\documentclass{article}

\usepackage{microtype}
\usepackage{graphicx}
\usepackage{subcaption}
\usepackage{booktabs} 
\usepackage{multirow}
\usepackage{hyperref}

\usepackage[preprint]{icml2026}

\usepackage{amsmath}
\usepackage{amssymb}
\usepackage{mathtools}
\usepackage{amsthm}

\usepackage[capitalize,noabbrev]{cleveref}

\theoremstyle{plain}
\newtheorem{theorem}{Theorem}[section]

\newtheorem{lemma}[theorem]{Lemma}
\newtheorem{corollary}[theorem]{Corollary}
\theoremstyle{definition}

\theoremstyle{remark}

\usepackage{enumitem}
\usepackage{pifont}
\usepackage[textsize=tiny]{todonotes}
\usepackage[table]{xcolor}
\definecolor{lightred}{rgb}{1, 0.8, 0.8}
\definecolor{lightgreen}{rgb}{0.9, 1, 0.9}
\icmltitlerunning{Understanding Post-Training Structural Changes in Large Language Models}

\begin{document}

\twocolumn[
  \icmltitle{Understanding Post-Training Structural Changes in Large Language Models}



  \icmlsetsymbol{equal}{*}

  \begin{icmlauthorlist}
    \icmlauthor{Xinyu He}{sch}
    \icmlauthor{Xianghui Cao}{sch}
  \end{icmlauthorlist}

  \icmlaffiliation{sch}{School of Automation, Southeast University, Nanjing, China}

  \icmlcorrespondingauthor{Xinyu He}{220242115@seu.edu.cn}
  \icmlcorrespondingauthor{Xianghui Cao}{xhcao@seu.edu.cn}

  \icmlkeywords{Machine Learning, ICML}

  \vskip 0.3in
]



\printAffiliationsAndNotice{}  

\begin{abstract}
  Post-training fundamentally alters the behavior of large language models (LLMs), yet its impact on the internal parameter space remains poorly understood. In this work, we conduct a systematic singular value decomposition (SVD) analysis of principal linear layers in pretrained LLMs, focusing on two widely adopted post-training methods: \textit{instruction tuning} and \textit{long-chain-of-thought (Long-CoT) distillation}. Our analysis reveals two unexpected and robust structural changes: \textbf{(1) a near-uniform geometric scaling of singular values across layers}; and \textbf{(2) highly consistent orthogonal transformations are applied to the left and right singular vectors of each matrix.} Based on these findings, We propose a simple yet effective framework to describe the coordinated dynamics of parameters in LLMs, which elucidates why post-training inherently relies on the foundational capabilities developed during pre-training. Further experiments demonstrate that singular value scaling underpins the temperature-controlled regulatory mechanisms of post-training, while the coordinated rotation of singular vectors encodes the essential semantic alignment. These results challenge the prevailing view of the parameter space in large models as a black box, uncovering the first clear regularities in how parameters evolve during training, and providing a new perspective for deeper investigation into model parameter changes.
\end{abstract}

\section{Introduction}

The remarkable success of large language models (LLMs) has been substantially facilitated by post-training techniques. With approaches such as instruction tuning \cite{ouyang2022traininglanguagemodelsfollow,zhang2024instructiontuninglargelanguage,peng2023instructiontuninggpt4}, alignment training \cite{schulman2017proximalpolicyoptimizationalgorithms,li2023reinforcementlearninghumanfeedback,rafailov2024directpreferenceoptimizationlanguage,deepseekai2025deepseekr1incentivizingreasoningcapability} and knowledge distillation \cite{xu2024surveyknowledgedistillationlarge,gu2024minillmknowledgedistillationlarge,mcdonald2024reducing,yang2024survey}, LLMs have become increasingly usable and better aligned with human intent \cite{guo2024largelanguagemodelbased,cai2025spatialbotprecisespatialunderstanding,feng2024largelanguagemodelbasedhumanagent}. Recent research on post-training has predominantly centered on algorithmic innovations such as \textit{Direct Preference Optimization} (DPO) \cite{rafailov2024directpreferenceoptimizationlanguage}, \textit{Group Relative Policy Optimization} (GRPO) \cite{deepseekai2025deepseekr1incentivizingreasoningcapability}, and \textit{Dynamic sAmpling 
Policy Optimization} (DAPO) \cite{yu2025dapo} to enhance the reasoning capabilities of LLMs. Alternatively, \textit{long-chain-of-thought (Long-CoT) distillation} offers a more straightforward and practiced approach, enabling smaller models to acquire reasoning ability by distilling long chains of thought from large RL-trained models \cite{deepseekai2025deepseekr1incentivizingreasoningcapability}. 

However, despite the empirical success of post-training, its underlying impact on the internal structure of model parameters remains insufficiently understood. Although recent studies have investigated post-training mechanisms and uncovered some novel insights \cite{du2025posttrainingreshapesllmsmechanistic,marks2024geometrytruthemergentlinear,jain2024mechanisticallyanalyzingeffectsfinetuning,lee2024mechanisticunderstandingalignmentalgorithms,panickssery2024steeringllama2contrastive,stolfo2024confidenceregulationneuronslanguage,katz2023visitvisualizinginterpretingsemantic,yao2025knowledgecircuitspretrainedtransformers}, their studies remain indirect—relying primarily on hidden representations or behavioral observations rather than exploring fundamental structural changes. Transformations in parameter space, especially weight matrices, which we often treat as black boxes, have not been systematically examined. \textbf{The extent to which post-training reshapes the representational capacity of the parameter space remains an unresolved problem.}

\vspace{-4pt}
\begin{figure*}[!htbp]
  \setlength{\abovecaptionskip}{2pt}
  \centering
  \includegraphics[width=0.88\textwidth]{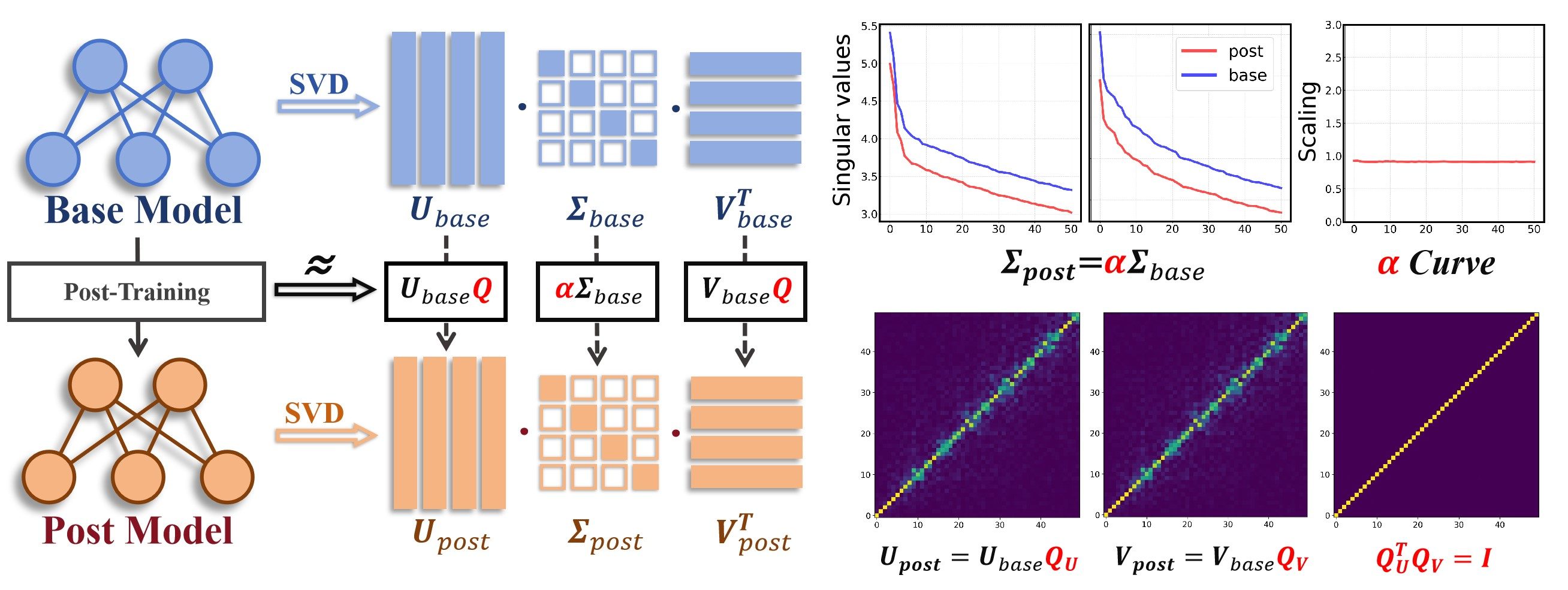}
  \caption{A simple but effective mathematical approximation to describe the effect of post-training on the parameter space. Performing SVD on weight matrices in the \textsc{base} model, post-training is equivalent to performing \textbf{linear scaling} on singular values and performing \textbf{consistent orthogonal transformations} on left and right singular vectors.}
  \vspace{-12pt}
  \label{head}
\end{figure*}

In this work, we present a systematic study on how post-training affects the parameter space of LLMs. Specifically, we focus on two supervised post-training methods: \textbf{instruction tuning} and \textbf{Long-CoT distillation}\footnote{For clarity and ease of reading, \textit{post-training} hereafter refers to both \textit{instruction tuning} and \textit{Long-CoT distillation} in the following sections unless otherwise specified.}. These methods underpin essential capabilities like instruction-following and reasoning, forming the basis for more advanced alignment techniques. To examine the structural impact of post-training, we leverage Singular Value Decomposition (SVD) to project the high-dimensional weight matrices into a set of orthogonal subspaces. By disentangling complex parameter structures into interpretable geometric components, this spectral perspective allows us to isolate and quantify the specific directional shifts and scaling variations induced by different post-training regimes. We apply this framework to the weight matrices within the Self-Attention modules and Feed-Forward Networks of publicly available models, and categorize models into three types: \textsc{base} models (e.g., \textit{Qwen2.5-Math-1.5B}  \cite{qwen2025qwen25technicalreport}), \textsc{instruct} models (trained via instruction tuning), and \textsc{reasoning} models (trained via long-CoT distillation). The latter two are collectively referred to as \textsc{post} models. This categorization enables systematic comparison of parameter space structural changes induced by different post-training methods.

Our empirical results reveal two unexpected regular patterns of post-training on the model's parameter space: \textbf{(1) Near-uniform geometric scaling of singular values}: Post-training preserves the singular value distribution of the \textsc{base} model while applying a global linear scaling factor. Notably, we observe anomalous scaling in the Attention module's $W_O$ matrix, which markedly distinguishes the \textsc{reasoning} models from the \textsc{instruct} models; \textbf{(2) Highly consistent orthogonal transformations}: The left and right singular vectors of each matrix undergo nearly identical orthogonal transformations during post-training, exhibiting shared, coordinated rotations, which essentially preserves the pre-trained semantic space.

Building upon these observations, \textbf{we propose a robust mathematical framework (Figure \ref{head}) to characterize the coordinated changes of parameters in LLMs.} From this parametric perspective, we elucidate why post-training inherently relies on the foundational capabilities established during pre-training. We then reveal that while uniform scaling reflects the temperature-controlled mechanisms of training, the shared orthogonal transformations encode the essential semantic alignment. Our findings also establish a parametric trajectory across post-training regimes, suggesting that distinct post-training methods yield essentially equivalent parametric effects.

We summarize our contributions as follows:
\begin{itemize}[leftmargin=10pt]
    \item \textbf{We experimentally report the first discovery of robust structural change laws governing the parameters of LLMs.} These two stable structural phenomena include near-uniform geometric scaling and consistent orthogonal transformations of each matrix, offering a novel perspective on the weight reshaping mechanism of models after post-training.
    \item \textbf{Formalization of a mathematical framework for post-training in parameter space.} We propose a simple yet effective mathematical framework that first formalizes the effects of post-training on model parameters and provides a theoretical explanation for why post-training must rely on the capabilities of pre-training.
    \item \textbf{Elucidation of the regulatory mechanisms of alignment.} Near-uniform geometric scaling of singular values correlates with post-training’s temperature-controlled regulatory mechanism, while singular vectors’ coordinated rotational dynamics encodes core semantic features acquired during post-training. Our findings also show that various post-training approaches produce equivalent parametric effects.  
\end{itemize}

\section{Related Work}
\textbf{Interpretability of Post-Training\quad} With the growing success of post-training, researchers have increasingly sought to uncover its underlying mechanisms. Several studies have attempted to investigate the impact of post-training on LLMs by constructing task-specific or instruction-formatted datasets \cite{du2025posttrainingreshapesllmsmechanistic,marks2024geometrytruthemergentlinear,jain2024mechanisticallyanalyzingeffectsfinetuning,lee2024mechanisticunderstandingalignmentalgorithms,panickssery2024steeringllama2contrastive,he2024learninggrokemergenceincontext}. However, since these studies treat the models more as black boxes, they provide limited insights into the structural changes in model parameters. Parallel lines of research have attempted to explain the behavior of large language models by analyzing individual neurons or sparse activation patterns, uncovering phenomena such as entropy neurons and task-specific circuits \cite{katz2023visitvisualizinginterpretingsemantic,yao2025knowledgecircuitspretrainedtransformers,gurnee2024universal,tang2024language,chen2024finding,yu2024neuronlevelknowledgeattributionlarge,gao2025hneuronsexistenceimpactorigin}. While these studies offer valuable insights, their scope is inherently limited, as they are often based on earlier models such as \textit{GPT-2} \cite{brown2020languagemodelsfewshotlearners}, reducing their relevance to contemporary architectures. Our analysis is data-agnostic, as we directly examine the full parameter space of the model rather than relying on input–output behavior. This perspective extends beyond previous studies that focus on individual neurons or isolated functional circuits, enabling a more global understanding of model structure.

\textbf{SVD in LLMs\quad} The optimal low-rank approximation property of SVD \cite{eckart1936approximation} has inspired a surge of SVD-based techniques for LLMs. Recent methods such as \textit{PiSSA} \cite{meng2024pissa}, \textit{SVFT} \cite{lingam2024svft}, and \textit{RaSA} \cite{he2025rasa} leverage dominant singular components to improve fine-tuning efficiency, while others employ SVD for quantization to reduce deployment costs \cite{li2024svdqunat,wang2024svd,qinsidobi,li2023loftq,yuan2023asvd}. Beyond its practical utility, SVD provides a principled framework for analyzing the internal structure of LLMs \cite{yang2023spectral, li2024happened}. For any weight matrix, reduced SVD decomposes the linear transformation into two orthogonal matrices spanning the input and output \textbf{semantic spaces} and a diagonal matrix governing directional scaling. In this view, the singular vectors define the basis of latent representations, determining how semantic information is projected and aligned across layers, while the singular values quantify the relative prominence of these features. This decomposition reveals that the geometry of LLM computation is encoded in the stability of its latent topologies, making SVD a vital lens for investigating how core knowledge is preserved or re-aligned during post-training.

\section{Preliminaries}
This section reviews the training pipeline and architectural components of LLMs. Given a vocabulary $\mathcal{V}$, we define LLMs as $\mathcal{M}:\mathcal{T}\rightarrow \mathcal{P}$, where $\mathcal{T}$ denotes the set of input token sequences $T_i=[t_1,t_2,...,t_n]_i \in \mathcal{T}$ and $\mathcal{P}$ is the probability space over $\mathcal{V}$. After $\mathcal{M}$ accepts sequences of input tokens $T_i$, a probability distribution $p_{\mathcal{M}} \in \mathcal{P}$ is output to predict the probability of the next token.

\textbf{Training Stages of LLMs\quad} LLMs are typically trained following a two-stage paradigm. The first stage, known as \text{pre-training}, involves optimizing a \textsc{base} model $\mathcal{M}_\text{base}$ to predict the next token given previous context, based on a large-scale corpus drawn from a large-scale distribution of natural language texts \cite{radford2018improving,sun2021ernie,yuan2022biobart}. The second stage, termed \textit{post-training}, further fine-tunes the pretrained model to align its behavior with specific objectives, such as following user instructions \cite{zhang2024instructiontuninglargelanguage} or performing complex reasoning \cite{deepseekai2025deepseekr1incentivizingreasoningcapability}. 

\textbf{Definition of \textsc{Base} and \textsc{Post} LLMs\quad} Depending on the post-training objective, the adapted model is referred to as an \textsc{instruct} model $\mathcal{M}_\text{Instruct}$ or a \textsc{reasoning} model $\mathcal{M}_\text{reasoning}$. The two models under discussion are collectively referred to as \textsc{post} models $\mathcal{M}_\text{post}$. The architectures of $\mathcal{M}_\text{base}$ and $\mathcal{M}_\text{post}$ are identical — all weight matrices share the same dimensionality, while the sole distinction lies in their respective parameterizations. In the main paper, $\mathcal{M}_\text{base}$ refers to \textit{Qwen2.5-Math-1.5B}, $\mathcal{M}_\text{Instruct}$ to its instruction-tuned variant \textit{Qwen2.5-Math-1.5B-Instruct}, and $\mathcal{M}_\text{reasoning}$ to the distilled \textit{reasoning} model \textit{DeepSeek-R1-Distill-Qwen-1.5B}. $\mathcal{M}_\text{Instruct}$ and $\mathcal{M}_\text{reasoning}$ can both be expressed as $\mathcal{M}_\text{post}$. Results for other models across different families and parameter scales are provided in the Appendix.

\begin{figure*}[!ht]
    \setlength{\abovecaptionskip}{2pt}
    \centering
    \includegraphics[width=0.89\linewidth]{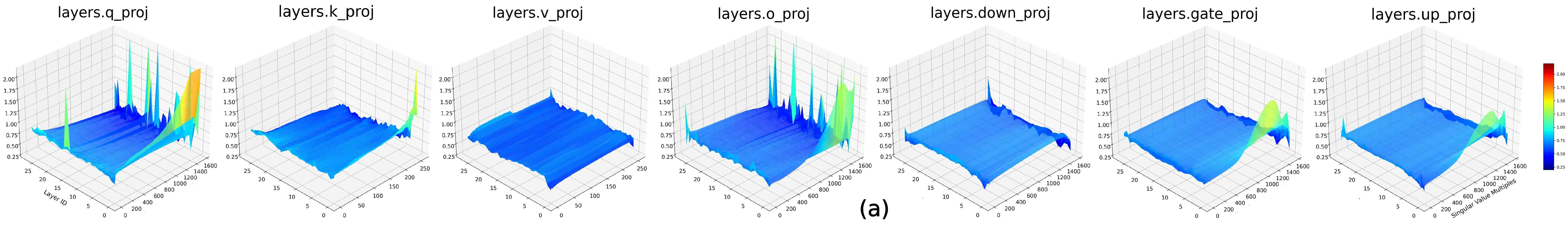}
    \includegraphics[width=0.89\linewidth]{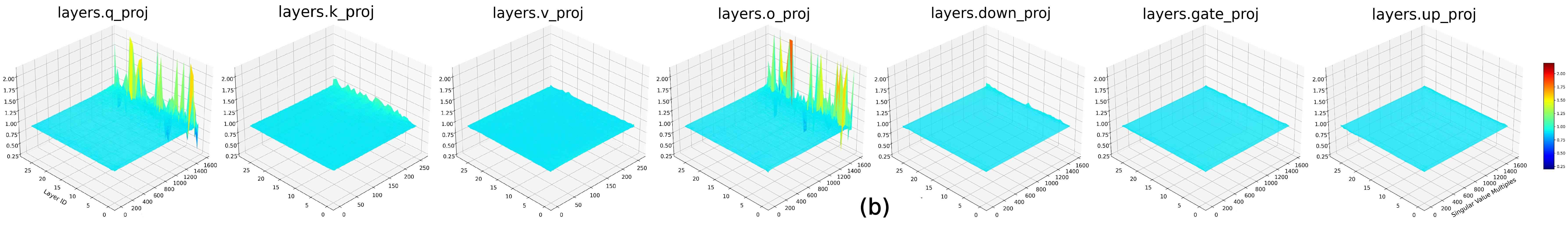}    
    \includegraphics[width=0.89\linewidth]{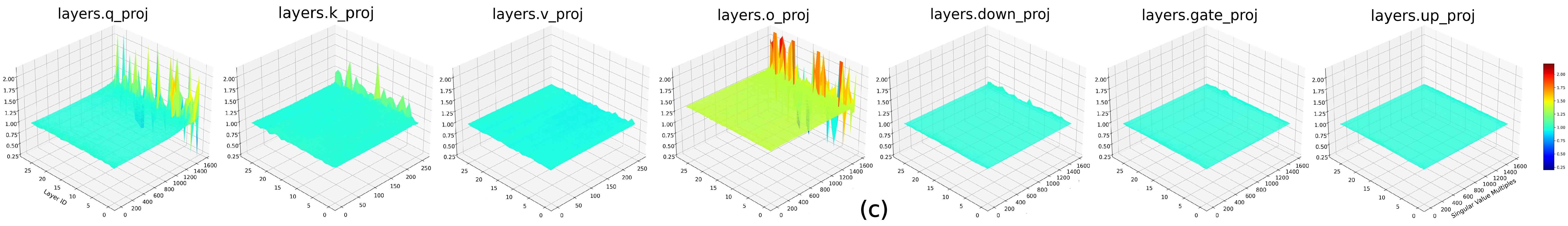} 
    \caption{The heatmaps of SVSMs comparing $\mathcal{M}_\text{base}$ with $\mathcal{M}'_\text{base}$, $\mathcal{M}_\text{Instruct}$ and $\mathcal{M}_\text{reasoning}$.  (a) indicates no regular pattern in the distribution of scaling factors between $\mathcal{M}'_\text{base}$ and $\mathcal{M}_\text{base}$. In both (b) and (c), the principal scaling exhibits a near-uniform distribution. While in (c), scaling factors of $W_O$ are significantly higher than those of other matrix types.}  
    \label{SVSMofBaseAndInstruct}
\end{figure*}

\textbf{Architectural Components of LLMs\quad} We focus on decoder-only Transformer-based models,  which constitute the foundation of state-of-the-art large language model systems \cite{GPT4,deepseekai2024deepseekv3,gemmateam2025gemma3technicalreport}. The Transformer architecture consists of two core components: the Self-Attention Module (SA) and the Feed-Forward Network (FFN) \cite{vaswani2023attentionneed}. Given an input hidden vector ${h}^T \in \mathbb{R}^{d_{model}}$, we consider the simplest form of attention calculation for concise illustration. The output of the SA is:
\begin{equation}
\begin{split}
  Attn(h) &= \text{softmax} \left( \frac{hW_Q \cdot [K_{\text{cache}} ; hW_K ]^T}{\sqrt{d}} \right) \\
  SA(h) &= Attn(h) \cdot [ V_{\text{cache}} ; hW_V] W_O
  \label{SA}
\end{split}
\end{equation}

where $W_Q, W_K, W_V, W_O \in \mathbb{R}^{{d_{\text{model}}} \times {d_{\text{model}}}}$ are learnable weight matrices,$\sqrt{d}$ is the scaling factor in the attention map, $K_{\text{cache}}$ and $V_{\text{cache}}$ are the key and value caches respectively, and $[...;...]$ denotes concatenation. Modern architectures like the Qwen2.5 series use GQA variants \cite{ainslie2023gqatraininggeneralizedmultiquery} to optimize attention computation, yet core projection matrices remain integral for defining the attention mechanism’s representational capacity. Given an input vector ${z}^T \in \mathbb{R}^{d_{model}}$, the output of the FFN, which employs the \textit{SwiGLU} activation function \cite{shazeer2020gluvariantsimprovetransformer}, is:
\begin{equation}
  FFN(z)=(SwiGLU(z\cdot W_{gate})\odot (z\cdot W_{up}))\cdot W_{down}
  \label{FFN}
\end{equation}

where $W^T_{down}, W_{gate}, W_{up} \in \mathbb{R}^{{d_{model}} \times {d_{mlp}}}$ are learnable weight matrices. Notably, GQA and SwiGLU-based FFNs have become fundamental building blocks adopted across numerous open-source LLMs, including \textit{Qwen} \cite{qwen2025qwen25technicalreport}, \textit{LLaMA} \cite{grattafiori2024llama3herdmodels}, \textit{Mistral} \cite{jiang2023mistral7b}, \textit{Phi-4} \cite{abdin2024phi4technicalreport}, \textit{gpt-oss} \cite{openai2025gptoss120bgptoss20bmodel}, \textit{Gemma} \cite{gemmateam2025gemma3technicalreport} and others \cite{glm2024chatglmfamilylargelanguage, yang2025baichuan2openlargescale,deepseekai2024deepseekllmscalingopensource}. We specifically focus on the weight matrices in SAs and FFNs, which account for the majority of parameters in LLMs.

\section{The Structural Changes of Singular Space After Post-Training}
\label{main_finding}
This section formally presents two unexpected structural changes that occur in the singular space of LLMs after post-training. Assuming that $m \leq n$, the reduced SVD of a matrix $W \in \mathbb{R}^{m \times n}$ is given by $W = U \Sigma V^{T}$, where $U \in \mathbb{R}^{{m \times m}}$ and $V^T \in \mathbb{R}^{{m \times n}}$ are matrices with orthogonality whose columns correspond to the left and right singular vectors respectively. The diagonal matrix $\Sigma = \operatorname{diag}(\sigma_1, \sigma_2, \dots, \sigma_n) \in \mathbb{R}^{n \times n}$ contains the singular values arranged in descending order.

\subsection{Near-Uniform Geometric Scaling of Singular Values}
Departing from the prevailing premise in previous studies \cite{lingam2024svft,he2025rasa} that post-training is prone to alter the singular value distribution, we observe that this process does not alter the overall singular value distribution established during pre-training in the \textsc {base} model. Instead, it exhibits \textbf{a near-uniform geometric scaling behavior}, characterized by approximately consistent scaling factors across the main singular values.

\begin{figure*}
    \setlength{\abovecaptionskip}{2pt}
    \centering
    \includegraphics[width=1.0\linewidth]{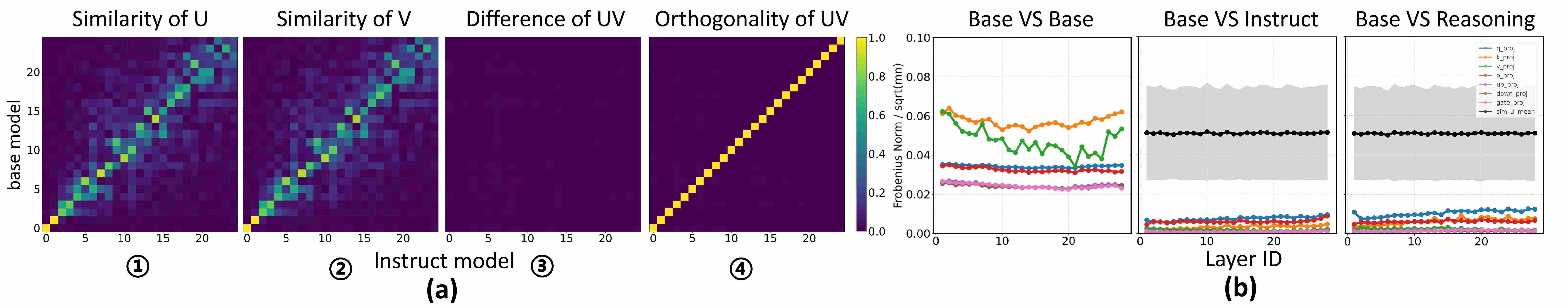}
    \caption{An example showing the orthogonality of singular vector similarity to the transformation performed. Only the first 25 dimensions are retained for clearer visualization. (a) shows the singular vector behavior of $W_O$ in the first Transformer block. Difference matrix (\ding{174}) represents $\smash{\big|sim^{(0)}_{U}-sim^{(0)}_{V}\big|}$, which is almost a zero matrix. \ding{175} is $\smash{I^{(0)}_{orth}}$ of $\smash{W^{(0)}_O}$. Most of its diagonal elements are close to 1, and the rest are basically 0. (b) extensively verifies the approximate equality of $\smash{Q^{(i)}_1}$ and $\smash{Q^{(i)}_2}$ comparing $\mathcal{M}_\text{base}$ to ${\mathcal{M}'_\text{base}}$ and $\mathcal{M}_\text{post}$.}  
    \label{sim_UV}
    \vspace{-12pt}
\end{figure*}

\textit{\textbf{Observation 1\quad}} For the $i$-th Transformer block of $\mathcal{M}_{A}$ and $\mathcal{M}_{B}$ of the same architecture, we perform reduced SVD on weight matrix:
\begin{equation}
    \begin{split}
  W^{(i)}_{A}=U^{(i)}_{A} \cdot diag(\sigma^{(i)}_{A,1},\sigma^{(i)}_{A,2},...,\sigma^{(i)}_{A,n})\cdot {V^{(i)}_{A}}^T \\
  W^{(i)}_{B}=U^{(i)}_{B}\cdot diag(\sigma^{(i)}_{B,1},\sigma^{(i)}_{B,2},...,\sigma^{(i)}_{B,n})\cdot {V^{(i)}_{B}}^T
  \end{split}
  \label{SVD_on_A_and_B}
\end{equation}
where $W^{(i)}_{A}\in \mathcal{M}_{A}$ and $W^{(i)}_{B}\in \mathcal{M}_{B}$ represent weight matrices of the same type in the $i$-th Transformer block (e.g. $W_Q$) but belonging to different models. To quantify the effect of post-training on the evolution of singular value distribution, we define the \textit{Singular Value Scaling Matrix} (SVSM) as:
\begin{equation}
  \begin{split}
  SVSM(\frac{\mathcal{M}_{B}}{\mathcal{M}_{A}})&=[Div^{(1)},Div^{(2)},...,Div^{(k)}] \\
  Div^{(i)}&=[\frac{\sigma^{(i)}_{B,1}}{\sigma^{(i)}_{A,1}},...,\frac{\sigma^{(i)}_{B,n}}{\sigma^{(i)}_{A,n}}]^T
  \end{split}
  \label{SVD_div}
\end{equation}
where $k$ corresponds to the depth of architecture $\mathcal{M}_{A}$ or $\mathcal{M}_{B}$. $\alpha^{(i)}={\sigma^{(i)}_{B,j}}/{\sigma^{(i)}_{A,j}},j=1,2,...,n$ is the scaling factor. SVSM actually describes the distribution of all scaling factors across layers. We plot the heatmaps of $\displaystyle{{SVSM(\tfrac{\mathcal{M}_\text{Instruct}}{\mathcal{M}_\text{base}})}}$ (Figure~\ref{SVSMofBaseAndInstruct}b) and $\smash{SVSM(\tfrac{\mathcal{M}_\text{reasoning}}{\mathcal{M}_\text{base}})}$ (Figure~\ref{SVSMofBaseAndInstruct}c) as examples. For reference comparison, we also show heatmaps of  ${SVSM(\tfrac{\mathcal{M}'_\text{base}}{\mathcal{M}_\text{base}})}$ where  $\smash{\mathcal{M}'_\text{base}}$ denotes \textit{Qwen2.5-1.5B}, which shares the same architecture but differs in pre-training data (Figure~\ref{SVSMofBaseAndInstruct}a). 

\textbf{Near-uniform Geometric Scaling\quad} A core phenomenon emerges: singular values of all weight types exhibit near-uniform geometric scaling, \textbf{with all cross-layer weight matrices sharing a single linear scaling factor}. This pattern prevails across both post-training paradigms corresponding to $\mathcal{M}_\text{Instruct}$ and $\mathcal{M}_\text{reasoning}$, showing a strictly consistent global cross-layer scaling effect relative to $\mathcal{M}_\text{base}$. Only tail singular values exhibit trivial fluctuations, which are negligible in magnitude and barely impact the overall linear transformation. Such behavior is well captured by $\smash{\Sigma_\text{post} \approx \alpha \Sigma_\text{base}}$, where $\alpha$ acts as the sole linear scaling factor shared by all cross-layer weight matrices and governing all principal singular values. In contrast, no such global cross-layer consistency exists between $\mathcal{M}'_\text{base}$ and $\mathcal{M}_\text{base}$.  

\textbf{The Scaling Specificity of $\mathbf{W_o}$\quad} We further observe that scaling factors of $W_O$ in $\mathcal{M}_\text{reasoning}$ consistently exceed those of other matrix types, which can be used to distinguish non-reasoning models. This pattern holds uniformly across all \textsc{reasoning} models in our study. Detailed quantitative data (Table \ref{svsf_table}) and visualizations of other models across different families and parameter scales are in Appendix \ref{Visualization_of_other_families_and_scales}.

\subsection{Consistent Orthogonal Transformations of Singular Vectors}
We investigate the similarity between the singular vectors of \textsc{base} models and \textsc{post} models. It is significant to find that the similarity matrices of both left and right singular vectors remain nearly identical after post-training, suggesting that the input and output subspaces undergo consistent orthogonal transformations during this process.

Combining Equation \ref{SVD_on_A_and_B}, the similarity matrices of $W^{(i)}_{A}$ and $W^{(i)}_{B}$ are defined as:
\begin{equation}
    \begin{split}
    sim^{(i)}_{U}(\frac{\mathcal{M}_{A}}{\mathcal{M}_{B}})&={U^{(i)}_{A}}^T\cdot{U^{(i)}_{B}}\\
    sim^{(i)}_{V}(\frac{\mathcal{M}_{A}}{\mathcal{M}_{B}})&={V^{(i)}_{A}}^T\cdot{V^{(i)}_{B}}
    \end{split}
    \label{sim}
\end{equation}
\textit{\textbf{Observation 2\quad}} The widely observed phenomenon can be expressed as $|{sim^{(i)}_{U}(\frac{\mathcal{M}_\text{base}}{\mathcal{M}_\text{post}})| \approx |sim^{(i)}_{V}(\frac{\mathcal{M}_\text{base}}{\mathcal{M}_\text{post}})}|$ (\ding{172}-\ding{174} in Figure \ref{sim_UV}a), where $|\cdot|$ takes the absolute value of each matrix element to remove the possible sign ambiguity of singular vectors, which implies that the input and output subspaces of LLMs are undergoing highly symmetrical changes.

Based on this observation, we can theoretically prove that the similarity matrices of the left and right singular vectors can be directly used to describe the transformation dynamics within the parameter space of LLMs, and \textbf{only rotate the orthogonal bases already formed during the pre-training of LLMs}:
\begin{corollary}\label{cor:ortho_transform_colspace}
For $\smash{\mathcal{M}_{\text{base}} \rightarrow \mathcal{M}_{\text{post}}}$, the left and right singular vectors of $\smash{W^{(i)}_{\text{base}}}$ to $\smash{W^{(i)}_{\text{post}}}$ undergo coordinated orthogonal transformations: $\smash{U^{(i)}_{\text{post}} = U^{(i)}_{\text{base}} Q^{(i)}_1}$, $\smash{V^{(i)}_{\text{post}} = V^{(i)}_{\text{base}} Q^{(i)}_2}$, with $\smash{Q^{(i)}_1 \approx Q^{(i)}_2 = \text{sim}^{(i)}_{U/V}}$, and $\smash{\text{col}(U^{(i)}_{\text{base}}) = \text{col}(U^{(i)}_{\text{post}})}$, $\smash{\text{col}(V^{(i)}_{\text{base}}) = \text{col}(V^{(i)}_{\text{post}})}$.
\end{corollary}
where $\smash{Q^{(i)}_1}$ and $\smash{Q^{(i)}_2}$ are transformation matrices. {The proof of corollary \ref{cor:ortho_transform_colspace} is given in Appendix \ref{Proof_of_Orthogonality}, which strongly reflects the collaborative and consistent variation of the input and output subspaces.} 

We validate this claim by leveraging the properties of orthogonal matrices: 
\begin{lemma}
If $Q^{(i)}_1 = Q^{(i)}_2$, then ${Q^{(i)}_1}^T Q^{(i)}_2 = I^{(i)}_{\text{orth}} = I$, where $Q^{(i)}_1 = {U^{(i)}_\text{base}}^{T} U^{(i)}_\text{post}$ and $Q^{(i)}_2 = {V^{(i)}_\text{base}}^{T} V^{(i)}_\text{post}$.
\end{lemma}
where $I\in \mathbb{R}^{n \times n}$ is the identity matrix. We quantify the orthogonality and the equality between ${Q^{(i)}_1}$ and ${Q^{(i)}_2}$ by measuring the proximity of $\smash{I^{(i)}_{orth}}$ to $I$, employing the normalized Frobenius norm ${\mathcal{NF}^{(i)} = \mathcal{F}^{(i)}(I^{(i)}_{orth}-I)/\sqrt{n^2}=\mathcal{F}^{(i)}(I^{(i)}_{orth}-I)/n}$ as our metric. {To eliminate the possibility of low $\smash{\mathcal{NF}^{(i)}}$ due to insufficient training, we also plot the mean and standard deviation of $\smash{\mathcal{NF}^{(i)}_{\mathrm{sim}} = \mathcal{F}^{(i)}((\mathrm{sim}^{(i)}_{U} - I)/n)}$ as line plots (shaded regions denote standard deviation) for all matrix types in each Transformer block.}

\textbf{Consistent Orthogonal Transformations\quad} \ding{175} in Figure \ref{sim_UV}a presents our visualization of $\smash{I^{(0)}_{orth}}$ for $W^{(0)}_O$, and Figure \ref{sim_UV}b illustrates $\smash{\mathcal{NF}^{(i)}}$ and $\smash{\mathcal{NF}^{(i)}_\text{sim}}$ in all the weight matrices of the layers. It can be observed that for $\mathcal{M}_\text{post}$, the values of ${\mathcal{NF}^{(i)}}$ are consistently and significantly lower than those of $\mathcal{M}'_\text{base}$ across all layers while $\smash{\mathcal{NF}^{(i)}_{\text{sim}}}$ sustains a persistently high magnitude, directly demonstrating that ${Q^{(i)}_1}$ and ${Q^{(i)}_2}$ are approximately equal orthogonal matrices throughout post-training. We can further conclude that the variation in the singular vectors can be approximately characterized by consistent orthogonal transformations, a property absent in different pretrained models (see Appendix~\ref{Visualization_of_other_families_and_scales_orth_sec2}). More detailed test results are in Appendix \ref{Visualization_of_other_families_and_scales_orth}.

\subsection{Extensiveness of Structural Changes}
We have conducted extensive identical observational experiments on the RL methods (including \textit{GRPO} and \textit{PPO}), across different model families, and throughout the entire post-training process of models. The results demonstrate that the two post-training structural changes in the singular space exhibit remarkable extensiveness and robustness. These changes are not constrained by training paradigms or model architectures, and remain stable from start to finish during training. Identical patterns are also observed in other functionally specific components of LLMs. \textbf{Such findings indicate that those stable phenomena is a fundamental law governing parameter variations in LLMs during post-training}. Details are provided in the Appendix \ref{case_study}.

\section{Analysis of Post-Training}
\label{Analysis}
\textbf{Mathematical Model\quad} Based on the observation of the aforementioned phenomena, \textbf{we propose a simple but novel mathematical model of the weight changes from $\smash{\mathcal{M}_\text{base} \rightarrow \mathcal{M}_\text{post}}$}, which prior work has struggled to describe formally \cite{du2025posttrainingreshapesllmsmechanistic,marks2024geometrytruthemergentlinear,jain2024mechanisticallyanalyzingeffectsfinetuning,lee2024mechanisticunderstandingalignmentalgorithms}. For $W_\text{base}\in \mathcal{M}_\text{base}$ and $W_\text{post}\in \mathcal{M}_\text{post}$, the changes imposed by post-training on the parameters can be approximated by \textbf{a linear factor $\boldsymbol{\alpha}$} and \textbf{an orthogonal matrix $\mathbf{Q}$}:
\begin{equation}
    \begin{split}
    W_\text{post}&=U_\text{post}\Sigma_\text{post}V^T_\text{post} \\
    &\approx (U_\text{base}Q)\cdot(\alpha \Sigma_\text{base})\cdot (V_\text{base}Q)^T
    \label{approx_W}
    \end{split}
\end{equation}

\textbf{The Essence of Post-Training\quad} Equation \ref{approx_W} provides a mechanistic perspective on how post-training builds upon pre-training. $\mathbf{Q}$ induces a rotation that aligns the subspaces spanned by $U_\text{base}$ and $V_\text{base}$ to those spanned by $U_\text{post}$ and $V_\text{post}$, preserving the intrinsic geometric structure of the foundational semantic space. Critically, this rotation enables the model to adapt to the post-training data distribution, thereby unlocking the model's underlying capabilities. \textbf{Thus, only when pre-training data is of high quality, forming an optimal semantic space with well-structured bases, can post-training fully release the model's fundamental potential by reorienting these bases.} This perspective formalizes why pre-training data quality sets the capability ceiling, while post-training serves as the key to activating that potential through parametric adaptation \cite{akter2025frontloadingreasoningsynergypretraining,qi2025evolmsearchlostlanguage,li2025tracing}.


This section experimentally verifies the correctness of Equation \ref{approx_W}, and shows that the near-uniform scaling of singular values corresponds to the temperature-controlled mechanism induced in post-training, while the coordinated rotation of singular vectors encodes the core semantic information learned in post-training. These results confirm a strong equivalence in the impact of post-training on the parameter space, regardless of variations in data distribution.

\subsection{Singular Value Scaling is a Temperature-Controlled Mechanism}
\label{temperature_mechanism}
\textbf{Verification Experiments\quad}To validate the effectiveness of singular value scaling in Equation \ref{approx_W}, a direct consequence is that models before and after singular value replacement should exhibit nearly identical performance. 

For $\mathcal{M}_{\text{post}}$, we apply Construction \ref{scaling SV with alpha} to all weight matrices across the transformer blocks, which replaces the singular values of $\mathcal{M}_{\text{post}}$ with the singular values of $\mathcal{M}_{\text{base}}$ scaled by a given linear factor $\alpha'$. In addition, to eliminate the influence of $\alpha'$, we apply Construction \ref{scaling SV with alpha_1} to construct the control group:
\begin{equation}
  W^{(i)}_\text{post} \leftarrow U^{(i)}_\text{post} \cdot (\boldsymbol{\alpha'\Sigma^{(i)}_\text{base}}) \cdot {V^{(i)}_\text{post}}^T
  \label{scaling SV with alpha}
\end{equation}
\begin{equation}
  W^{(i)}_\text{post} \leftarrow U^{(i)}_\text{post} \cdot (\boldsymbol{\alpha'\Sigma'^{(i)}_\text{base}}) \cdot {V^{(i)}_\text{post}}^T
  \label{scaling SV with alpha_1}
\end{equation}
$\Sigma'^{(i)}_\text{base}$ denotes the singular values of $\mathcal{M}'_{\text{base}}$. As shown in Figure \ref{SVSMofBaseAndInstruct}, its singular value distribution differs significantly from that of $\mathcal{M}_{\text{base}}$.We denote the resulting model after substitution of singular values as $\mathcal{M}^{\text{replaced}}_{\text{post}}$ and $\mathcal{M}^{\text{controlled}}_{\text{post}}$. The choice of $\alpha'$ is shown in Table \ref{alpha'_table}. We then evaluate both $\mathcal{M}_{\text{post}}$ and $\mathcal{M}^{\text{replaced}}_{\text{post}}$ on four standard benchmarks: GSM8K \citep{cobbe2021trainingverifierssolvemath}, MATH-500 \citep{math500}, MMLU (dev split) \citep{mmlu}, and GPQA \citep{rein2023gpqagraduatelevelgoogleproofqa}.Performance is measured under a token limit of 1024. The results are shown in Table \ref{replaced_acc}. 

\begin{table}[!htbp]
  \caption{Performance comparison between original and replaced models across GSM8K, MATH-500, MMLU, and GPQA with pass@1 accuracy (\%). all evaluations are conducted with three independent repetitions, and the average values are reported.}
  \label{replaced_acc}
  \begin{center}
    \begin{small}
      \resizebox{0.45\textwidth}{!}{%
        \begin{tabular}{llll}
          \toprule
          \textsc{Base} Models & \textsc{Replaced} Types  & GSM8K  & MATH-500 \\
          \midrule
          \multirow{13}{*}[-0.5ex]{\shortstack{\textit{Qwen2.5-}\\ \textit{Math-1.5B}}}
          & $\mathcal{M}_\text{Instruct}$ & 85.14$\pm$0.14 & 65.47$\pm$0.90 \\
          & \cellcolor{lightgreen}$\boldsymbol{\mathcal{M}^{\textbf{replaced}}_\textbf{Instruct}}$ & \cellcolor{lightgreen}\textbf{85.59}$\boldsymbol{\pm}$\textbf{0.09} & \cellcolor{lightgreen}\textbf{61.67}$\boldsymbol{\pm}$\textbf{0.57} \\
          & \cellcolor{lightred}$\boldsymbol{\mathcal{M}^{\textbf{controlled}}_\textbf{Instruct}}$ & \cellcolor{lightred}\textbf{0.00}$\boldsymbol{\pm}$\textbf{0.00} & \cellcolor{lightred}\textbf{0.00}$\boldsymbol{\pm}$\textbf{0.00} \\
          & $\mathcal{M}_\text{reasoning}$ & 62.88$\pm$0.59 & 32.73$\pm$1.64 \\
          & \cellcolor{lightgreen}$\boldsymbol{\mathcal{M}^{\textbf{replaced}}_\textbf{reasoning}}$ & \cellcolor{lightgreen}\textbf{69.45}$\boldsymbol{\pm}$\textbf{0.43} & \cellcolor{lightgreen}\textbf{41.46}$\boldsymbol{\pm}$\textbf{0.53} \\
          & \cellcolor{lightred}$\boldsymbol{\mathcal{M}^{\textbf{controlled}}_\textbf{reasoning}}$ & \cellcolor{lightred}\textbf{0.00}$\boldsymbol{\pm}$\textbf{0.00} & \cellcolor{lightred}\textbf{0.00}$\boldsymbol{\pm}$\textbf{0.00} \\
          \cmidrule{2-4}
          & \textsc{Replaced} Types  & MMLU (dev) & GPQA \\
          \cmidrule{2-4}
          & $\mathcal{M}_\text{Instruct}$ & 48.04$\pm$0.60 & 30.44$\pm$0.36 \\
          & \cellcolor{lightgreen}$\boldsymbol{\mathcal{M}^{\textbf{replaced}}_\textbf{Instruct}}$ & \cellcolor{lightgreen}\textbf{49.47}$\boldsymbol{\pm}$\textbf{0.29} & \cellcolor{lightgreen}\textbf{25.99}$\boldsymbol{\pm}$\textbf{0.70} \\
          & \cellcolor{lightred}$\boldsymbol{\mathcal{M}^{\textbf{controlled}}_\textbf{Instruct}}$ & \cellcolor{lightred}\textbf{0.00}$\boldsymbol{\pm}$\textbf{0.00} & \cellcolor{lightred}\textbf{0.00}$\boldsymbol{\pm}$\textbf{0.00} \\
          & $\mathcal{M}_\text{reasoning}$ & 25.02$\pm$0.59 & 7.02$\pm$0.44 \\
          & \cellcolor{lightgreen}$\boldsymbol{\mathcal{M}^{\textbf{replaced}}_\textbf{reasoning}}$ & \cellcolor{lightgreen}\textbf{35.52}$\boldsymbol{\pm}$\textbf{0.81} & \cellcolor{lightgreen}\textbf{9.45}$\boldsymbol{\pm}$\textbf{1.59} \\
          & \cellcolor{lightred}$\boldsymbol{\mathcal{M}^{\textbf{controlled}}_\textbf{reasoning}}$ & \cellcolor{lightred}\textbf{0.00}$\boldsymbol{\pm}$\textbf{0.00} & \cellcolor{lightred}\textbf{0.00}$\boldsymbol{\pm}$\textbf{0.00} \\
          \bottomrule
        \end{tabular}%
      }
    \end{small}
  \end{center}
  \vskip -0.1in
\end{table}

It can be observed that $\mathcal{M}^{\text{replaced}}_{\text{post}}$ maintains the performance of $\mathcal{M}_{\text{post}}$, while $\mathcal{M}^{\text{controlled}}_{\text{post}}$ suffers from significant model collapse, which once again illustrates the importance of Equation \ref{approx_W} and verifies that post-training does not alter the singular value distribution of the original model. Detailed experimental setups, the selection method of $\alpha'$, and results across different model scales and families are provided in Appendix \ref{test_replaced_model}.

\begin{figure}[!htbp]
    \vspace{-8pt}
    \centering
    \includegraphics[width=0.85\linewidth]{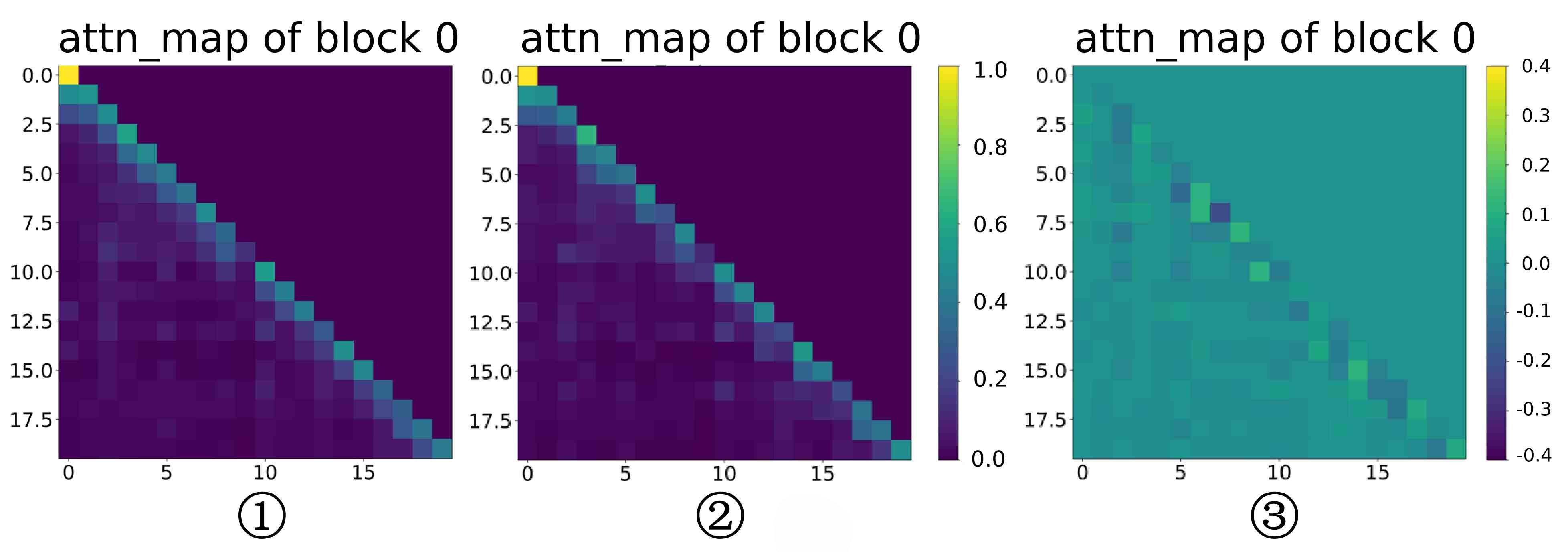}
    \caption{Visualization of attention map before and after replacing the singular values. \ding{172} shows the original attention heads, while \ding{173} presents the averaged attention heads from the modified model. \ding{174} illustrates the differences between those patterns. }
    \label{attn_behavior_one_1}
    \vspace{-12pt}
\end{figure}

\textbf{Singular Value Scaling is a Temperature-Controlled Mechanism\quad}We rigorously prove that singular value scaling is equivalent to adjusting the attention temperature (see proof in Appendix \ref{Proof_of_Replacement}), and thus equivalent to modulating the attention map and the final output confidence of models, as illustrated in Figure \ref{attn_behavior_one_1}. 

Due to its global consistency, this uniform scaling forces each layer to adjust the latent representation to the same extent, \textbf{thereby smoothing or concentrating the probability distribution of the final signals through cumulative effects. It will influence the output token length of LLM, as we measured in Table \ref{replaced_token_length}.} A strong manifestation is the abnormally high value of $W_o$ in $\mathcal{M}_{\text{reasoning}}$, which concentrates the confidence of $\mathcal{M}_{\text{reasoning}}$ toward a few tokens. When the model is uncertain about whether to terminate, this concentration makes the $<$eos$>$ token less likely to be sampled, thereby encouraging longer reasoning chains. This observation provides a valuable perspective on the feasibility of \textit{test-time scaling} \cite{muennighoff2025s1simpletesttimescaling}.

\begin{figure}[!htbp]
    \vspace{-6pt}
    \centering
    \includegraphics[width=0.85\linewidth]{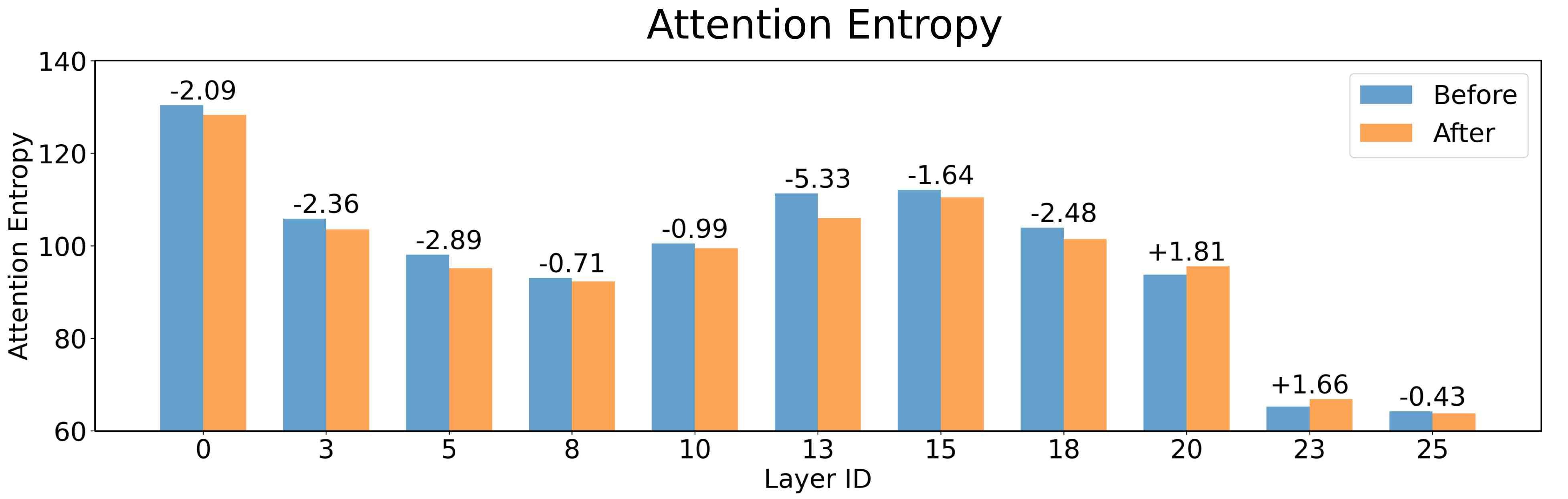}
    \caption{Visualization of the attention entropy before and after replacing the singular values. This suggests that post-training does not structurally modify the singular value spectrum.}
    \label{attn_behavior_one_2}
    \vspace{-12pt}
\end{figure}

\textbf{Structural Invariance\quad} However, This phenomenon suggests that post-training adjustments to singular values do not fundamentally reshape the model's internal representational logic. By constructing $\mathcal{M}_{\text{replaced}}$ via Construction \ref{temperature-controlled} and tracking the \textit{attention entropy} $\mathcal{H}$ \citep{kumar2019calibrationencoderdecodermodels}, we observe that the entropy profile remains remarkably stable before and after replacement (Figure \ref{attn_behavior_one_2}).

Such invariance implies that the \textbf{core semantic information is primarily encoded within the rotations of singular vectors rather than being heavily carried by the magnitude of singular values.} Consequently, singular value replacement functions as a temperature-controlled mechanism: it recalibrates the energy distribution and decisiveness of the model while preserving the underlying semantic framework established during pre-training. More detailed results are provided in Appendix \ref{Attention_Behavior}.

\subsection{Consistent Orthogonal Transformations Encode Core Semantic Information}

\textbf{Pairing Relationship in Semantic Space\quad} We argue that the consistent orthogonal transformation $Q$ is a structural prerequisite for preserving the model's core knowledge. By maintaining the \textbf{pairing relationship} between the input and output singular vectors, post-training ensures that the fundamental semantic space constructed during pre-training remains stable and coherent. This equivalence of subspaces represents a unified realignment of the coordinates in the latent representation, enabling the model to efficiently adapt to new tasks while preventing the destruction of the underlying knowledge topology. Consequently, any decoupling of this coherence would disrupt the functional association between input and output features, thereby destroying the fundamental semantic space formed during pre-training and leading to a catastrophic collapse of the model’s reasoning capabilities.

\begin{table}[!htbp]
  \caption{Performance comparison between original and \textsc{restoration} models across GSM8K, MATH-500, MMLU, and GPQA with pass@1 accuracy (\%).}
  \label{restoration_acc}
  \begin{center}
    \begin{small}
      \resizebox{0.45\textwidth}{!}{%
        \begin{tabular}{llll}
          \toprule
          \textsc{Base} Models & \textsc{Restoration} Types  & GSM8K  & MATH-500 \\
          \midrule
          \multirow{13}{*}[-0.5ex]{\shortstack{\textit{Qwen2.5-}\\ \textit{Math-1.5B}}} 
          & $\mathcal{M}_\text{Instruct}$ & 85.14$\pm$0.14 & 65.47$\pm$0.90 \\
          & \cellcolor{lightred}${\mathcal{M}^{\textbf{ablation}}_\textbf{Instruct}}$ & \cellcolor{lightred}0.00$\pm$0.00 & \cellcolor{lightred}0.00$\pm$0.00 \\
          & \cellcolor{lightgreen}$\boldsymbol{\mathcal{M}^{\textbf{restoration}}_\textbf{Instruct}}$ & \cellcolor{lightgreen}\textbf{84.53}$\boldsymbol{\pm}$\textbf{0.25} & \cellcolor{lightgreen}\textbf{66.20}$\boldsymbol{\pm}$\textbf{0.16} \\
          & $\mathcal{M}_\text{reasoning}$ & 62.88$\pm$0.59 & 32.73$\pm$1.64 \\
          & \cellcolor{lightred}${\mathcal{M}^{\textbf{ablation}}_\textbf{reasoning}}$ & \cellcolor{lightred}0.00$\pm$0.00 & \cellcolor{lightred}0.00$\pm$0.00 \\
          & \cellcolor{lightgreen}$\boldsymbol{\mathcal{M}^{\textbf{restoration}}_\textbf{reasoning}}$ & \cellcolor{lightgreen}\textbf{61.54}$\boldsymbol{\pm}$\textbf{1.19} & \cellcolor{lightgreen}\textbf{30.93}$\boldsymbol{\pm}$\textbf{0.57} \\
          \cmidrule{2-4}
          & \textsc{Restoration} Types  & MMLU (dev) & GPQA \\
          \cmidrule{2-4}
          & $\mathcal{M}_\text{Instruct}$ & 48.04$\pm$0.60 & 30.44$\pm$0.36 \\
          & \cellcolor{lightred}${\mathcal{M}^{\textbf{ablation}}_\textbf{Instruct}}$ & \cellcolor{lightred}0.00$\pm$0.00 & \cellcolor{lightred}0.00$\pm$0.00 \\
          & \cellcolor{lightgreen}$\boldsymbol{\mathcal{M}^{\textbf{restoration}}_\textbf{Instruct}}$ & \cellcolor{lightgreen}\textbf{41.28}$\boldsymbol{\pm}$\textbf{0.44} & \cellcolor{lightgreen}\textbf{27.69}$\boldsymbol{\pm}$\textbf{0.29} \\
          & $\mathcal{M}_\text{reasoning}$ & 25.02$\pm$0.59 & 7.02$\pm$0.44 \\
          & \cellcolor{lightred}${\mathcal{M}^{\textbf{ablation}}_\textbf{reasoning}}$ & \cellcolor{lightred}0.00$\pm$0.00 & \cellcolor{lightred}0.00$\pm$0.00 \\
          & \cellcolor{lightgreen}$\boldsymbol{\mathcal{M}^{\textbf{restoration}}_\textbf{reasoning}}$ & \cellcolor{lightgreen}\textbf{29.00}$\boldsymbol{\pm}$\textbf{0.44} & \cellcolor{lightgreen}\textbf{6.75}$\boldsymbol{\pm}$\textbf{0.27} \\
          \bottomrule
        \end{tabular}%
      }
    \end{small}
  \end{center}
  \vskip -0.1in
\end{table}

\textbf{Verification Experiments\quad} To validate the point mentioned above and the effectiveness of Equation \ref{approx_W}, we design a controlled experiment with two comparative settings: In the first setting (\textsc{Ablation}), we remove the orthogonal transformation applied to the output subspaces of $W_\text{post}$ (Construction \ref{reduce_Q}), while preserving the transformation on the input subspaces. In the second setting (\textsc{Restoration}), we restore coherence by applying to the output subspaces the same orthogonal transformation derived from the input subspaces (Construction \ref{retain_Q}).
\begin{equation}
  W^{(i)}_\text{post} \leftarrow U^{(i)}_\text{post}\Sigma_\text{post} \cdot \boldsymbol{{V^{(i)}_\text{base}}^T}
  \label{reduce_Q}
\end{equation}
\begin{equation}
\begin{split}
  W^{(i)}_\text{post} \leftarrow U^{(i)}_\text{post}\Sigma_\text{post} \cdot \boldsymbol{({V^{(i)}_\text{base}}\cdot {U^{(i)}_\text{base}}^T{U^{(i)}_\text{post}})^T}
\end{split}
  \label{retain_Q}
\end{equation}
For the original model, the \textsc{ablation} model ($\mathcal{M}^{\textbf{ablation}}_\textbf{post}$) and the \textsc{restoration} model ($\mathcal{M}^{\textbf{restoration}}_\textbf{post}$), all weight matrices in SAs are modified according to Constructions \ref{reduce_Q} and \ref{retain_Q}. The same experimental setup as in Table \ref{replaced_acc} is adopted to evaluate the performance of restoration models on four datasets, with the results reported in Table \ref{restoration_acc}.

The performance of \textsc{ablation} models produce \textbf{nonsensical outputs} across different tasks,leading to \textbf{0\%} accuracy across all evaluation metrics. In contrast, \textsc{restoration} models recover meaningful outputs, further supporting the hypothesis of consistent orthogonal transformations in LLMs. The results across different model scales and families are provided in the Appendix \ref{performance_restoration_models}.

\textbf{Orthogonal Consistency and Model Integrity\quad} To further investigate how consistent orthogonal transformations safeguard the coherence of the semantic space, we analyze the latent representations using \textit{Centered Kernel Alignment} (CKA) \citep{kornblith2019similarityneuralnetworkrepresentations}. This metric allows us to quantify the geometric similarity between the representational structures of different models. 

\begin{figure}[!htbp]
    \vspace{-4pt}
    \centering
    \setlength{\abovecaptionskip}{1pt}
    \includegraphics[width=0.8\linewidth]{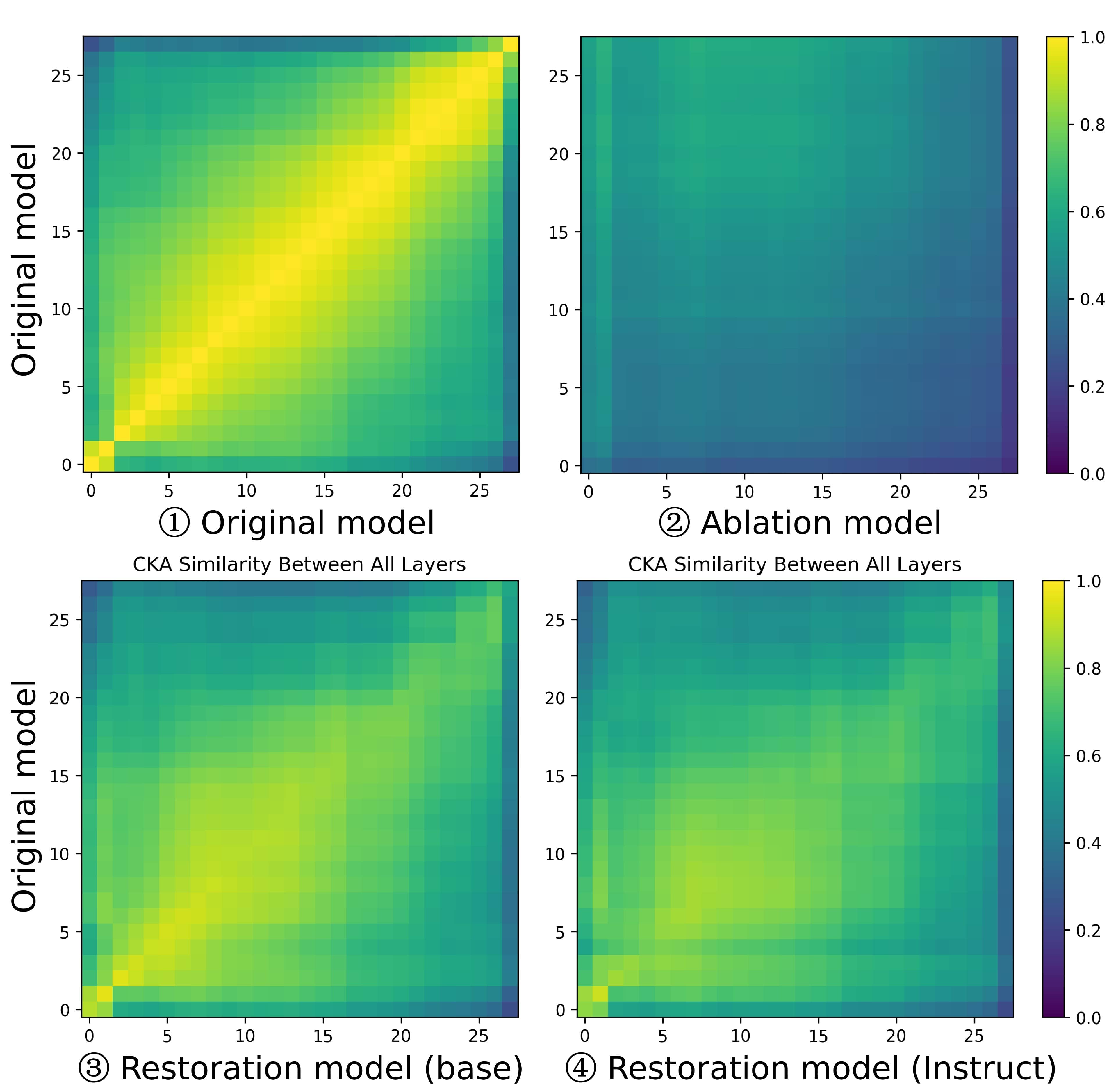}
    \caption{Heatmaps of CKA under different settings. The $\mathcal{M}_{\text{original}}$ in heatmaps is $\mathcal{M}_{\text{reasoning}}$. \ding{173} corresponds to the ablation in Construction~\ref{reduce_Q}, which substantially disrupts the original model’s representational structure. \ding{174} and \ding{175}, corresponding to restorations via Constructions~\ref{retain_Q} and \ref{retain_Q_in}, effectively recover the original hidden representations.}
    \label{cka}
    \vspace{-4pt}
\end{figure}

As shown in Figure~\ref{cka}, the results reveal that the ablation (\ding{173}) leads to an immediate and significant disruption of the representational geometry starting from the very first layer. This suggests that breaking the coupled $Q$ transformation does not merely introduce cumulative errors, but causes a \textbf{fundamental rupture} in the pre-trained feature topology. In contrast, the restoration process (\ding{174}) effectively reinstates the original representational structure, achieving high similarity scores with the \textsc{base} model. This empirical recovery underscores that the consistent $Q$ transformation is the critical mechanism for \textbf{encoding core semantic information}, as it preserves the essential input-output pairings required for meaningful inference. Additional experimental settings and results are provided in Appendix~\ref{cka_restoration_models}.

\textbf{Equivalence of Parameter Impacts Across Post-Training\quad} Beyond internal consistency, we prove that \textsc{post} models trained on divergent distributions remain mutually transformable via these shared orthogonal operators (see Appendix \ref{shared}). As shown by the CKA restoration in Figure \ref{cka} (\ding{175}), the latent space of $\mathcal{M}_{\text{Instruct}}$ can be effectively mapped to that of other $\mathcal{M}_{\text{post}}$ through Construction \ref{retain_Q_in}:
\begin{equation}
  W^{(i)}_\text{post} \leftarrow U^{(i)}_\text{post}\Sigma_\text{post} \cdot \boldsymbol{({V^{(i)}_\text{Instruct}}\cdot {U^{(i)}_\text{Instruct}}^T{U^{(i)}_\text{post}})^T}
  \label{retain_Q_in}
\end{equation}
The $\mathcal{M}_{\text{post}}$ in Construction \ref{retain_Q_in} is $\mathcal{M}_{\text{reasoning}}$. This parametric equivalence offers a novel perspective on \textit{catastrophic forgetting}: \textbf{forgetting occurs when task-specific updates \textbf{overwrite the shared orthogonal backbone}, thereby de-aligning the universal semantic mappings inherited from the pre-trained base.} By situating forgetting within the disruption of these coupled transformations, we provide a concrete geometric basis for understanding how LLMs lose prior capabilities during sequential fine-tuning.

Notably, Appendix \ref{case_study} reveals that the RL-based method also follows the pattern of Equation \ref{approx_W}. This indicates that the improvement in model generalization is essentially driven by \textbf{a more generalized data distribution induced by training methods}, while different training strategies effectively influence the extent to which parameters fit the generalized data. This provides a new understanding of the underlying parameter mechanisms for previous research \cite{lai2026reinforcementfinetuningnaturallymitigates, chu2025sftmemorizesrlgeneralizes,yue2025doesreinforcementlearningreally}.

\section{Conclusion, Discussion, and Limitations}
The paper establishes a unified and interpretable framework for understanding how post-training reshapes the internal structure of large language models. We identify two robust phenomena: a near-uniform geometric scaling of singular values and highly consistent orthogonal transformations of singular vectors. This perspective allows us to first reveal the parametric mechanism by which post-training is bounded by the capacity ceiling of pre-training. Through extensive experiments and theoretical analysis, we uncover a temperature-controlled mechanism for singular value scaling and demonstrate that singular vectors encode the majority of semantic information. Moreover, we prove the equivalence of parameter impacts under different post-training procedures. Understanding how pre-training establishes such a capacity ceiling remains an important direction for future work. We believe this finding may guide additional potential applications (see Appendix \ref{potential_applications} for a related discussion) and provide parameter-space-level guidance for subsequent research.

\newpage
\section*{Impact Statement}

This paper presents work that aims to advance the field of Machine Learning. Our research primarily focuses on the parameter space of LLMs, and neither the subjects of study nor the content involves uncontrolled interactions with the real world. The methodology designed in this paper aims to enhance our understanding of the model’s parameter space and does not entail any foreseeable social or ethical implications.


\bibliography{example_paper}
\bibliographystyle{icml2026}

\newpage
\appendix
\onecolumn
\section{Singular Value Scaling Across Models of different families and sizes}
\label{Visualization_of_other_families_and_scales}
In the main paper, we introduce the SVSMs of \textit{Qwen2.5-Math-1.5B} as the \textsc{base} model. This section continues to present comparisons of models with different post-training methods based on \textsc{base} models \textit{Qwen2.5-Math-7B}, \textit{Llama-3.1-8B}, and \textit{Qwen2.5-14B} in \citet{deepseekai2025deepseekr1incentivizingreasoningcapability}. The different \textsc{post} versions of these models are described in the Appendix \ref{models}. We will also provide a detailed analysis of the cross-layer stability of the near-uniform geometric scaling.
\subsection{SVSMs}
\begin{figure}[!htbp]
    \vspace{-14pt}
    \centering
    \includegraphics[width=0.8\linewidth]{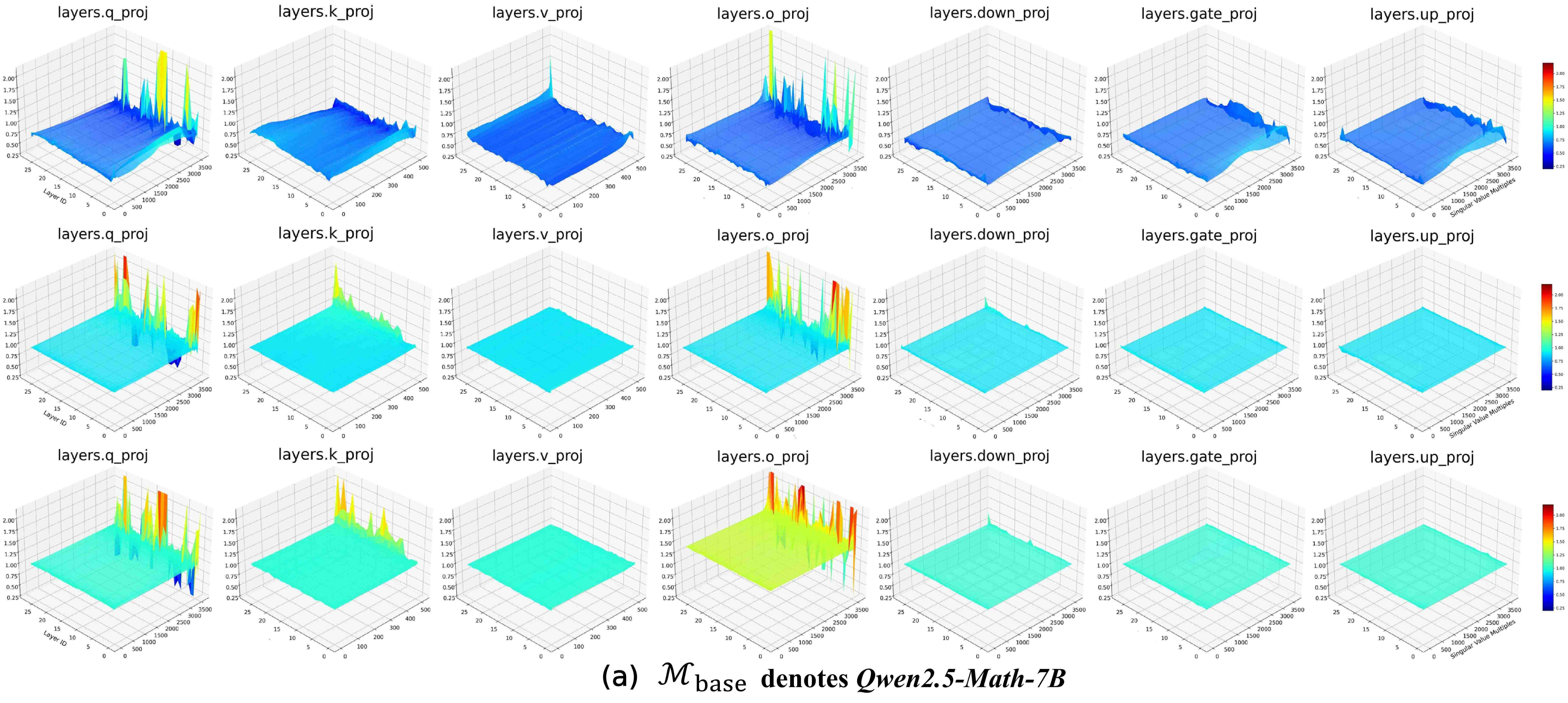}
    \includegraphics[width=0.8\linewidth]{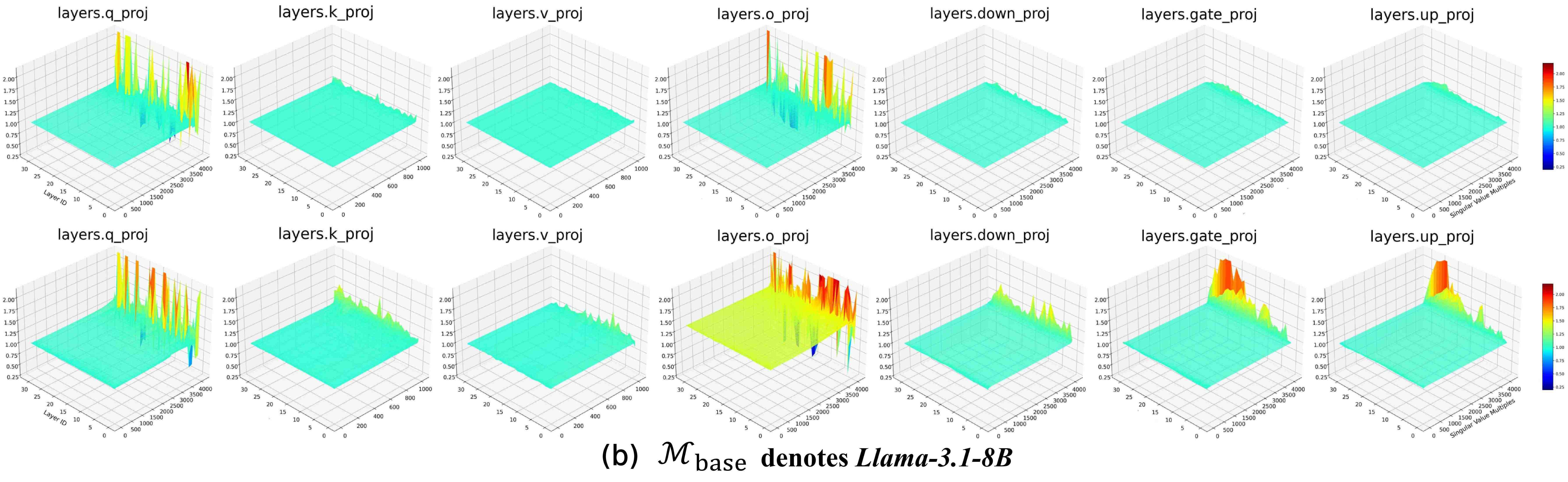}    
    \includegraphics[width=0.8\linewidth]{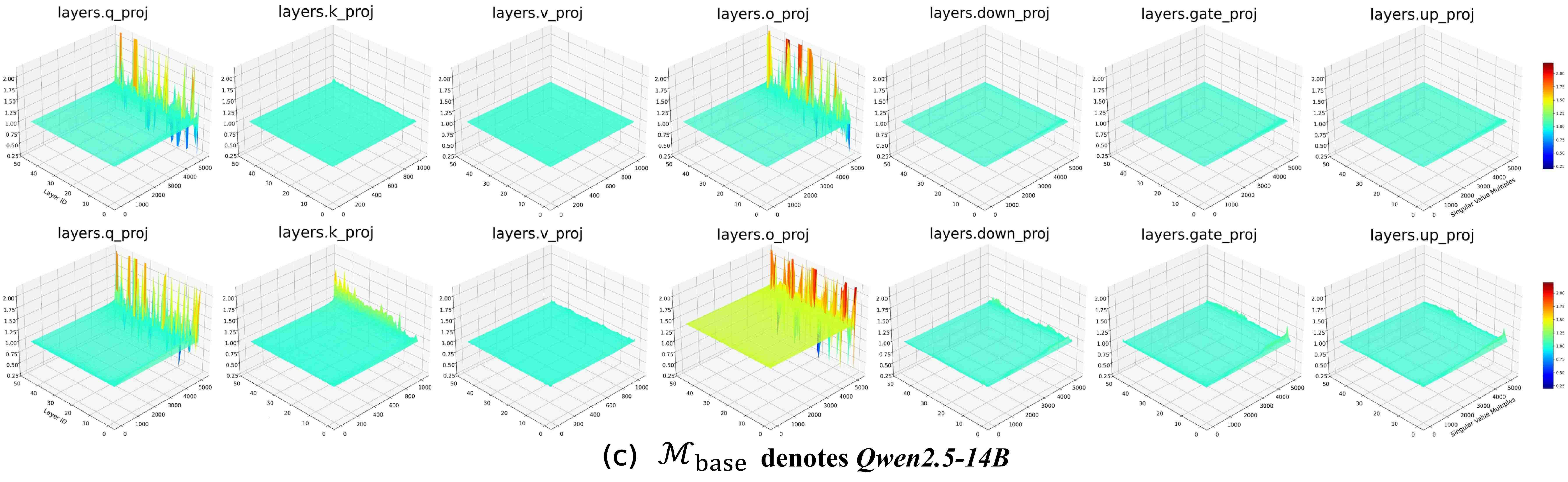} 
    \caption{The heatmaps of SVSMs. The \textsc{base} models of (a), (b) and (c) are \textit{Qwen2.5-Math-7B}, \textit{Llama-3.1-8B} and \textit{Qwen2.5-14B} respectively. Unlike \textit{Qwen2.5-Math-7B} which has different pretrained versions like \textit{Qwen2.5-7B}, only \textsc{instruct} version and \textsc{reasoning} version of the latter two models are compared.}  
    \label{SVSMofBaseAndInstruct_appendix}
    \vspace{-16pt}
\end{figure}

Figure \ref{SVSMofBaseAndInstruct_appendix} shows SVSMs of different \textsc{base} models. We empirically observe a consistent pattern of singular value scaling across different post-training methods, where the principal singular values exhibit identical scaling ratios across different layers. This phenomenon universally manifests in all weight matrices. Notably, the $W_O$ matrices in all \textsc{reasoning} models demonstrate significantly higher overall scaling ratios compared to other weight matrices.

\subsection{Cross-layer Stability of Singular Value Scaling}
\begin{figure}[!htbp]
    \vspace{-8pt}
    \centering
    \includegraphics[width=0.8\linewidth]{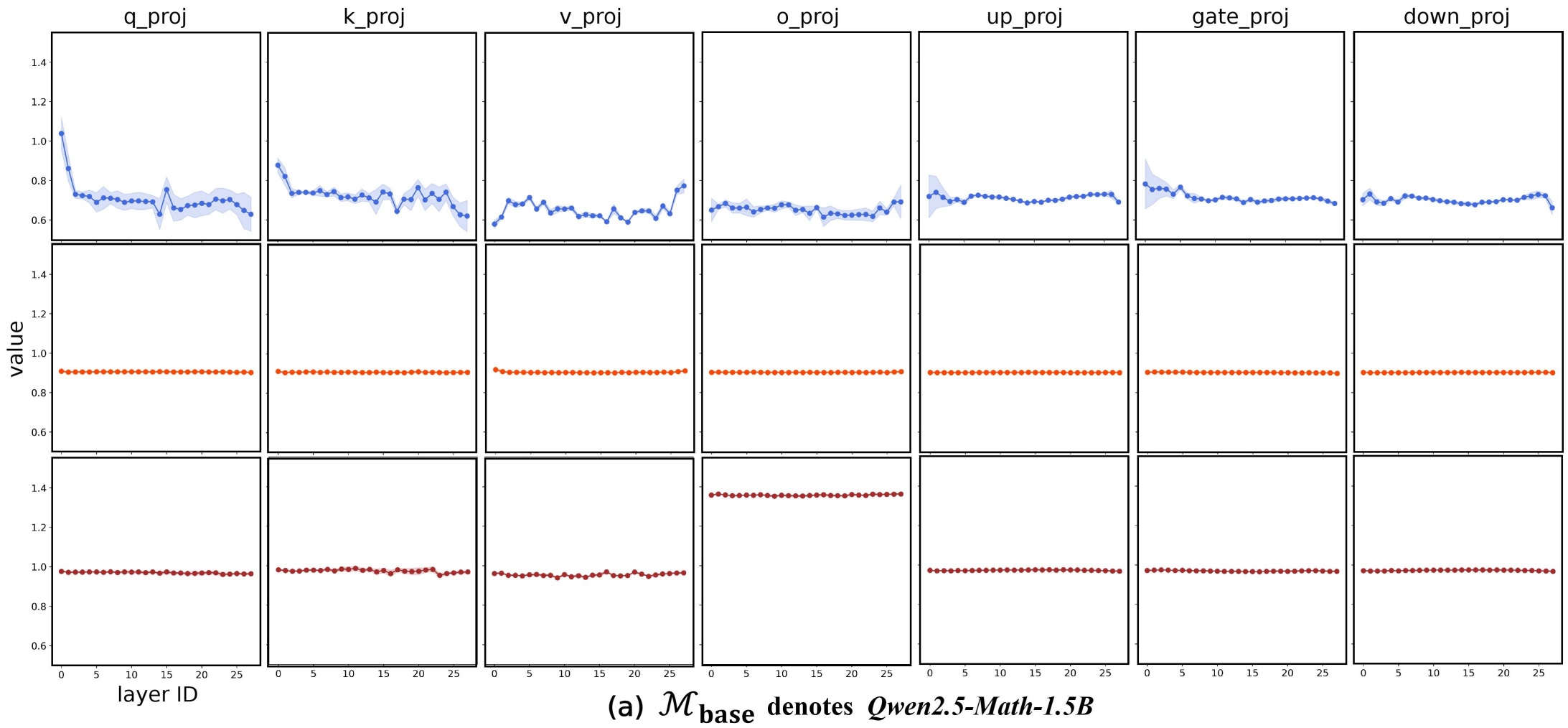}
    \includegraphics[width=0.8\linewidth]{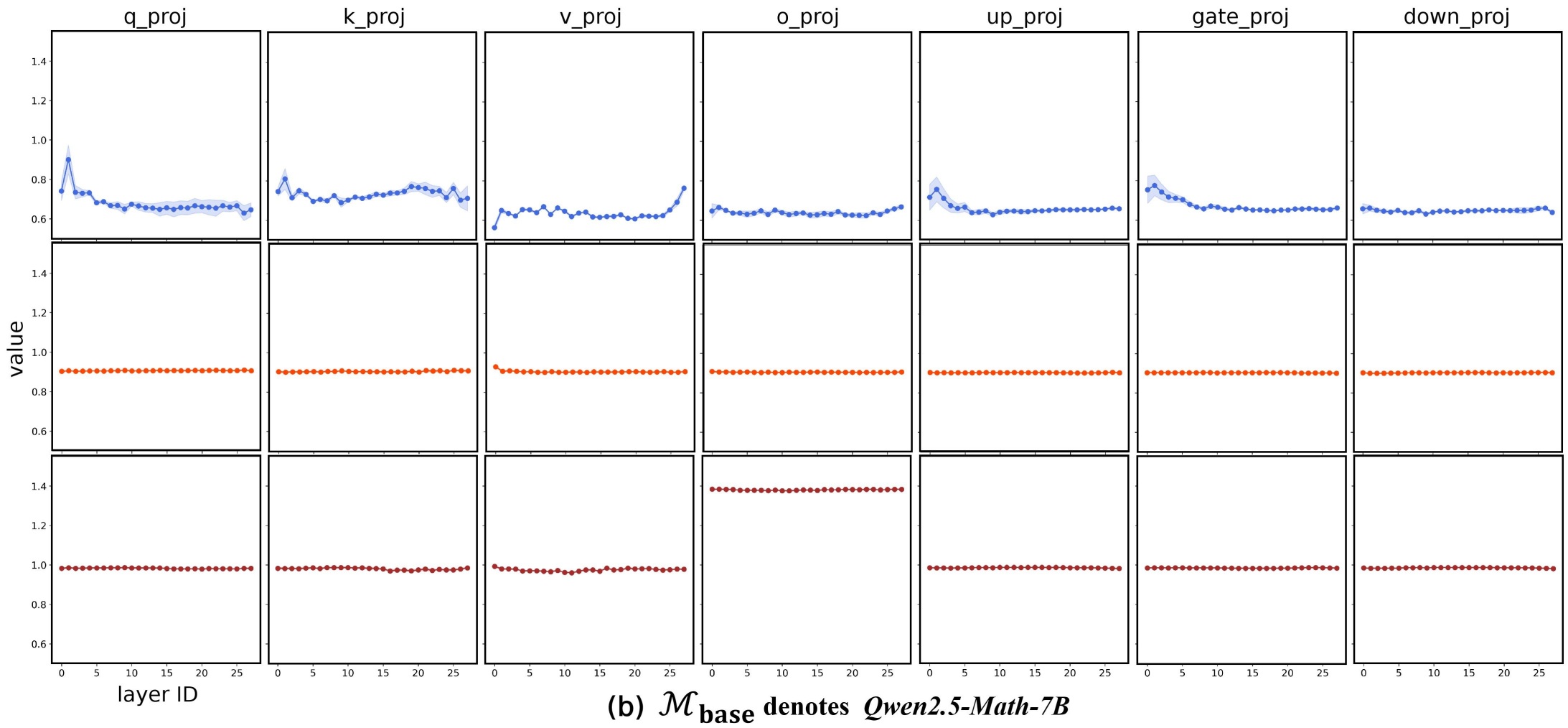}
    \includegraphics[width=0.8\linewidth]{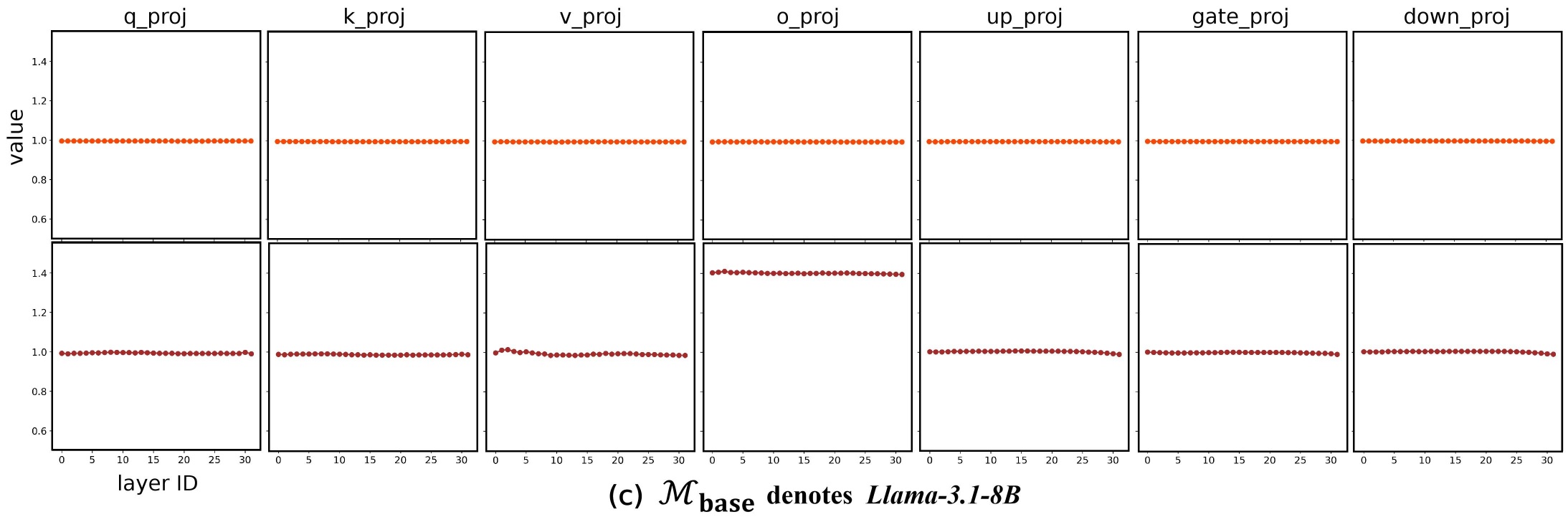}
    \includegraphics[width=0.8\linewidth]{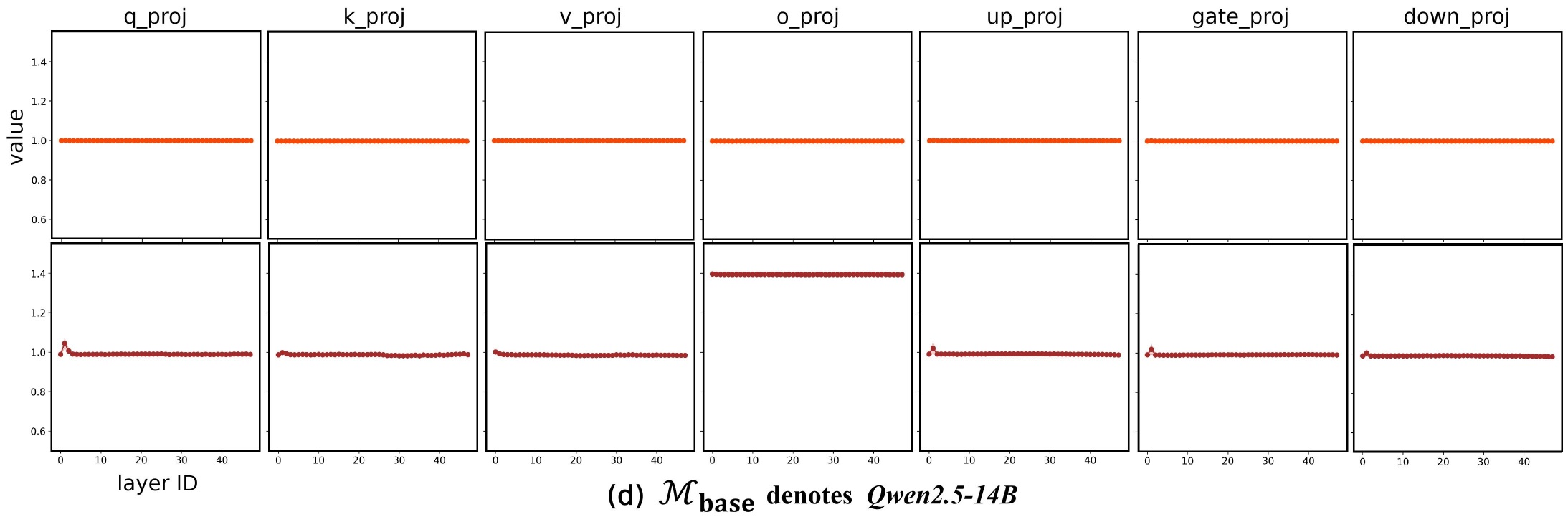}
    \caption{The bandwidth plot shows the distribution ( $mean\pm std$ ) of the scaling factors for the top 90\% singular values in each layer. The blue line indicates comparison with $\mathcal{M}'_\text{base}$, while the light orange and brown curves correspond to comparisons with $\mathcal{M}_\text{instruct}$ and $\mathcal{M}_\text{reasoning}$ respectively. }  
    \label{band}
    \vspace{-8pt}
\end{figure}

Figure \ref{band} shows the mean (dark line) and standard deviation (light band) of the scaling factors for the top 90\% principal singular values across all Transformer blocks. 

As can be seen from the figure, both the \textsc{instruct} and \textsc{reasoning} models show stability in singular value scaling, which is both per-layer (almost no broadband is visible in the \textsc{instruct} and \textsc{reasoning} models) and cross-layer (the values in each layer are almost the same). Table \ref{svsf_table} further reports the overall mean and standard deviation of the scaling factors for the top 90\% singular values across all layers. As shown, the standard deviation across different \textsc{base} models is substantially larger than that between each \textsc{base} model and its corresponding \textsc{post} model (e.g., 37.39× std for  in \textit{Qwen2.5-Math-1.5B} between $\mathcal{M}'_\text{base}$ and $\mathcal{M}_\text{Instruct}$), and the maximum variation of $\mathcal{M}_\text{post}$ remains within 1\%, demonstrating the stability of the singular value scaling phenomenon and further reinforcing our claim.
\begin{table}[!htbp]
  \caption{Global layer statistics of the scaling of the top 90\% singular values ( $mean\pm std$ ), measured for different model families and parameter scales.}
  \label{svsf_table}
  \centering
  \resizebox{0.8\textwidth}{!}{
  \begin{tabular}{llllll}
    \toprule
    \multirow{1}{*}{}        & $SVSM(\frac{\cdot}{\mathcal{M}_\text{base}})$ & $W_Q$& $W_K$& $W_V$& $W_O$\\
    \midrule
    \multirow{3}{*}{\textit{Qwen2.5-Math-1.5B}}& ${\mathcal{M}'_\text{base}}$& $0.6709\pm0.1728$& $0.7017\pm0.0903$& $0.6465\pm0.0432$&$0.6293\pm0.1272$\\ 
                                 & $\mathcal{M}_\text{Intruct}$  & $0.9071\pm0.0046$& $0.9084\pm0.0053$& $0.9026\pm0.0036$& $0.9041\pm0.0036$    \\
                                 & ${\mathcal{M}_\text{reasoning}}$& $0.9710\pm0.0131$& $0.9723\pm0.0109$& $0.9513\pm0.0103$&$1.3551\pm0.0058$\\
    \midrule
    \multirow{3}{*}{\textit{Qwen2.5-Math-7B}}& ${\mathcal{M}'_\text{base}}$& $0.6621\pm0.0827$& $0.7033\pm0.0688$& $0.6388\pm0.0368$&$0.6257\pm0.0317$\\
                                 & $\mathcal{M}_\text{Intruct}$  & $0.9074\pm0.0043$& $0.9103\pm0.0111$& $0.9040\pm0.0047$& $0.9056\pm0.0027$    \\
                                 & ${\mathcal{M}_\text{reasoning}}$ & $0.9837\pm0.0036$& $0.9823\pm0.0072$& $0.9737\pm0.0072$& $1.3800\pm0.0031$    \\

    \midrule
    \multirow{2}{*}{\textit{Llama-3.1-8B}} & $\mathcal{M}_\text{Intruct}$  & $0.9960\pm0.0017$& $0.9951\pm0.0008$& $0.9957\pm0.0009$& $0.9975\pm0.0027$     \\
                                 & ${\mathcal{M}_\text{reasoning}}$ & $1.0041\pm0.0181$& $0.9898\pm0.0058$& $0.9930\pm0.0093$& $1.4112\pm0.0187$\\

    \midrule
    \multirow{2}{*}{\textit{Qwen2.5-14B}} & $\mathcal{M}_\text{Intruct}$  & $0.9990\pm0.0006$& $0.9989\pm0.0003$& $0.9989\pm0.0002$& $0.9989\pm0.0002$     \\
                                 & ${\mathcal{M}_\text{reasoning}}$ & $0.9937\pm0.0142$& $0.9901\pm0.0064$& $0.9861\pm0.0031$& $1.3952\pm0.0017$      \\
    \bottomrule
  \end{tabular}}

    \vspace{1em}
    
  \resizebox{0.68\textwidth}{!}{
  \begin{tabular}{lllll}
    \toprule
    \multirow{1}{*}{}        & $SVSM(\frac{\cdot}{\mathcal{M}_\text{base}})$ & $W_{up}$& $W_{gate}$& $W_{down}$\\
    \midrule
    \multirow{3}{*}{\textit{Qwen2.5-Math-1.5B}} & ${\mathcal{M}'_\text{base}}$& $0.7242\pm0.0882$& $0.7282\pm0.1179$&$0.6967\pm0.0274$\\
                                 & $\mathcal{M}_\text{Intruct}$  & $0.9016\pm0.0010$& $0.9018\pm0.0017$& $0.9019\pm0.0010$    \\
                                 & ${\mathcal{M}_\text{reasoning}}$& $0.9720\pm0.0023$& $0.9687\pm0.0035$& $0.9714\pm0.0026$      \\
    \midrule
    \multirow{3}{*}{\textit{Qwen2.5-Math-7B}} & ${\mathcal{M}'_\text{base}}$& $0.6693\pm0.0454$& $0.6791\pm0.0514$&$0.6495\pm0.0140$\\
                                 & $\mathcal{M}_\text{Intruct}$  & $0.9021\pm0.0014$& $0.9025\pm0.0013$& $0.9024\pm0.0016$    \\
                                 & ${\mathcal{M}_\text{reasoning}}$ & $0.9847\pm0.0020$& $0.9839\pm0.0019$& $0.9843\pm0.0021$   \\

    \midrule
    \multirow{2}{*}{\textit{Llama-3.1-8B}} & $\mathcal{M}_\text{Intruct}$  & $0.9961\pm0.0003$& $0.9957\pm0.0003$& $0.9961\pm0.0003$     \\
                                 & ${\mathcal{M}_\text{reasoning}}$ & $1.0036\pm0.0041$& $0.9988\pm0.0033$& $1.0035\pm0.0044$      \\

    \midrule
    \multirow{2}{*}{\textit{Qwen2.5-14B}} & $\mathcal{M}_\text{Intruct}$  & $0.9991\pm0.0021$& $0.9991\pm0.0015$& $0.9990\pm0.0006$     \\
                                 & ${\mathcal{M}_\text{reasoning}}$ & $0.9922\pm0.0132$& $0.9924\pm0.0119$& $0.9909\pm0.0062$      \\
    \bottomrule
  \end{tabular}}
\end{table}

\section{Consistent Orthogonal Transformations Across Models of different families and sizes}
\label{Visualization_of_other_families_and_scales_orth}
In this section, we compare $\smash{\mathcal{NF}^{(i)}}$ between the \textsc{base} and \textsc{post} versions of \textit{Qwen2.5-Math-7B}, \textit{Llama-3.1-8B}, and \textit{Qwen2.5-14B}. We also visualize the similarity, difference, and orthogonality matrices of the left and right singular vectors of $W_Q$, $W_K$, $W_V$, and $W_O$ (using the first and last Transformer blocks as examples), and discuss whether such orthogonal consistency is already present in the pre-training stage. 
\subsection{Visualizing Orthogonal Consistency Across Models of different families and sizes}
\begin{figure}[!htbp]
    \centering
    \includegraphics[width=0.85\linewidth]{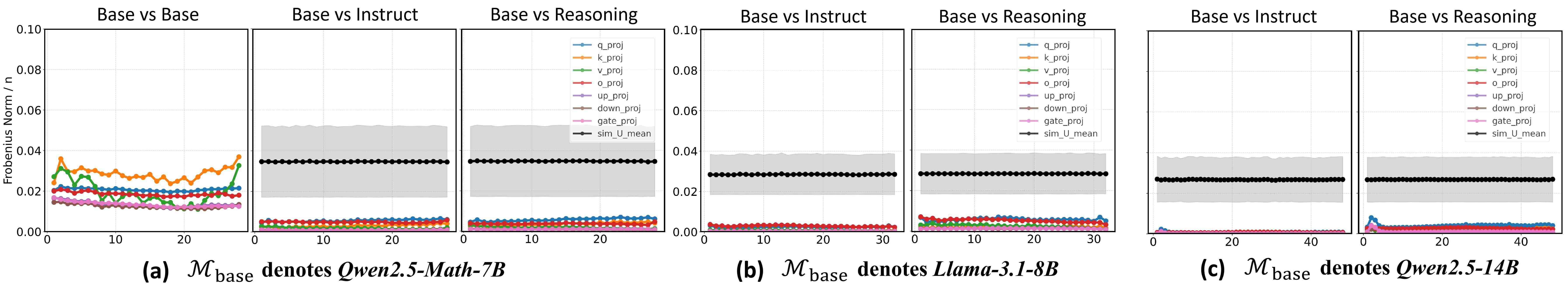}
    \caption{Extensively verifies the equality of $\smash{Q^{(i)}_1}$ and $\smash{Q^{(i)}_2}$ comparing $\mathcal{M}_\text{base}$ to $\mathcal{M}_\text{post}$ by $\smash{\mathcal{NF}^{(i)}}$.}  
    \label{orth_all}
\end{figure}

\begin{figure}[!htbp]
    \centering
    \includegraphics[width=0.8\linewidth]{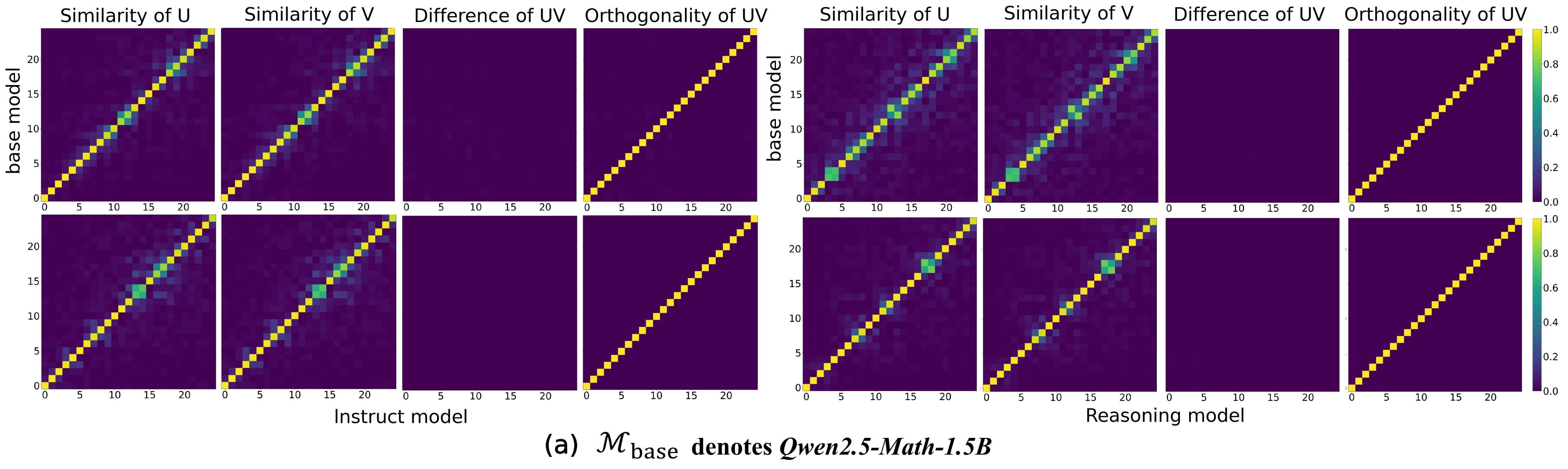}
    \includegraphics[width=0.8\linewidth]{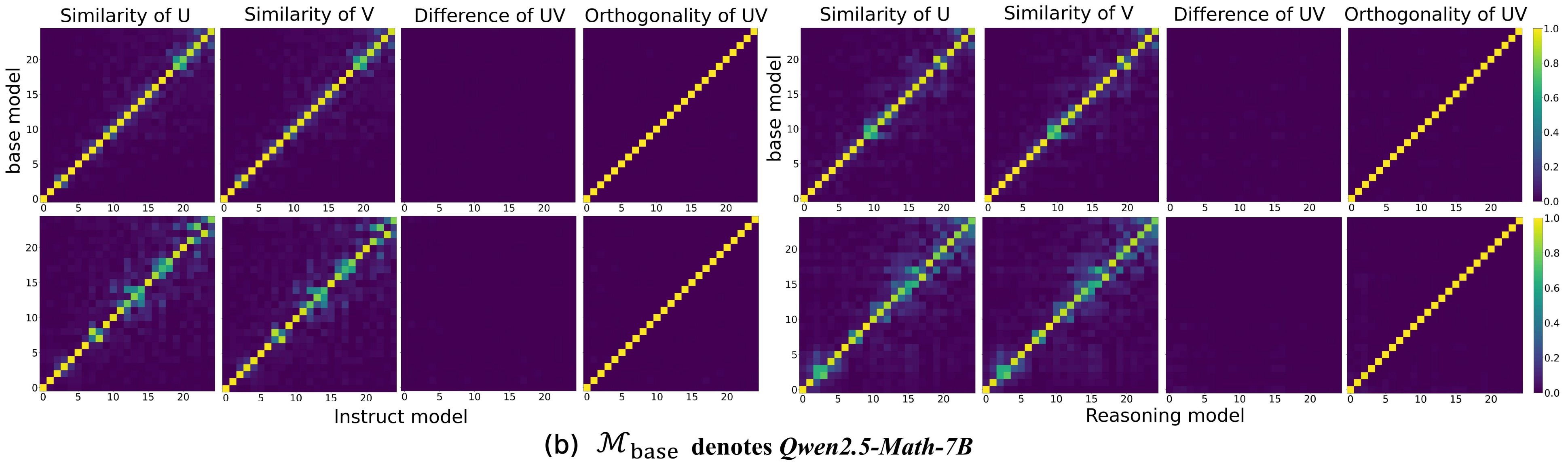}
    \includegraphics[width=0.8\linewidth]{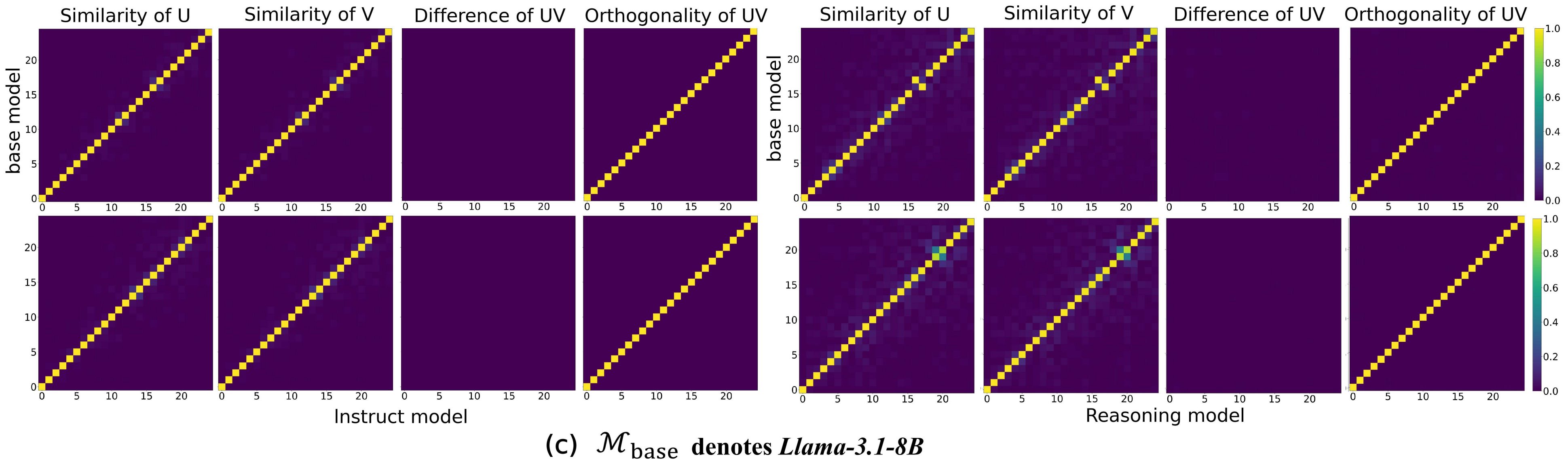}
    \includegraphics[width=0.8\linewidth]{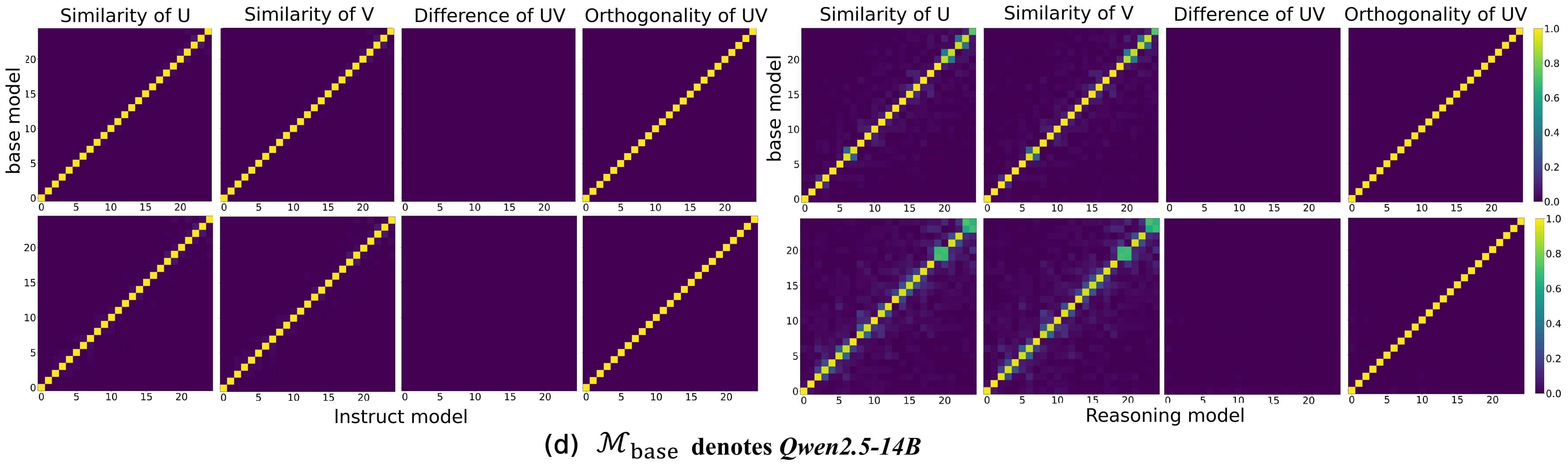}
    \caption{Visualizations of the similarity, difference and orthogonality matrices of the left and right singular vectors of the first and last Transformer block's $W_Q$ before and after post-training across models of different scales.}  
    \label{orth_all_visi_WQ}
\end{figure}

\begin{figure}[!htbp]
    \centering
    \includegraphics[width=0.8\linewidth]{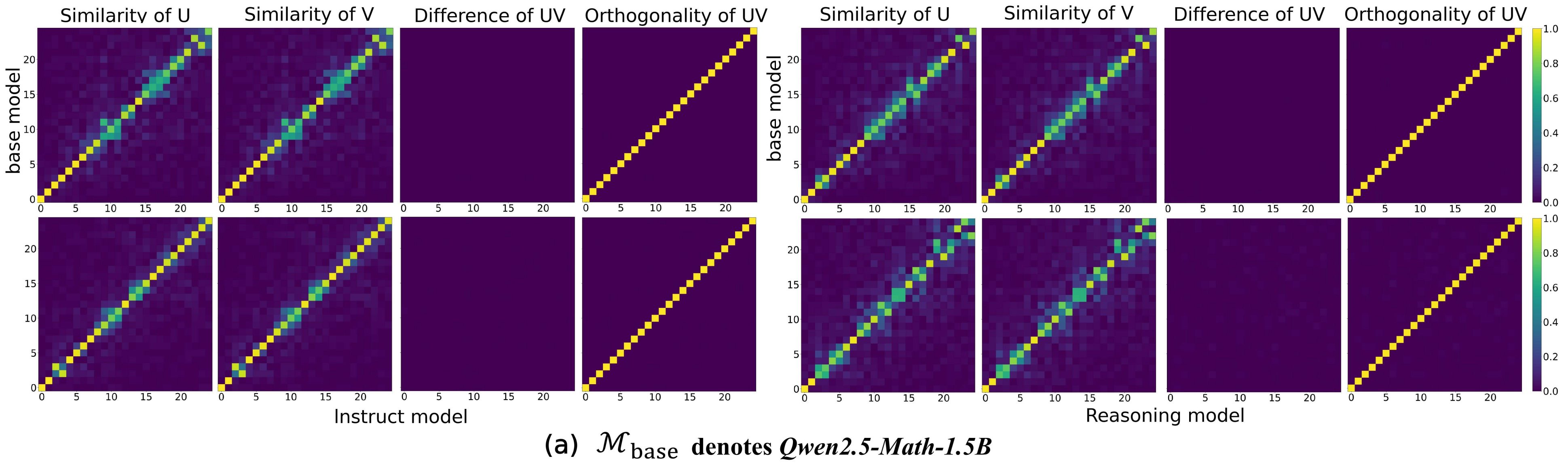}
    \includegraphics[width=0.8\linewidth]{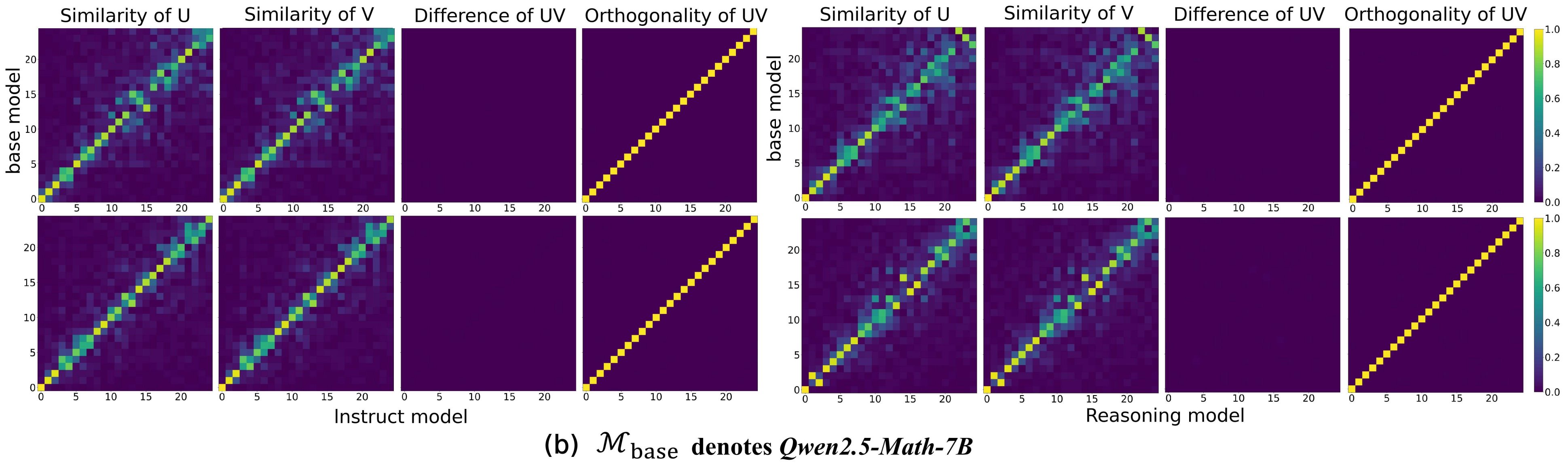}
    \includegraphics[width=0.8\linewidth]{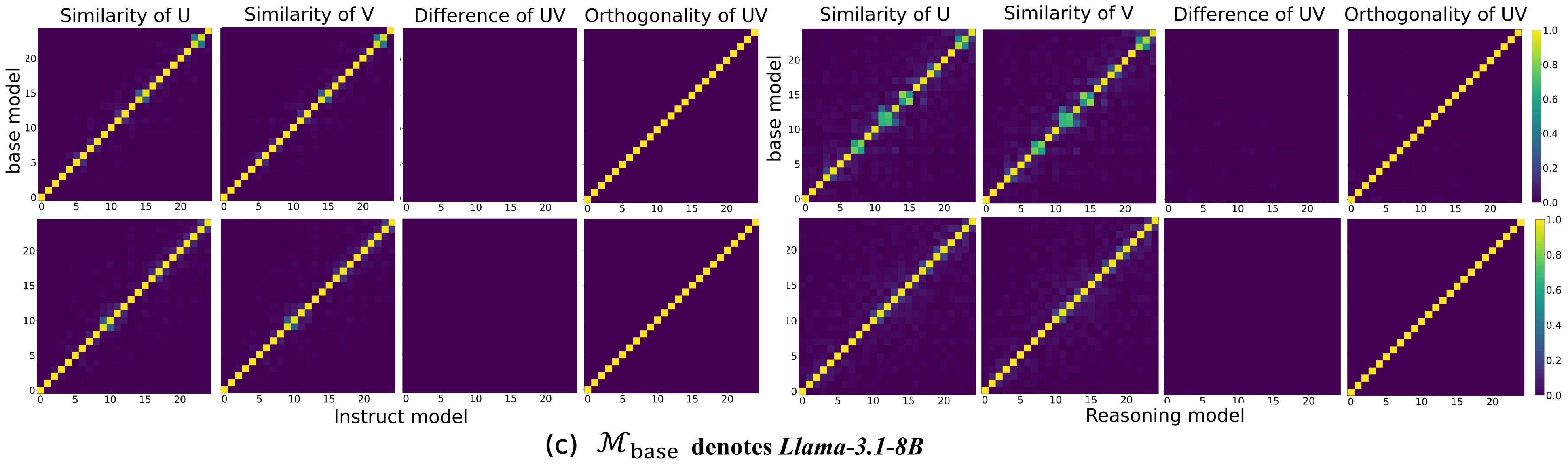}
    \includegraphics[width=0.8\linewidth]{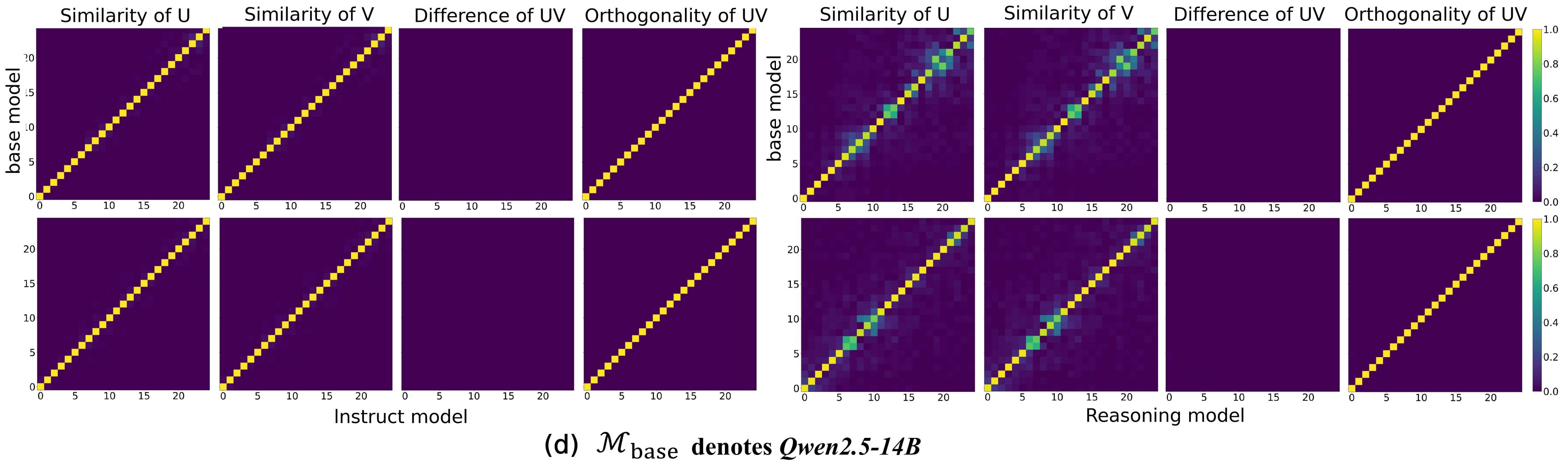}
    \caption{Visualizations of the similarity, difference and orthogonality matrices of the left and right singular vectors of the first and last Transformer block's $W_K$ before and after post-training across models of different scales.}  
    \label{orth_all_visi_WK}
\end{figure}

\begin{figure}[!htbp]
    \centering
    \includegraphics[width=0.8\linewidth]{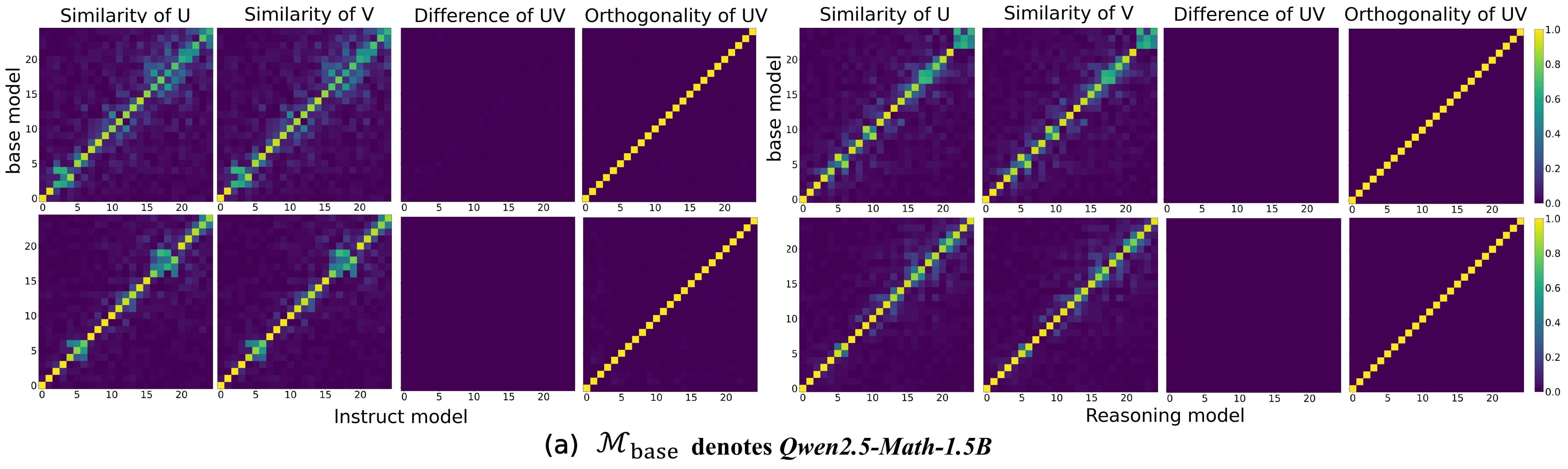}
    \includegraphics[width=0.8\linewidth]{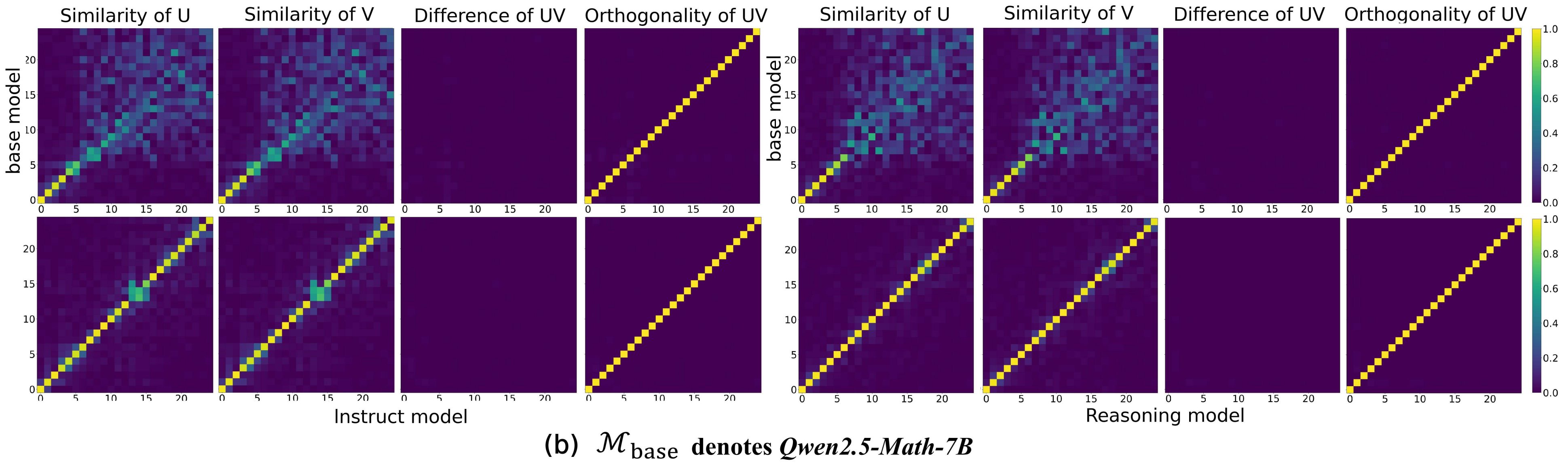}
    \includegraphics[width=0.8\linewidth]{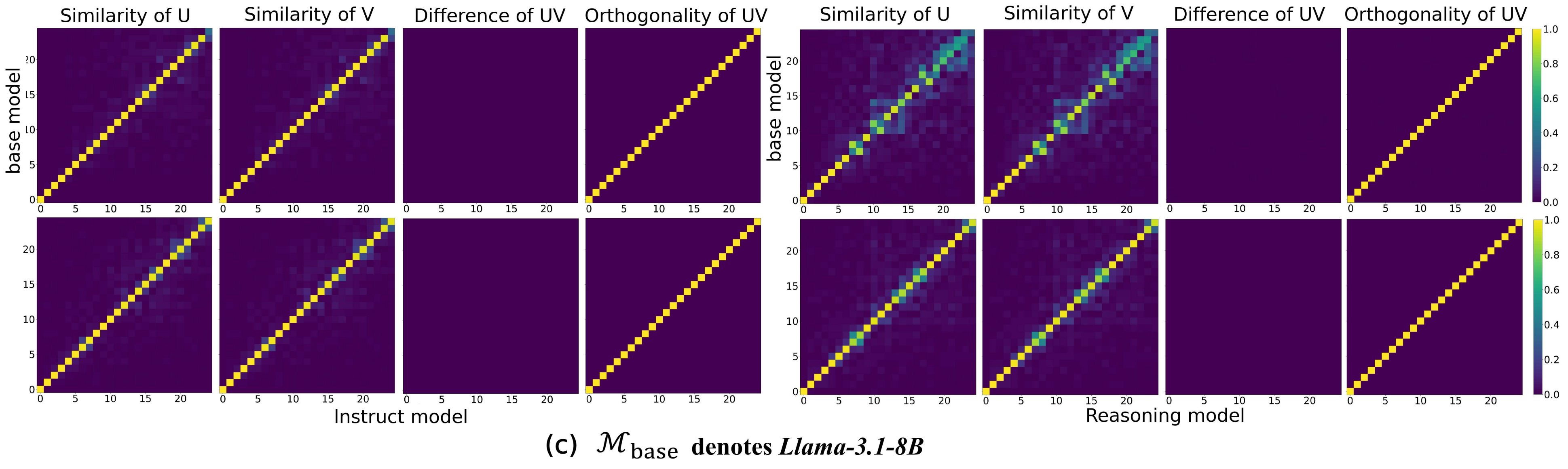}
    \includegraphics[width=0.8\linewidth]{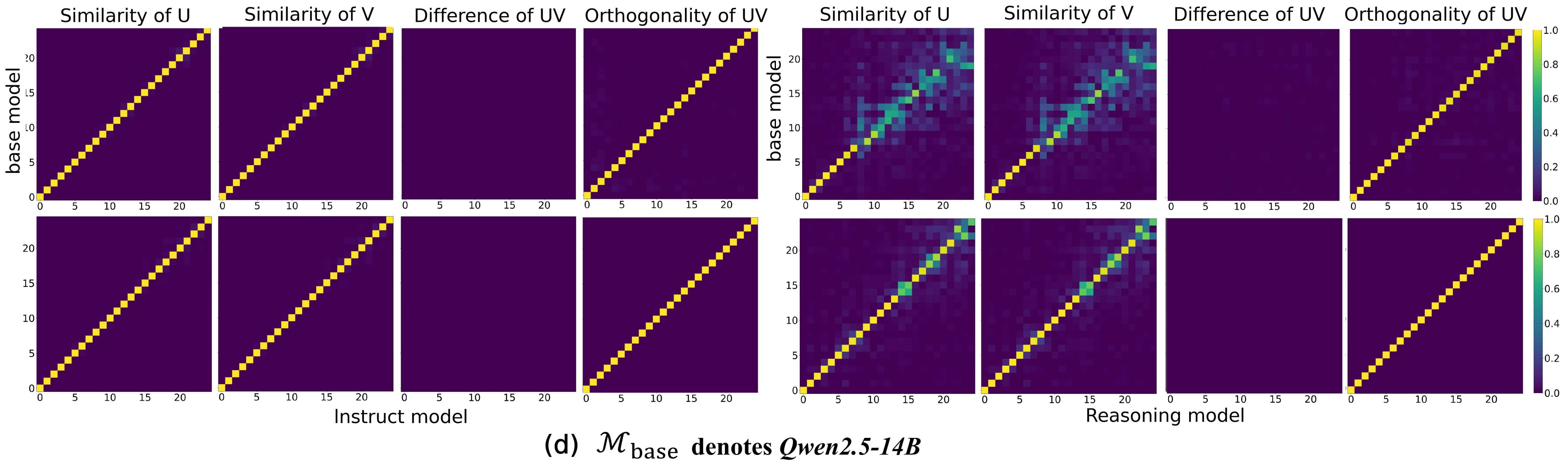}
    \caption{Visualizations of the similarity, difference and orthogonality matrices of the left and right singular vectors of the first and last Transformer block's $W_V$ before and after post-training across models of different scales.}  
    \label{orth_all_visi_WV}
\end{figure}

\begin{figure}[!htbp]
    \centering
    \includegraphics[width=0.8\linewidth]{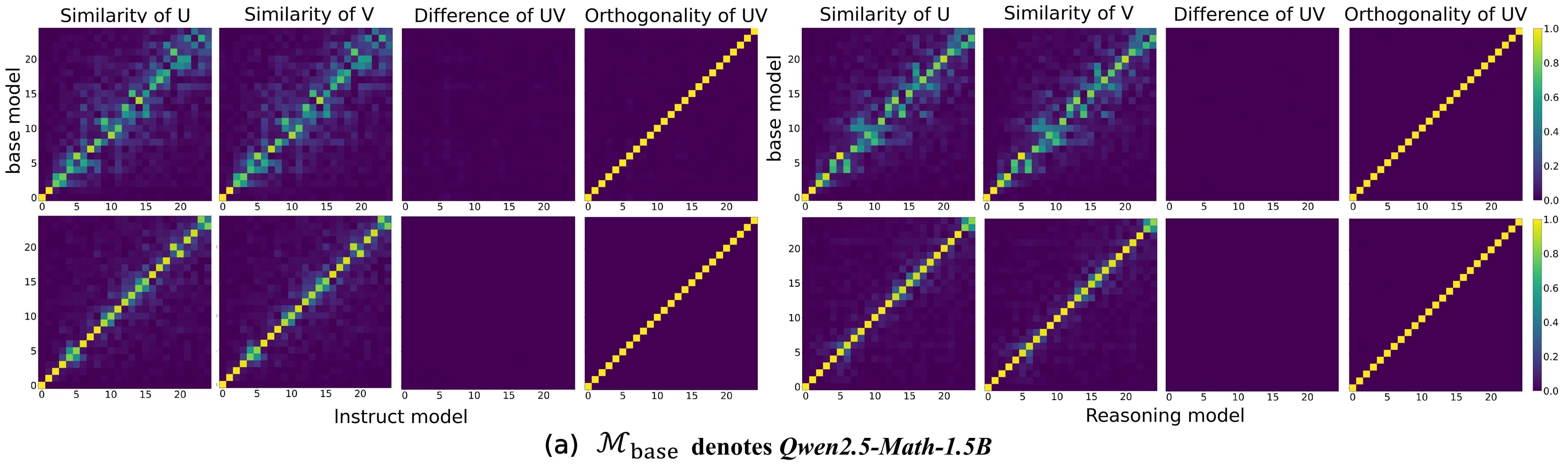}
    \includegraphics[width=0.8\linewidth]{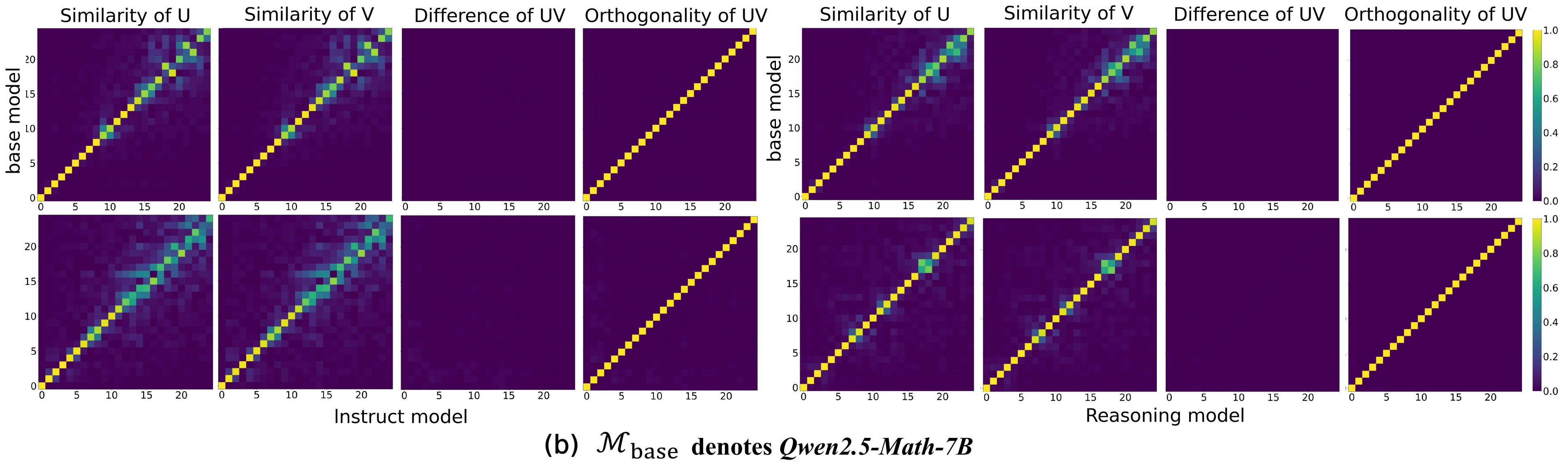}
    \includegraphics[width=0.8\linewidth]{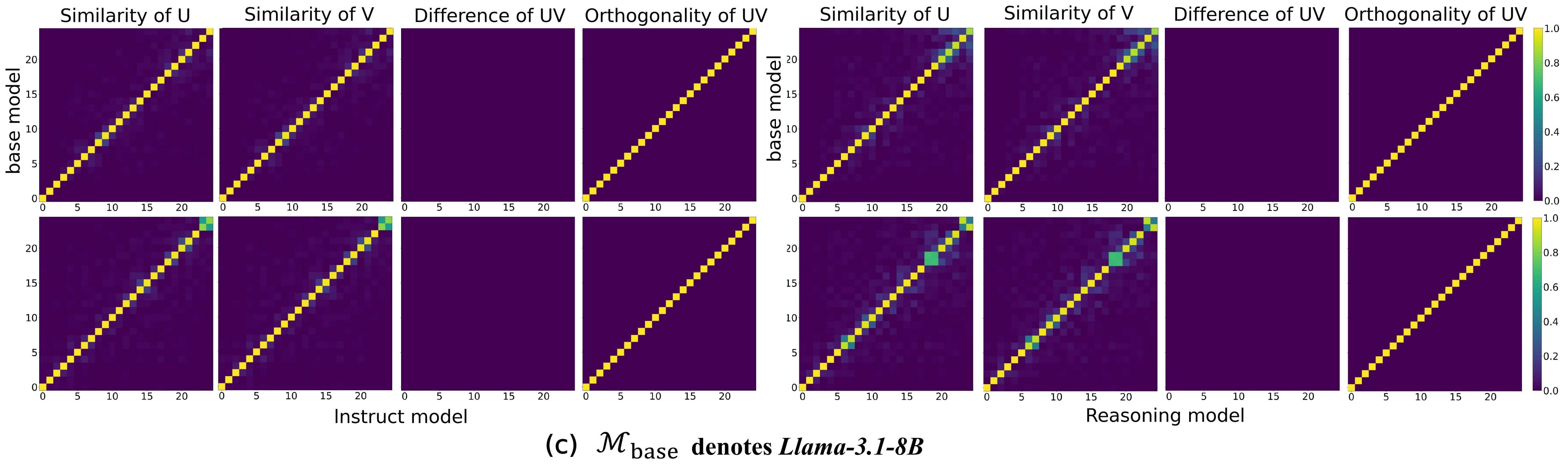}
    \includegraphics[width=0.8\linewidth]{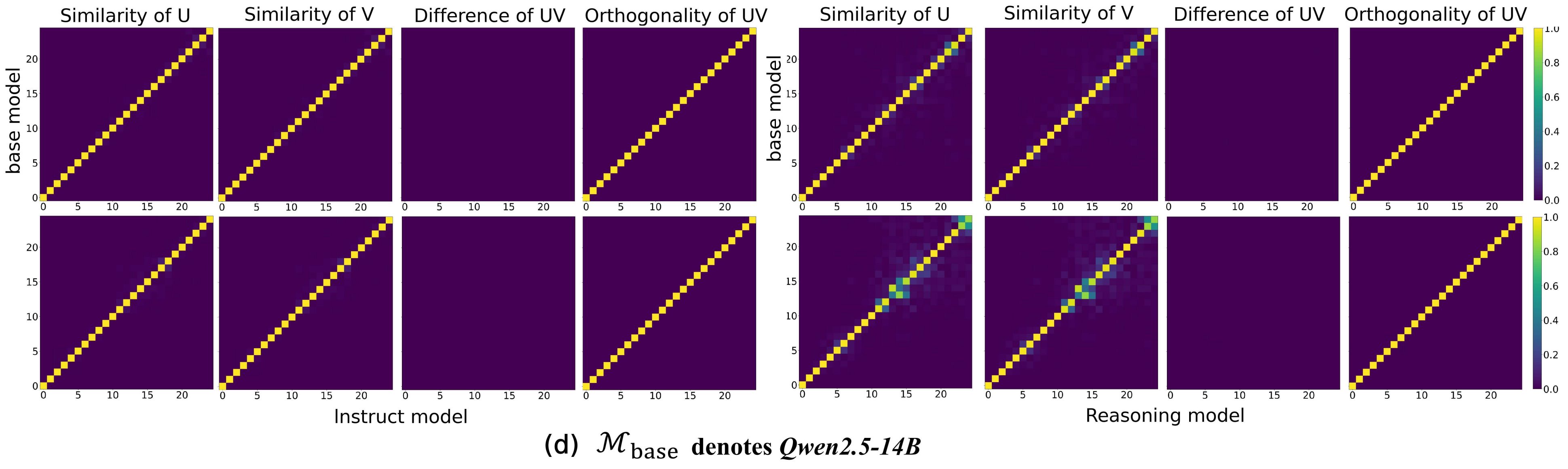}
    \caption{Visualizations of the similarity, difference and orthogonality matrices of the left and right singular vectors of the first and last Transformer block's $W_O$ before and after post-training across models of different scales.}  
    \label{orth_all_visi}
\end{figure}

As shown in Figure \ref{orth_all}, the $\smash{\mathcal{NF}^{(i)}}$ values across different \textsc{post} versions consistently remain low, in contrast to the higher values observed among the pre-training variants (Figure \ref{orth_all}a, \textit{Base vs Base}). This indicates that, despite variations in model scale and post-training methods, each matrix exhibits a high degree of consistency in the orthogonal transformations ($\smash{Q^{(i)}_1}$ and $\smash{Q^{(i)}_2}$) applied to its singular vectors. This phenomenon is illustrated more clearly in Figure \ref{orth_all_visi_WQ}-\ref{orth_all_visi}, where most orthogonality matrices closely approximate the identity matrix.


\newpage
We also observe that the similarity matrices of the left and right singular vectors are mostly concentrated along the diagonal. As shown in Appendix \ref{Visualization_of_other_families_and_scales}, post-training does not alter the distribution of singular values of the weight matrices. When taken together with our current observation, this indirectly supports the view that post-training acts as a perturbation to the pretrained subspaces.

\subsection{Transformations of Singular Vectors After Pre-Training}
\label{Visualization_of_other_families_and_scales_orth_sec2}
\begin{figure}[!htbp]
    \centering
    \includegraphics[width=0.8\linewidth]{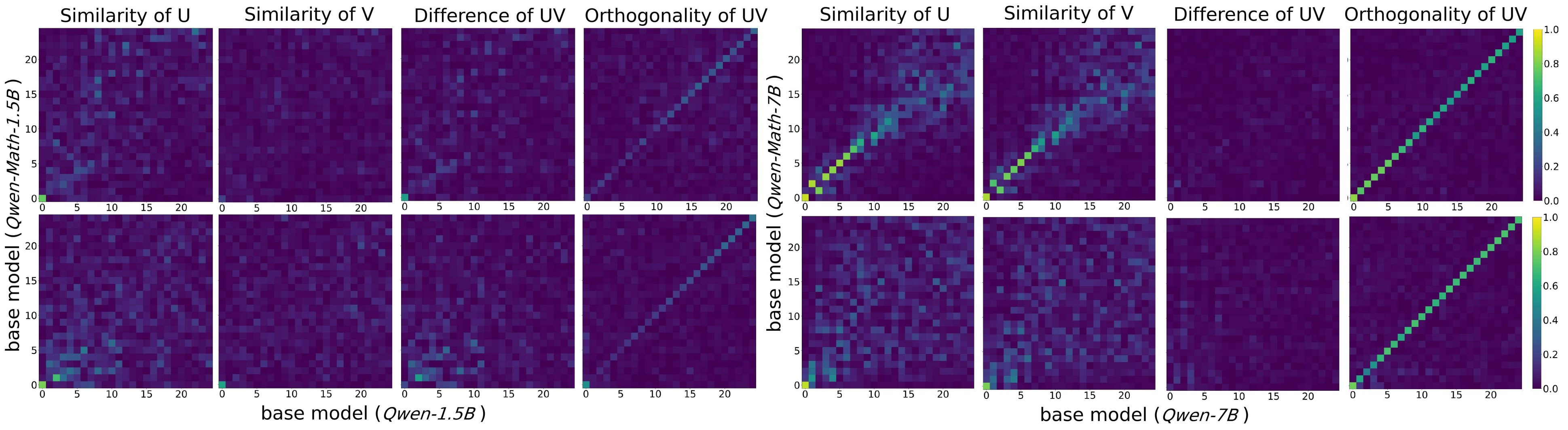}
    \caption{Visualizations of the similarity, difference and orthogonality matrices of the left and right singular vectors of the first and last Transformer block's $W_O$ between $\mathcal{M}_\text{base}$ and $\mathcal{M}'_\text{base}$.}  
    \label{orth_all_base}
\end{figure}

The similarity matrices of the left and right singular vectors across different \textsc{base} models do not exhibit strong diagonal dominance, suggesting substantial divergence in their pretrained subspaces (Figure \ref{orth_all_base}). Despite this divergence, we observe a subtle and consistent pattern in the orthogonal transformations between the left and right singular vectors. This subtle consistency likely indicates that coordinated rotation is a fundamental characteristic during model training, and the failure to adapt to the data distribution leads to the introduction of new exploratory patterns on the original basis, which in turn results in this discrepancy. We can calibrate $U_\text{post},V_\text{post}$ in Equation \ref{approx_W}:
\begin{equation}
    \begin{split}
        U_\text{post}=U_\text{base}(Q\cdot\Delta Q_1)\\
        V_\text{post}=V_\text{base}(Q\cdot\Delta Q_2)
    \end{split}
    \label{calibration}
\end{equation}
The matrices $\Delta Q_1$ and $\Delta Q_2$ represent small-angle components that capture fine-grained deviations superimposed on the coordinated transformation of the left and right singular vectors during training. It is worth noting that $\Delta Q_1$ and $\Delta Q_2$ correspond to the variations' errors, but only one of them is strictly an orthogonal matrix, while the other is merely a projection matrix. These residual transformation correspond to the perturbation term $I_\text{orth}$. From this perspective, the amount of data used in post-training is substantially smaller than in pre-training. As a result, the accumulated perturbations introduced during post-training are also much smaller than the large-scale transformations of the left and right singular vectors induced during pre-training. It is reasonable to postulate that the accumulation of such errors precisely constitutes a significant factor in reshaping the semantic bases of \textsc{base} models. Given that the cumulative deviations introduced by $\Delta Q_1$ and $\Delta Q_2$ remain sufficiently small, the overall transformations of the singular space can be well-approximated as coherent orthogonal rotations. This also supports the validity of the approximation made in Equation \ref{approx_W}.

\section{Experiments on Different Replaced Models}
\label{replaced_appendix}
This section will conduct the same experiments as presented in the main paper on models of varying scales and families, aiming to verify the universality and generalizability of the near-uniform geometric scaling phenomenon of singular values. The evaluation will include tests on four standard benchmark datasets, along with visualizations of attention entropy.

\subsection{Performance of Different Replaced Models}
\label{test_replaced_model}
The purpose of performing Construction \ref{scaling SV with alpha} on $\mathcal{M}_{\text{post}}$ is to verify that the singular value distribution of $\mathcal{M}_{\text{post}}$ can be reconstructed through the linear factor $\alpha'$ and the singular value distribution of $\mathcal{M}_{\text{base}}$, thereby validating the rationality of Equation \ref{approx_W}. This verification critically depends on the selection of $\alpha'$. Our choice of $\alpha'$ is based on Table \ref{svsf_table}, as it reflects the overall distribution of singular value scaling factors. We obtain the final $\alpha'$ values for each type of weight matrix in the \textsc{post} models by rounding the mean of these scaling factors, as presented in Table \ref{alpha'_table}.
\begin{table}[!htbp]
  \caption{$\alpha'$ values (right) assigned based on mean singular value scaling factors (left) of weight matrices per type (from Table \ref{svsf_table}).}
  \label{alpha'_table}
  \centering
  \resizebox{0.78\textwidth}{!}{
  \begin{tabular}{llllll}
    \toprule
    \multirow{1}{*}{}        & \textsc{post} Types & $W_Q$& $W_K$& $W_V$& $W_O$\\
    \midrule
    \multirow{2}{*}{\textit{Qwen2.5-Math-1.5B}} & $\mathcal{M}_\text{Intruct}$  & $0.9071\rightarrow\textbf{0.9}$& $0.9084\rightarrow\textbf{0.9}$& $0.9026\rightarrow\textbf{0.9}$& $0.9041\rightarrow\textbf{0.9}$    \\
                                 & ${\mathcal{M}_\text{reasoning}}$& $0.9710\rightarrow\textbf{1.0}$& $0.9723\rightarrow\textbf{1.0}$& $0.9513\rightarrow\textbf{1.0}$&$1.3551\rightarrow\textbf{1.4}$\\
    \midrule
    \multirow{2}{*}{\textit{Qwen2.5-Math-7B}}& $\mathcal{M}_\text{Intruct}$  & $0.9074\rightarrow\textbf{0.9}$& $0.9103\rightarrow\textbf{0.9}$& $0.9040\rightarrow\textbf{0.9}$& $0.9056\rightarrow\textbf{0.9}$    \\
                                 & ${\mathcal{M}_\text{reasoning}}$ & $0.9837\rightarrow\textbf{1.0}$& $0.9823\rightarrow\textbf{1.0}$& $0.9737\rightarrow\textbf{1.0}$& $1.3800\rightarrow\textbf{1.4}$    \\

    \midrule
    \multirow{2}{*}{\textit{Llama-3.1-8B}} & $\mathcal{M}_\text{Intruct}$  & $0.9960\rightarrow\textbf{1.0}$& $0.9951\rightarrow\textbf{1.0}$& $0.9957\rightarrow\textbf{1.0}$& $0.9975\rightarrow\textbf{1.0}$     \\
                                 & ${\mathcal{M}_\text{reasoning}}$ & $1.0041\rightarrow\textbf{1.0}$& $0.9898\rightarrow\textbf{1.0}$& $0.9930\rightarrow\textbf{1.0}$& $1.4112\rightarrow\textbf{1.4}$\\

    \midrule
    \multirow{2}{*}{\textit{Qwen2.5-14B}} & $\mathcal{M}_\text{Intruct}$  & $0.9990\rightarrow\textbf{1.0}$& $0.9989\rightarrow\textbf{1.0}$& $0.9989\rightarrow\textbf{1.0}$& $0.9989\rightarrow\textbf{1.0}$     \\
                                 & ${\mathcal{M}_\text{reasoning}}$ & $0.9937\rightarrow\textbf{1.0}$& $0.9901\rightarrow\textbf{1.0}$& $0.9861\rightarrow\textbf{1.0}$& $1.3952\rightarrow\textbf{1.4}$      \\
    \bottomrule
  \end{tabular}}

    \vspace{1em}
    
  \resizebox{0.65\textwidth}{!}{
  \begin{tabular}{lllll}
    \toprule
    \multirow{1}{*}{}        & \textsc{post} Types & $W_{up}$& $W_{gate}$& $W_{down}$\\
    \midrule
    \multirow{2}{*}{\textit{Qwen2.5-Math-1.5B}}& $\mathcal{M}_\text{Intruct}$  & $0.9016\rightarrow\textbf{0.9}$& $0.9018\rightarrow\textbf{0.9}$& $0.9019\rightarrow\textbf{0.9}$    \\
                                 & ${\mathcal{M}_\text{reasoning}}$& $0.9720\rightarrow\textbf{1.0}$& $0.9687\rightarrow\textbf{1.0}$& $0.9714\rightarrow\textbf{1.0}$      \\
    \midrule
    \multirow{2}{*}{\textit{Qwen2.5-Math-7B}}& $\mathcal{M}_\text{Intruct}$  & $0.9021\rightarrow\textbf{0.9}$& $0.9025\rightarrow\textbf{0.9}$& $0.9024\rightarrow\textbf{0.9}$    \\
                                 & ${\mathcal{M}_\text{reasoning}}$ & $0.9847\rightarrow\textbf{1.0}$& $0.9839\rightarrow\textbf{1.0}$& $0.9843\rightarrow\textbf{1.0}$   \\

    \midrule
    \multirow{2}{*}{\textit{Llama-3.1-8B}} & $\mathcal{M}_\text{Intruct}$  & $0.9961\rightarrow\textbf{1.0}$& $0.9957\rightarrow\textbf{1.0}$& $0.9961\rightarrow\textbf{1.0}$     \\
                                 & ${\mathcal{M}_\text{reasoning}}$ & $1.0036\rightarrow\textbf{1.0}$& $0.9988\rightarrow\textbf{1.0}$& $1.0035\rightarrow\textbf{1.0}$      \\

    \midrule
    \multirow{2}{*}{\textit{Qwen2.5-14B}} & $\mathcal{M}_\text{Intruct}$  & $0.9991\rightarrow\textbf{1.0}$& $0.9991\rightarrow\textbf{1.0}$& $0.9990\rightarrow\textbf{1.0}$     \\
                                 & ${\mathcal{M}_\text{reasoning}}$ & $0.9922\rightarrow\textbf{1.0}$& $0.9924\rightarrow\textbf{1.0}$& $0.9909\rightarrow\textbf{1.0}$      \\
    \bottomrule
  \end{tabular}}
\end{table}

In our experiments, the output parameters of the LLMs are configured with a temperature of 0.2, a top\_p of 0.95, and a maximum output token limit of 1024. This setting ensures stable generation while maintaining moderate diversity for subsequent statistical analysis. System prompts are provided in Appendix \ref{system_prompt}. Each model is executed three times on the test set, with the final performance reported as the average score and variance. The results are presented in Table \ref{replaced_acc_all}. The mean and variance of the average length of output tokens across three test runs are also reported in Table \ref{replaced_token_length}.

\begin{table}[!htbp]
  \caption{Performance comparison between original and replaced models across GSM8K, MATH-500, MMLU, and GPQA with pass@1 accuracy (\%)}.
  \label{replaced_acc_all}
  \centering
  \renewcommand{\arraystretch}{1.5}
  \resizebox{0.68\textwidth}{!}{
  \begin{tabular}{llllll}
    \toprule
    \textsc{Base} Models & \textsc{Replaced} Types  & GSM8K  & MATH-500 & MMLU (dev) & GPQA \\
    \midrule
    \multirow{4}{*}{\shortstack{\textit{Qwen2.5-}\\\textit{Math-7B}}} 
    & $\mathcal{M}_\text{Instruct}$ & 95.75$\pm$0.12 & 70.06$\pm$0.50 & 55.90$\pm$0.16 & 27.14$\pm$0.49 \\
    & $\boldsymbol{\mathcal{M}^{\textbf{replaced}}_\textbf{Instruct}}$ & \textbf{95.25}$\boldsymbol{\pm}$\textbf{0.06} & \textbf{73.00}$\boldsymbol{\pm}$\textbf{0.43} & \textbf{55.20}$\boldsymbol{\pm}$\textbf{0.16} & \textbf{27.22}$\boldsymbol{\pm}$\textbf{0.41} \\
    & $\mathcal{M}_\text{reasoning}$ & 62.70$\pm$1.05 & 47.60$\pm$0.33 & 58.71$\pm$0.91 & 14.73$\pm$0.97 \\
    & $\boldsymbol{\mathcal{M}^{\textbf{replaced}}_\textbf{reasoning}}$ & \textbf{72.28}$\boldsymbol{\pm}$\textbf{0.42} & \textbf{53.66}$\boldsymbol{\pm}$\textbf{0.81} & \textbf{60.69}$\boldsymbol{\pm}$\textbf{1.03} & \textbf{18.01}$\boldsymbol{\pm}$\textbf{0.87} \\
    \midrule
    \multirow{4}{*}{\textit{Llama-3.1-8B}} 
    & $\mathcal{M}_\text{Instruct}$ & 34.70$\pm$1.24 & 31.46$\pm$1.06 & 67.48$\pm$0.44 & 21.21$\pm$0.29 \\
    & $\boldsymbol{\mathcal{M}^{\textbf{replaced}}_\textbf{Instruct}}$ & \textbf{34.92}$\boldsymbol{\pm}$\textbf{0.37} & \textbf{32.60}$\boldsymbol{\pm}$\textbf{1.14} & \textbf{65.26}$\boldsymbol{\pm}$\textbf{0.57} & \textbf{20.11}$\boldsymbol{\pm}$\textbf{0.76} \\
    & $\mathcal{M}_\text{reasoning}$ & 60.17$\pm$0.07 & 32.73$\pm$0.41 & 52.51$\pm$1.47 & 11.40$\pm$0.17 \\
    & $\boldsymbol{\mathcal{M}^{\textbf{replaced}}_\textbf{reasoning}}$ & \textbf{68.72}$\boldsymbol{\pm}$\textbf{0.43} & \textbf{29.73}$\boldsymbol{\pm}$\textbf{0.90} & \textbf{52.16}$\boldsymbol{\pm}$\textbf{1.29} & \textbf{9.17}$\boldsymbol{\pm}$\textbf{0.51} \\
    \midrule
    \multirow{4}{*}{\textit{Qwen2.5-14B}} 
    & $\mathcal{M}_\text{Instruct}$ & 94.24$\pm$0.29 & 70.53$\pm$0.34 & 90.63$\pm$0.16 & 36.65$\pm$0.36 \\
    & $\boldsymbol{\mathcal{M}^{\textbf{replaced}}_\textbf{Instruct}}$ & \textbf{94.11}$\boldsymbol{\pm}$\textbf{0.25} & \textbf{69.13}$\boldsymbol{\pm}$\textbf{0.09} & \textbf{89.93}$\boldsymbol{\pm}$\textbf{1.01} & \textbf{35.60}$\boldsymbol{\pm}$\textbf{1.48} \\
    & $\mathcal{M}_\text{reasoning}$ & 70.61$\pm$0.46 & 53.13$\pm$0.25 & 77.89$\pm$0.76 & 19.48$\pm$0.55 \\
    & $\boldsymbol{\mathcal{M}^{\textbf{replaced}}_\textbf{reasoning}}$ & \textbf{79.49}$\boldsymbol{\pm}$\textbf{0.42} & \textbf{52.33}$\boldsymbol{\pm}$\textbf{0.25} & \textbf{75.79}$\boldsymbol{\pm}$\textbf{1.03} & \textbf{19.02}$\boldsymbol{\pm}$\textbf{0.32} \\
    \bottomrule
  \end{tabular}}
\end{table}

\begin{table}[!htbp]
  \caption{Comparison of average length of output tokens between Original and Replaced Models across GSM8K, MATH-500, MMLU, and GPQA.}
  \label{replaced_token_length}
  \centering
  \renewcommand{\arraystretch}{1.5}
  \resizebox{0.68\textwidth}{!}{
  \begin{tabular}{llllll}
    \toprule
    \textsc{Base} Models & \textsc{Replaced} Types  & GSM8K  & MATH-500 & MMLU (dev) & GPQA \\
    \midrule
    \multirow{4}{*}{\shortstack{\textit{Qwen2.5-}\\\textit{Math-1.5B}}} 
    & $\mathcal{M}_\text{Instruct}$ & 305.01$\pm$1.54 & 542.32$\pm$1.21 & 402.60$\pm$3.13 & 633.82$\pm$5.09 \\
    & $\boldsymbol{\mathcal{M}^{\textbf{replaced}}_\textbf{Instruct}}$ & \textbf{302.92}$\boldsymbol{\pm}$\textbf{2.54} & \textbf{527.03}$\boldsymbol{\pm}$\textbf{4.11} & \textbf{408.09}$\boldsymbol{\pm}$\textbf{4.31} & \textbf{610.73}$\boldsymbol{\pm}$\textbf{8.94} \\
    & $\mathcal{M}_\text{reasoning}$ & 539.82$\pm$6.86 & 911.55$\pm$5.55 & 619.34$\pm$13.82 & 952.00$\pm$18.83 \\
    & $\boldsymbol{\mathcal{M}^{\textbf{replaced}}_\textbf{reasoning}}$ & \textbf{427.41}$\boldsymbol{\pm}$\textbf{5.33} & \textbf{864.71}$\boldsymbol{\pm}$\textbf{8.03} & \textbf{590.98}$\boldsymbol{\pm}$\textbf{15.42} & \textbf{939.18}$\boldsymbol{\pm}$\textbf{9.91} \\
    \midrule
    \multirow{4}{*}{\shortstack{\textit{Qwen2.5-}\\\textit{Math-7B}}} 
    & $\mathcal{M}_\text{Instruct}$ & 299.46$\pm$3.17 & 551.34$\pm$4.39 & 372.53$\pm$5.91 & 567.34$\pm$4.96 \\
    & $\boldsymbol{\mathcal{M}^{\textbf{replaced}}_\textbf{Instruct}}$ & \textbf{304.21}$\boldsymbol{\pm}$\textbf{2.91} & \textbf{549.13}$\boldsymbol{\pm}$\textbf{2.53} & \textbf{378.34}$\boldsymbol{\pm}$\textbf{4.51} & \textbf{533.19}$\boldsymbol{\pm}$\textbf{5.98} \\
    & $\mathcal{M}_\text{reasoning}$ & 729.16$\pm$7.64 & 795.40$\pm$9.01 & 514.15$\pm$6.91 & 933.15$\pm$9.97 \\
    & $\boldsymbol{\mathcal{M}^{\textbf{replaced}}_\textbf{reasoning}}$ & \textbf{451.27}$\boldsymbol{\pm}$\textbf{9.28} & \textbf{726.08}$\boldsymbol{\pm}$\textbf{6.14} & \textbf{488.30}$\boldsymbol{\pm}$\textbf{15.17} & \textbf{891.63}$\boldsymbol{\pm}$\textbf{6.07} \\
    \midrule
    \multirow{4}{*}{\textit{Llama-3.1-8B}} 
    & $\mathcal{M}_\text{Instruct}$ & 166.47$\pm$4.22 & 359.19$\pm$6.02 & 35.79$\pm$1.43 & 236.35$\pm$7.38 \\
    & $\boldsymbol{\mathcal{M}^{\textbf{replaced}}_\textbf{Instruct}}$ & \textbf{146.05}$\boldsymbol{\pm}$\textbf{2.18} & \textbf{451.38}$\boldsymbol{\pm}$\textbf{7.71} & \textbf{41.42}$\boldsymbol{\pm}$\textbf{3.36} & \textbf{251.64}$\boldsymbol{\pm}$\textbf{3.06} \\
    & $\mathcal{M}_\text{reasoning}$ & 627.14$\pm$8.71 & 931.14$\pm$14.80 & 721.64$\pm$11.13 & 989.41$\pm$7.43 \\
    & $\boldsymbol{\mathcal{M}^{\textbf{replaced}}_\textbf{reasoning}}$ & \textbf{651.23}$\boldsymbol{\pm}$\textbf{11.34} & \textbf{970.02}$\boldsymbol{\pm}$\textbf{15.14} & \textbf{751.02}$\boldsymbol{\pm}$\textbf{8.29} & \textbf{994.00}$\boldsymbol{\pm}$\textbf{4.31} \\
    \midrule
    \multirow{4}{*}{\textit{Qwen2.5-14B}} 
    & $\mathcal{M}_\text{Instruct}$ & 281.95$\pm$7.21 & 550.02$\pm$6.17 & 89.69$\pm$1.18 & 240.16$\pm$6.55 \\
    & $\boldsymbol{\mathcal{M}^{\textbf{replaced}}_\textbf{Instruct}}$ & \textbf{299.14}$\boldsymbol{\pm}$\textbf{5.11} & \textbf{530.65}$\boldsymbol{\pm}$\textbf{5.93} & \textbf{87.56}$\boldsymbol{\pm}$\textbf{2.43} & \textbf{241.67}$\boldsymbol{\pm}$\textbf{6.39} \\
    & $\mathcal{M}_\text{reasoning}$ & 583.01$\pm$4.57 & 897.61$\pm$8.81 & 487.54$\pm$7.68 & 924.63$\pm$7.90 \\
    & $\boldsymbol{\mathcal{M}^{\textbf{replaced}}_\textbf{reasoning}}$ & \textbf{410.97}$\boldsymbol{\pm}$\textbf{7.81} & \textbf{847.14}$\boldsymbol{\pm}$\textbf{2.06} & \textbf{514.09}$\boldsymbol{\pm}$\textbf{6.90} & \textbf{933.15}$\boldsymbol{\pm}$\textbf{5.10} \\
    \bottomrule
  \end{tabular}}
\end{table}

Experimental results demonstrate that models exhibit nearly identical performance before and after singular value replacement. This further validates that post-training does not alter the singular value distribution of pre-trained models, thereby supporting our conclusion.

We also observe that the performance of some \textsc{reasoning} models improves after singular value replacement. One possible explanation is that Construction \ref{scaling SV with alpha} effectively eliminates noise arising from precision limitations or heterogeneous data during singular value adjustment of $\mathcal{M}_\textbf{base}$s' weight matrices in post-training phases. This reduction in noise consequently enables more efficient token consumption for simpler tasks (e.g., the notable decrease in output token count for $\mathcal{M}^{\textbf{replaced}}_\textbf{reasoning}$ of \textit{Qwen2.5-Math-7B} on GSM8K). These observations suggest that post-training processes exert theoretically derivable influences on the singular values of weight matrices. We identify this phenomenon as a crucial direction for future theoretical investigation.

\subsection{Attention Entropy of Different Replaced Models}
\label{Attention_Behavior}
To demonstrate that singular value scaling is similar to a temperature-controlled mechanism, we perform the following operation on all weight matrices $W_\text{post}$ of the \textsc{post} models:
\begin{equation}
    W_\text{post} \leftarrow U_\text{post}\Sigma_\text{base}V^T_\text{post}
    \label{temperature-controlled}
\end{equation}
Construction \ref{temperature-controlled} replaces the singular values of \textsc{post} models’ weight matrices with those from \textsc{base} models. To evaluate the impact of this substitution, we monitor the attention entropy $\mathcal{H}$. A substantial change in entropy suggests a shift in the distribution of attention scores, indicating a structural change. Otherwise, the effect may be interpreted as a soft temperature modulation. 

We input example questions from different domains \cite{cobbe2021trainingverifierssolvemath,talmor2019commonsenseqaquestionansweringchallenge,mmlu,rein2023gpqagraduatelevelgoogleproofqa} into replaced models $\mathcal{M}_\text{replaced}$ and observe their attention scores prior to generating the first token. Specifically, we track the average attention distribution from each attention head in Transformer blocks 0, 3, 5, 8, 10, 13, 15, 18, 20, 23, and 25, and compute the corresponding attention entropy. 

\begin{figure}[!htbp]
    \centering
    \includegraphics[width=0.8\linewidth]{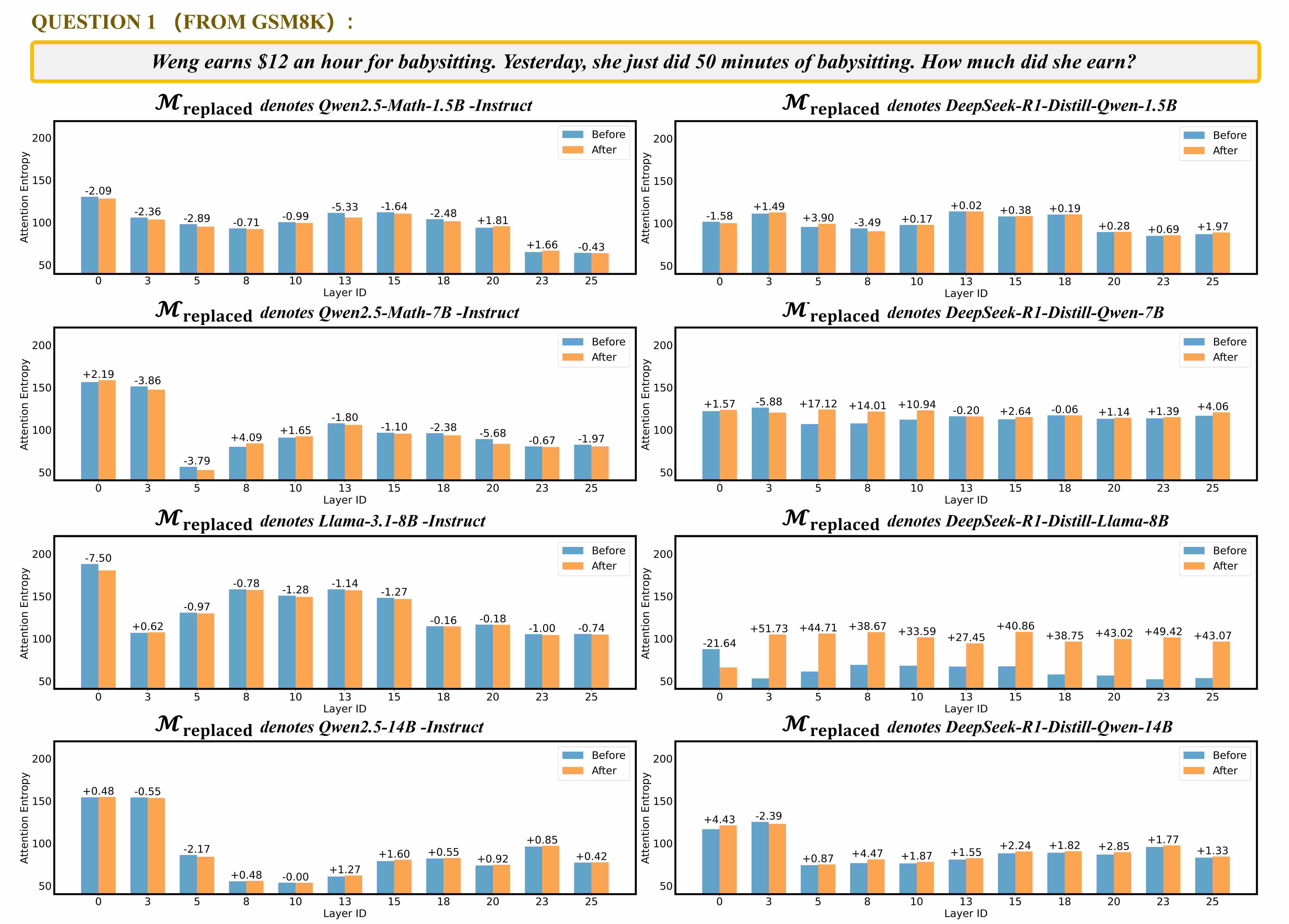}
    \caption{Attention entropy for different $\mathcal{M}_\text{replaced}$. The example input is from GSM8K.}  
    \label{attn_entropy_01}
\end{figure}
\begin{figure}[!htbp]
    \centering
    \includegraphics[width=0.8\linewidth]{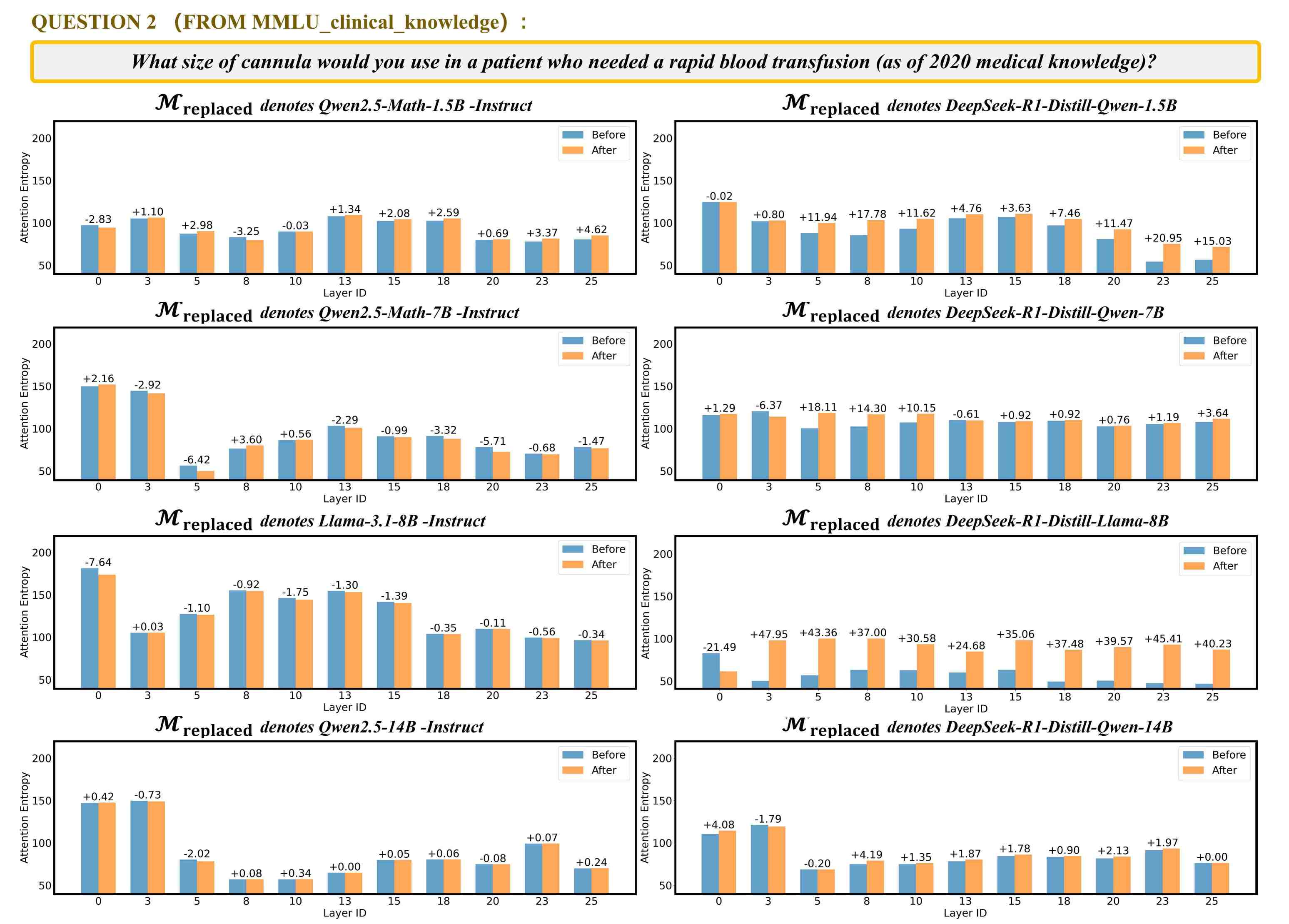}
    \caption{Attention entropy for different $\mathcal{M}_\text{replaced}$. The example input is from MMLU (clinical knowledge).}  
    \label{attn_entropy_02}
\end{figure}
\begin{figure}[!htbp]
    \centering
    \includegraphics[width=0.8\linewidth]{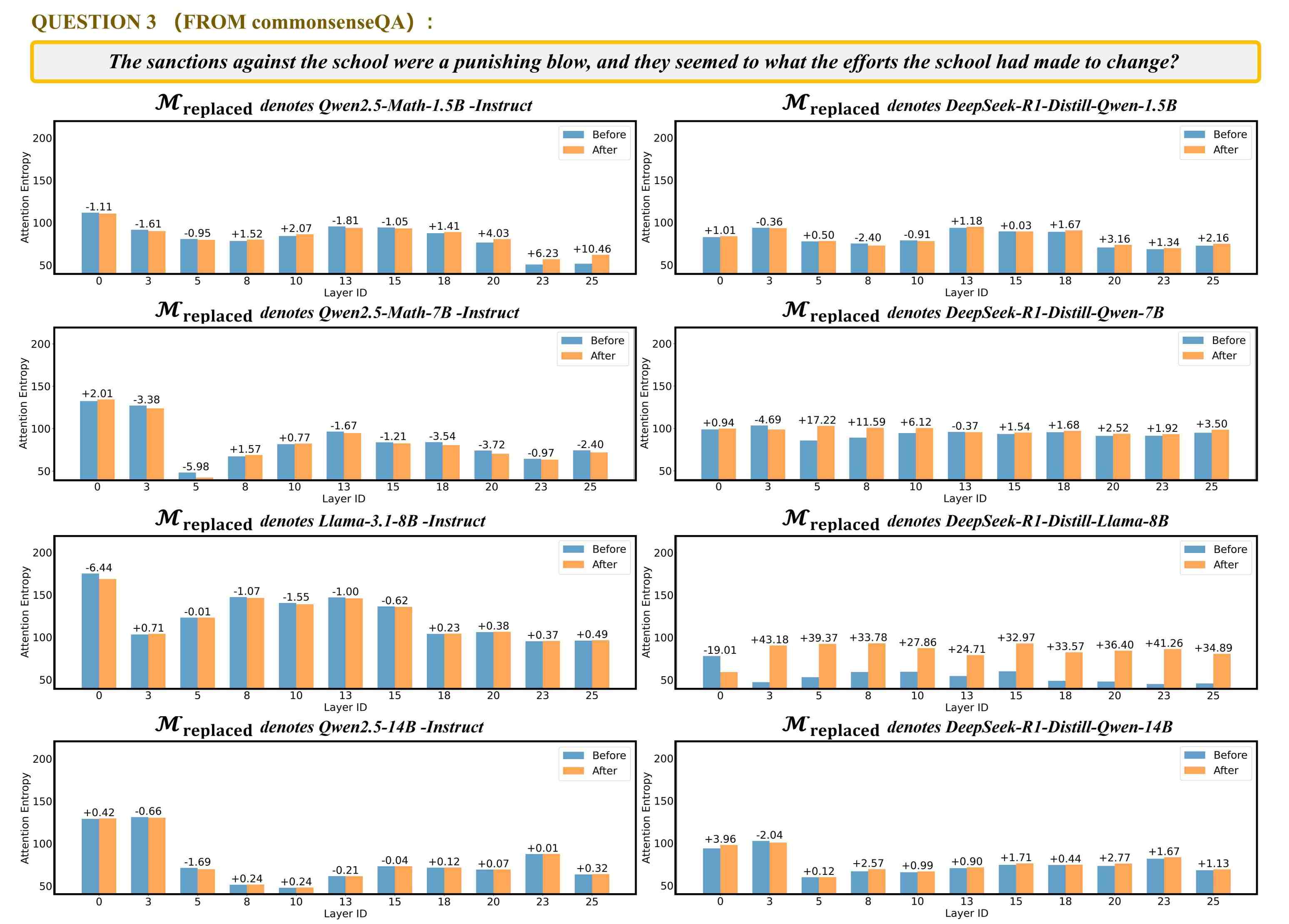}
    \caption{Attention entropy for different $\mathcal{M}_\text{replaced}$. The example input is from CommonsenseQA.}  
    \label{attn_entropy_03}
\end{figure}
\begin{figure}[!htbp]
    \centering
    \includegraphics[width=0.8\linewidth]{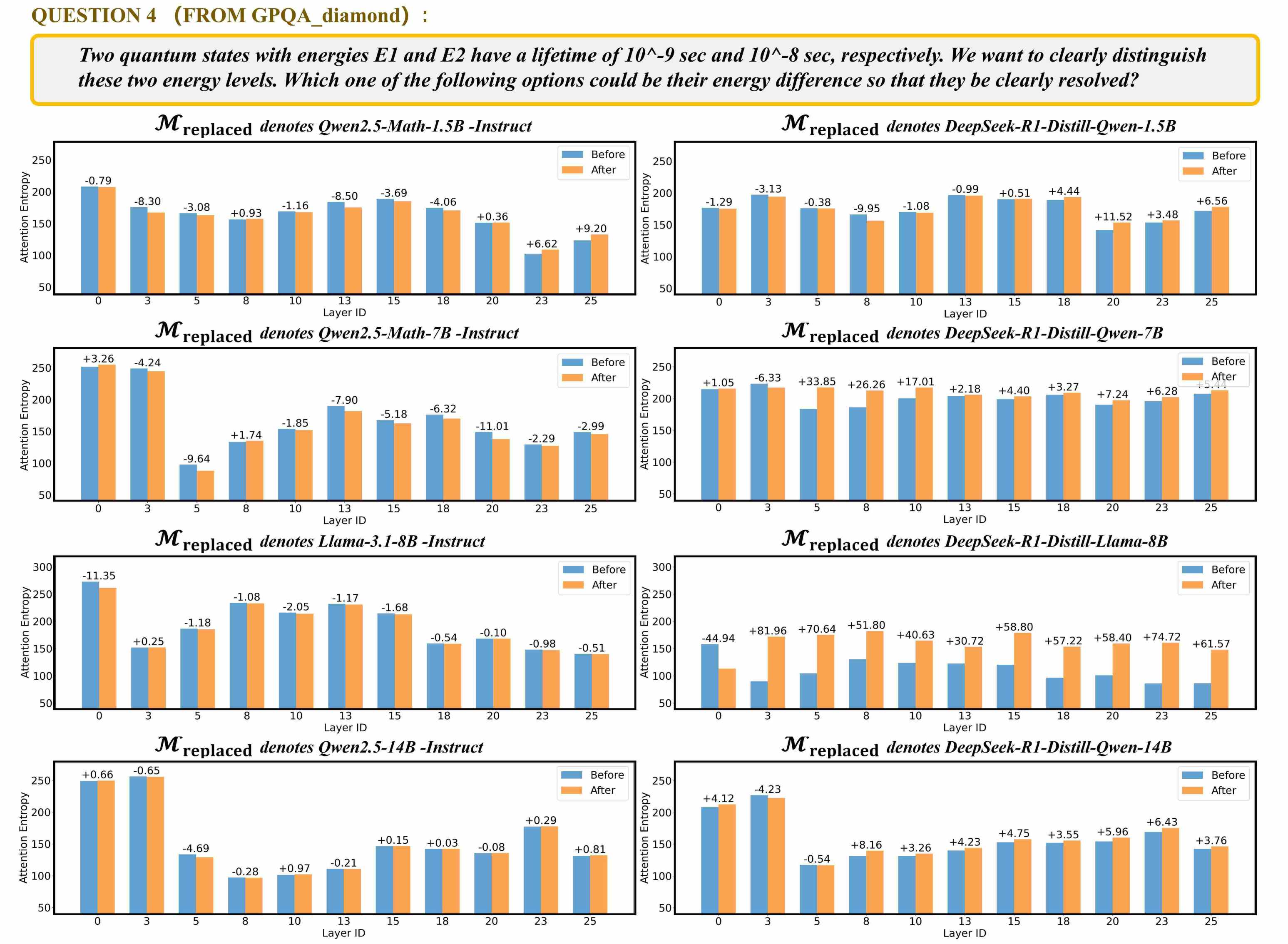}
    \caption{Attention entropy for different $\mathcal{M}_\text{replaced}$. The example input is from GPQA (diamond).}  
    \label{attn_entropy_04}
\end{figure}
 \newpage
The replaced models $\mathcal{M}_\text{replaced}$, spanning diverse architectures and parameter scales, consistently preserve the attention entropy of their original counterparts across a range of examples . This robustness persists even under higher scaling of the singular values in the $W_O$ of \textsc{reasoning} models. In particular, \textit{Qwen}-based models exhibit minimal sensitivity to such modifications, with attention entropy remaining largely unchanged (Figures \ref{attn_entropy_01}, \ref{attn_entropy_02}, \ref{attn_entropy_03}, \ref{attn_entropy_04}). In contrast, \textit{LLaMA}-based \textsc{reasoning} models show an increase in attention entropy when the overall scale of $W_O$ singular values is reduced, consistent with a more uniform distribution of attention scores. Importantly, these effects are largely invariant to extreme amplification of singular values in the long tail of the spectrum. 

\section{Experiments on verifying the consistency of orthogonal transformations}
\label{restoration}
This section highlights the critical importance of orthogonal consistency. While the main paper only demonstrates that disrupting orthogonal transformations in SA output subspaces can be compensated by preserving orthogonality in input subspaces, we present here a more extensive set of experimental results. We apply Construction \ref{reduce_Q} to matrices in $\mathcal{M}_\text{post}$ to obtain $\mathcal{M}^\text{ablation}_\text{post}$, and use Construction \ref{retain_Q} to derive $\mathcal{M}^\text{restoration}_\text{post}$. These operations model the destruction and subsequent restoration of the output subspaces in the weight matrices. Similarly, we apply Constructions \ref{reduce_Q_input} and \ref{retain_Q_input} to the input subspaces, as a symmetric counterpart to Constructions \ref{reduce_Q} and \ref{retain_Q}:
\begin{equation}
  W^{(i)}_\text{post} \leftarrow \boldsymbol{U^{(i)}_\text{base}}\Sigma_\text{post} \cdot {V^{(i)}_\text{post}}^T
  \label{reduce_Q_input}
\end{equation}
\begin{equation}
  W^{(i)}_\text{post} \leftarrow \boldsymbol{(U^{(i)}_\text{base}Q)}\cdot\Sigma_\text{post}{V^{(i)}_\text{post}}^T = \boldsymbol{(U^{(i)}_\text{base}\cdot {V^{(i)}_\text{base}}^T{V^{(i)}_\text{post}})}\cdot \Sigma_\text{post} {V^{(i)}_\text{post}}^T
  \label{retain_Q_input}
\end{equation}

Constructions \ref{reduce_Q}, \ref{retain_Q}, \ref{reduce_Q_input}, and \ref{retain_Q_input} provide an intuitive demonstration of the orthogonal consistency between the left and right singular vectors of each weight matrix in the model. For each $\mathcal{M}_\text{post}$, we apply the transformations from Constructions \ref{reduce_Q}, \ref{retain_Q}, \ref{reduce_Q_input}, and \ref{retain_Q_input} to all SA or FFN modules. These operations disrupt the orthogonal transformations of either the input or output subspaces, and attempt to restore them using the corresponding orthogonal mappings. This yields eight model variants: $\mathcal{M}^{SA,out}_\text{ablation}$, $\mathcal{M}^{SA,out}_\text{restoration}$, $\mathcal{M}^{SA,in}_\text{ablation}$, $\mathcal{M}^{SA,in}_\text{restoration}$, $\mathcal{M}^{FFN,out}_\text{ablation}$, $\mathcal{M}^{FFN,out}_\text{restoration}$, $\mathcal{M}^{FFN,in}_\text{ablation}$, and $\mathcal{M}^{FFN,in}_\text{restoration}$. The superscript indicates whether the operation is applied to the input or output subspaces of all weight matrices in SAs or FFNs, while the subscript denotes whether the operation is destructive or restorative. We perform ablation and restoration operations on SAs and FFNs separately, to prevent model collapse caused by excessive cumulative errors when restoring all weight matrices simultaneously. Additionally, this approach enables independent validation of the co-rotation phenomenon between the input-output subspaces of SAs and FFNs, avoiding excessive cumulative errors that could interfere with experimental observations.

\subsection{Performance of Different Restoration Models}
\label{performance_restoration_models}
We report the performance of all \textsc{restoration} models on GSM8K, MATH-500, MMLU (dev split), and GPQA. All experimental configurations remain consistent with Appendix \ref{test_replaced_model}, specifically with the temperature set to 0.2, top\_p to 0.95, and a maximum output token length of 1024. The system prompts are as detailed in Appendix \ref{system_prompt}. For each of the four datasets, we measure the results three times and report their pass@1 accuracy (\%). All \textsc{ablation} models were unable to produce valid outputs, inevitably yielding a pass@1 accuracy of 0\% in every evaluation. As these uniformly null results do not provide additional empirical insight, we refrain from reporting them in detail.
The complete results are shown in Table \ref{additional_acc_resotration} and \ref{additional_token_length_resotration}. 

Most \textsc{restoration} models successfully recover the original performance, validating the consistency of co-rotational alignment between input and output subspaces and confirming Equation \ref{approx_W}. We further observe that orthogonal substitutions in the output subspaces are more stable than in the input subspaces: $\mathcal{M}^{\_,in}_\text{restoration}$ often performs far worse than $\mathcal{M}^{\_,out}_\text{restoration}$, indicating directional rotational error (Appendix \ref{Visualization_of_other_families_and_scales_orth_sec2}). Errors appear to accumulate along the input-to-output pathway, while reverse elimination can cause collapse. This suggests an inherent asymmetry in co-rotation speed, with one subspace consistently leading the other—an intriguing phenomenon warranting further study.

\begin{table}[!htbp]
  \caption{Performance comparison between original and \textsc{restoration} models across GSM8K, MATH-500, MMLU, and GPQA with pass@1 accuracy (\%). The "-" indicates model collapse.}
  \label{additional_acc_resotration}
  \centering
  \renewcommand{\arraystretch}{1.5}
  \resizebox{0.78\textwidth}{!}{
  \begin{tabular}{lllllll}
    \toprule
    \textsc{base} Models       & \textsc{post} Types      & \textsc{restoration} Types  & GSM8K  & MATH-500 & MMLU (dev) & GPQA \\
    \midrule
    \multirow{10}{*}{\shortstack{\textit{Qwen2.5-}\\\textit{Math-1.5B}}} & \multirow{5}{*}{$\mathcal{M}_\text{Instruct}$}   & $\mathcal{M}_\text{original}$ & 85.14$\pm$0.14 & 65.47$\pm$0.90 & 48.04$\pm$0.60 & 30.44$\pm$0.36 \\
                                                                      & & $\mathcal{M}^{SA,in}_\text{restoration}$ & 84.53$\pm$0.25 & 66.20$\pm$0.16 & 41.28$\pm$0.44 & 27.69$\pm$0.29 \\
                                                                      & & $\mathcal{M}^{SA,out}_\text{restoration}$ & 84.03$\pm$0.29 & 66.47$\pm$1.79 & 38.25$\pm$2.30 & 29.34$\pm$2.65 \\
                                                                      & & $\mathcal{M}^{FFN,in}_\text{restoration}$ & 61.54$\pm$0.19 & 53.00$\pm$0.20 & 31.81$\pm$0.41 & 28.79$\pm$0.83 \\
                                                                      & & $\mathcal{M}^{FFN,out}_\text{restoration}$ & 84.51$\pm$0.18 & 66.07$\pm$0.31 & 41.17$\pm$0.88 & 22.97$\pm$1.10 \\
    \cmidrule(l){2-7}
                                      & \multirow{5}{*}{$\mathcal{M}_\text{Reasoning}$}   & $\mathcal{M}_\text{original}$ & 62.88$\pm$0.59 & 32.73$\pm$1.64 & 25.02$\pm$0.59 & 7.02$\pm$0.44 \\
                                                                      & & $\mathcal{M}^{SA,in}_\text{restoration}$ & 61.54$\pm$1.19 & 30.93$\pm$0.57 & 29.00$\pm$0.44 & 6.75$\pm$0.27 \\
                                                                      & & $\mathcal{M}^{SA,out}_\text{restoration}$ & 61.96$\pm$1.71 & 32.06$\pm$0.25 & 28.30$\pm$1.77 & 3.45$\pm$1.23 \\
                                                                      & & $\mathcal{M}^{FFN,in}_\text{restoration}$ & 60.60$\pm$1.25 & 53.60$\pm$0.43 & 25.49$\pm$1.07 & 12.81$\pm$1.44 \\
                                                                      & & $\mathcal{M}^{FFN,out}_\text{restoration}$ & 76.05$\pm$0.71 & 56.46$\pm$0.34 & 32.51$\pm$3.03 & 16.71$\pm$1.81 \\
    \midrule
    \multirow{10}{*}{\shortstack{\textit{Qwen2.5-}\\\textit{Math-7B}}} & \multirow{5}{*}{$\mathcal{M}_\text{Instruct}$}   & $\mathcal{M}_\text{original}$ & 95.75$\pm$0.12 & 70.06$\pm$0.50 & 55.90$\pm$0.16 & 27.14$\pm$0.49 \\
                                                                      & & $\mathcal{M}^{SA,in}_\text{restoration}$ & 95.15$\pm$0.41 & 73.20$\pm$0.33 & 55.18$\pm$0.18 & 24.85$\pm$0.17 \\
                                                                      & & $\mathcal{M}^{SA,out}_\text{restoration}$ & 94.31$\pm$0.98 & 72.40$\pm$0.53 & 53.10$\pm$1.46 & 20.80$\pm$1.60 \\
                                                                      & & $\mathcal{M}^{FFN,in}_\text{restoration}$ & 86.10$\pm$0.53 & 68.60$\pm$1.40 & 54.04$\pm$0.61 & 25.07$\pm$0.98 \\
                                                                      & & $\mathcal{M}^{FFN,out}_\text{restoration}$ & 94.21$\pm$0.86 & 70.93$\pm$1.51 & 55.44$\pm$3.35 & 25.89$\pm$1.44 \\
    \cmidrule(l){2-7}
                                      & \multirow{5}{*}{$\mathcal{M}_\text{Reasoning}$}   & $\mathcal{M}_\text{original}$ & 62.70$\pm$1.05 & 47.60$\pm$0.33 & 58.71$\pm$0.91 & 14.73$\pm$0.97 \\
                                                                      & & $\mathcal{M}^{SA,in}_\text{restoration}$ & 63.21$\pm$0.91 & 52.80$\pm$0.28 & 58.48$\pm$0.65 & 22.99$\pm$1.19 \\
                                                                      & & $\mathcal{M}^{SA,out}_\text{restoration}$ & 64.34$\pm$2.29 & 50.93$\pm$1.36 & 59.06$\pm$0.73 & 21.34$\pm$0.69 \\
                                                                      & & $\mathcal{M}^{FFN,in}_\text{restoration}$ & 82.46$\pm$0.90 & 65.60$\pm$2.91 & 48.42$\pm$0.70 & 22.71$\pm$1.13 \\
                                                                      & & $\mathcal{M}^{FFN,out}_\text{restoration}$ & 58.83$\pm$1.66 & 60.07$\pm$1.75 & 58.83$\pm$0.73 & 20.16$\pm$2.42 \\
    \midrule
    \multirow{10}{*}{\textit{Llama-3.1-8B}} & \multirow{5}{*}{$\mathcal{M}_\text{Instruct}$}   & $\mathcal{M}_\text{original}$ & 34.70$\pm$1.24 & 31.46$\pm$1.06 & 67.48$\pm$0.44 & 21.21$\pm$0.29 \\
                                                                      & & $\mathcal{M}^{SA,in}_\text{restoration}$ & 30.15$\pm$0.82 & 30.40$\pm$0.75 & 65.49$\pm$0.43 & 22.32$\pm$0.09 \\
                                                                      & & $\mathcal{M}^{SA,out}_\text{restoration}$ & 31.18$\pm$1.17 & 33.13$\pm$1.70 & 63.74$\pm$2.66 & 25.07$\pm$2.16 \\
                                                                      & & $\mathcal{M}^{FFN,in}_\text{restoration}$ & 24.13$\pm$2.12 & 23.40$\pm$1.91 & 59.64$\pm$0.93 & 22.61$\pm$1.19 \\
                                                                      & & $\mathcal{M}^{FFN,out}_\text{restoration}$ & 43.97$\pm$2.06 & 23.26$\pm$1.28 & 63.62$\pm$2.92 & 21.98$\pm$1.29 \\
    \cmidrule(l){2-7}
                                      & \multirow{5}{*}{$\mathcal{M}_\text{Reasoning}$}   & $\mathcal{M}_\text{original}$ & 60.17$\pm$0.07 & 32.73$\pm$0.41 & 52.51$\pm$1.47 & 11.40$\pm$0.17 \\
                                                                      & & $\mathcal{M}^{SA,in}_\text{restoration}$ & 60.30$\pm$1.54 & 29.60$\pm$0.49 & 42.22$\pm$0.59 & 8.77$\pm$0.60 \\
                                                                      & & $\mathcal{M}^{SA,out}_\text{restoration}$ & 61.25$\pm$0.78 & 34.87$\pm$1.17 & 47.13$\pm$2.28 & 6.81$\pm$1.63 \\
                                                                      & & $\mathcal{M}^{FFN,in}_\text{restoration}$ & 39.87$\pm$1.13 & 15.33$\pm$3.89 & 38.95$\pm$0.70 & 8.99$\pm$2.13 \\
                                                                      & & $\mathcal{M}^{FFN,out}_\text{restoration}$ & 38.76$\pm$1.09 & 25.00$\pm$2.31 & 47.83$\pm$1.93 & 7.53$\pm$1.50 \\
    \midrule
    \multirow{10}{*}{\textit{Qwen2.5-14B}} & \multirow{5}{*}{$\mathcal{M}_\text{Instruct}$}   & $\mathcal{M}_\text{original}$ & 94.24$\pm$0.29 & 70.53$\pm$0.34 & 90.63$\pm$0.16 & 36.65$\pm$0.36 \\
                                                                      & & $\mathcal{M}^{SA,in}_\text{restoration}$ & 94.09$\pm$0.34 & 68.86$\pm$0.50 & 88.42$\pm$0.29 & 37.60$\pm$0.34 \\
                                                                      & & $\mathcal{M}^{SA,out}_\text{restoration}$ & 93.91$\pm$1.52 & 73.67$\pm$0.92 & 88.07$\pm$1.95 & 32.51$\pm$0.63 \\
                                                                      & & $\mathcal{M}^{FFN,in}_\text{restoration}$ & 93.63$\pm$0.38 & 71.33$\pm$0.83 & 82.57$\pm$3.58 & 28.89$\pm$1.66 \\
                                                                      & & $\mathcal{M}^{FFN,out}_\text{restoration}$ & 94.87$\pm$0.64 & 73.60$\pm$1.11 & 88.30$\pm$0.73 & 34.05$\pm$3.40 \\
    \cmidrule(l){2-7}
                                      & \multirow{5}{*}{$\mathcal{M}_\text{Reasoning}$}   & $\mathcal{M}_\text{original}$ & 70.61$\pm$0.46 & 53.13$\pm$0.25 & 77.89$\pm$0.76 & 19.48$\pm$0.55 \\
                                                                      & & $\mathcal{M}^{SA,in}_\text{restoration}$ & 75.72$\pm$0.25 & 56.46$\pm$0.24 & 76.37$\pm$1.85 & 21.94$\pm$0.86 \\
                                                                      & & $\mathcal{M}^{SA,out}_\text{restoration}$ & 76.32$\pm$1.69 & 56.33$\pm$1.70 & 78.83$\pm$3.06 & 17.17$\pm$1.91 \\
                                                                      & & $\mathcal{M}^{FFN,in}_\text{restoration}$ & - & - & - & - \\
                                                                      & & $\mathcal{M}^{FFN,out}_\text{restoration}$ & 82.15$\pm$1.41 & 62.60$\pm$1.39 & 76.84$\pm$3.35 & 27.06$\pm$3.95 \\
    \bottomrule
  \end{tabular}}
\end{table}

\begin{table}[!htbp]
  \caption{Comparison of average length of output tokens between original and \textsc{restoration} Models across GSM8K, MATH-500, MMLU, and GPQA. The "-" indicates model collapse.}
  \label{additional_token_length_resotration}
  \centering
  \renewcommand{\arraystretch}{1.5}
  \resizebox{0.8\textwidth}{!}{
  \begin{tabular}{lllllll}
    \toprule
    \textsc{base} Models       & \textsc{post} Types      & \textsc{restoration} Types  & GSM8K  & MATH-500 & MMLU (dev) & GPQA \\
    \midrule
    \multirow{10}{*}{\shortstack{\textit{Qwen2.5-}\\\textit{Math-1.5B}}} & \multirow{5}{*}{$\mathcal{M}_\text{Instruct}$}   & $\mathcal{M}_\text{original}$ & 305.01$\pm$1.54 & 542.32$\pm$1.21 & 402.60$\pm$3.13 & 633.82$\pm$5.09 \\
                                                                      & & $\mathcal{M}^{SA,in}_\text{restoration}$ & 309.47$\pm$15.81 & 523.06$\pm$5.87 & 435.36$\pm$8.72 & 646.07$\pm$6.98 \\
                                                                      & & $\mathcal{M}^{SA,out}_\text{restoration}$ & 287.12$\pm$6.99 & 558.05$\pm$3.83 & 447.05$\pm$8.25 & 631.88$\pm$3.64 \\
                                                                      & & $\mathcal{M}^{FFN,in}_\text{restoration}$ & 422.87$\pm$25.85 & 587.42$\pm$7.66 & 532.19$\pm$4.54 & 792.16$\pm$7.86 \\
                                                                      & & $\mathcal{M}^{FFN,out}_\text{restoration}$ & 320.65$\pm$8.86 & 499.06$\pm$13.76 & 443.56$\pm$1.18 & 617.73$\pm$2.57 \\
    \cmidrule(l){2-7}
                                      & \multirow{5}{*}{$\mathcal{M}_\text{Reasoning}$}   & $\mathcal{M}_\text{original}$ & 539.82$\pm$6.86 & 911.55$\pm$5.55 & 619.34$\pm$13.82 & 952.00$\pm$18.83 \\
                                                                     & & $\mathcal{M}^{SA,in}_\text{restoration}$ & 504.75$\pm$24.05 & 916.60$\pm$8.58 & 659.16$\pm$8.78 & 920.66$\pm$13.58 \\
                                                                      & & $\mathcal{M}^{SA,out}_\text{restoration}$ & 518.82$\pm$10.24 & 910.68$\pm$19.32 & 661.64$\pm$13.52 & 968.31$\pm$4.19 \\
                                                                      & & $\mathcal{M}^{FFN,in}_\text{restoration}$ & 356.13$\pm$11.35 & 692.21$\pm$6.48 & 466.14$\pm$10.31 & 872.22$\pm$16.03 \\
                                                                      & & $\mathcal{M}^{FFN,out}_\text{restoration}$ & 422.74$\pm$4.12 & 755.90$\pm$5.98 & 502.26$\pm$8.86 & 819.93$\pm$4.54 \\
    \midrule
    \multirow{10}{*}{\shortstack{\textit{Qwen2.5-}\\\textit{Math-7B}}} & \multirow{5}{*}{$\mathcal{M}_\text{Instruct}$}   & $\mathcal{M}_\text{original}$ & 299.46$\pm$3.17 & 551.34$\pm$4.39 & 372.53$\pm$5.91 & 567.34$\pm$4.96 \\
                                                                      & & $\mathcal{M}^{SA,in}_\text{restoration}$ & 320.01$\pm$9.72 & 561.23$\pm$4.63 & 411.70$\pm$3.47 & 665.44$\pm$10.30 \\
                                                                      & & $\mathcal{M}^{SA,out}_\text{restoration}$ & 307.38$\pm$7.85 & 565.77$\pm$15.30 & 420.34$\pm$9.38 & 672.78$\pm$7.23 \\
                                                                      & & $\mathcal{M}^{FFN,in}_\text{restoration}$ & 382.13$\pm$8.09 & 552.38$\pm$3.86 & 642.14$\pm$10.25 & 846.68$\pm$8.97 \\
                                                                      & & $\mathcal{M}^{FFN,out}_\text{restoration}$ & 286.25$\pm$22.59 & 510.28$\pm$11.25 & 345.16$\pm$8.75 & 535.02$\pm$5.42 \\
    \cmidrule(l){2-7}
                                      & \multirow{5}{*}{$\mathcal{M}_\text{Reasoning}$}   & $\mathcal{M}_\text{original}$ & 729.16$\pm$7.64 & 795.40$\pm$9.01 & 514.15$\pm$6.91 & 933.15$\pm$9.97 \\
                                                                      & & $\mathcal{M}^{SA,in}_\text{restoration}$ & 791.97$\pm$21.19 & 617.83$\pm$4.76 & 457.57$\pm$2.16 & 863.81$\pm$2.92 \\
                                                                      & & $\mathcal{M}^{SA,out}_\text{restoration}$ & 796.48$\pm$5.62 & 778.33$\pm$5.57 & 451.87$\pm$7.65 & 877.55$\pm$17.99 \\
                                                                      & & $\mathcal{M}^{FFN,in}_\text{restoration}$ & 423.84$\pm$8.60 & 809.49$\pm$8.49 & 388.25$\pm$7.09 & 824.16$\pm$3.86 \\
                                                                      & & $\mathcal{M}^{FFN,out}_\text{restoration}$ & 442.44$\pm$14.48 & 691.19$\pm$7.95 & 444.99$\pm$12.73 & 823.32$\pm$13.92 \\
    \midrule
    \multirow{10}{*}{\textit{Llama-3.1-8B}} & \multirow{5}{*}{$\mathcal{M}_\text{Instruct}$}   & $\mathcal{M}_\text{original}$ & 166.47$\pm$4.22 & 359.19$\pm$6.02 & 35.79$\pm$1.43 & 236.35$\pm$7.38 \\
                                                                      & & $\mathcal{M}^{SA,in}_\text{restoration}$ & 183.11$\pm$8.15 & 324.01$\pm$2.05 & 32.51$\pm$8.96 & 243.30$\pm$10.17 \\
                                                                      & & $\mathcal{M}^{SA,out}_\text{restoration}$ & 169.65$\pm$4.65 & 343.88$\pm$18.92 & 48.50$\pm$6.12 & 254.77$\pm$9.58 \\
                                                                      & & $\mathcal{M}^{FFN,in}_\text{restoration}$ & 150.22$\pm$3.90 & 278.5$\pm$11.29 & 5.33$\pm$1.24 & 6.01$\pm$1.42 \\
                                                                      & & $\mathcal{M}^{FFN,out}_\text{restoration}$ & 173.32$\pm$7.98 & 247.75$\pm$13.73 & 11.01$\pm$1.41 & 38.74$\pm$1.11 \\
    \cmidrule(l){2-7}
                                      & \multirow{5}{*}{$\mathcal{M}_\text{Reasoning}$}   & $\mathcal{M}_\text{original}$ & 627.14$\pm$8.71 & 931.14$\pm$14.80 & 721.64$\pm$11.13 & 989.41$\pm$7.43 \\
                                                                      & & $\mathcal{M}^{SA,in}_\text{restoration}$ & 410.23$\pm$6.32 & 833.03$\pm$11.39 & 755.99$\pm$15.07 & 989.68$\pm$3.84 \\
                                                                      & & $\mathcal{M}^{SA,out}_\text{restoration}$ & 431.48$\pm$18.15 & 888.37$\pm$17.35 & 768.72$\pm$11.06 & 998.85$\pm$6.39 \\
                                                                      & & $\mathcal{M}^{FFN,in}_\text{restoration}$ & 309.76$\pm$24.51 & 953.37$\pm$14.71 & 684.11$\pm$19.56 & 975.54$\pm$17.14 \\
                                                                      & & $\mathcal{M}^{FFN,out}_\text{restoration}$ & 457.27$\pm$10.21 & 833.03$\pm$11.39 & 672.14$\pm$9.32 & 972.02$\pm$4.06 \\
    \midrule
    \multirow{10}{*}{\textit{Qwen2.5-14B}} & \multirow{5}{*}{$\mathcal{M}_\text{Instruct}$}   & $\mathcal{M}_\text{original}$ & 281.95$\pm$7.21 & 550.02$\pm$6.17 & 89.69$\pm$1.18 & 240.16$\pm$6.55 \\
                                                                     & & $\mathcal{M}^{SA,in}_\text{restoration}$ & 279.14$\pm$7.21 & 444.63$\pm$13.24 & 101.63$\pm$8.73 & 283.74$\pm$9.02 \\
                                                                      & & $\mathcal{M}^{SA,out}_\text{restoration}$ & 182.34$\pm$4.57 & 850.45$\pm$11.08 & 99.50$\pm$5.92 & 275.19$\pm$6.80 \\
                                                                      & & $\mathcal{M}^{FFN,in}_\text{restoration}$ & 288.07$\pm$14.29 & 442.79$\pm$4.03 & 89.41$\pm$3.21 & 188.08$\pm$5.28 \\
                                                                      & & $\mathcal{M}^{FFN,out}_\text{restoration}$ & 282.67$\pm$6.75 & 431.10$\pm$6.25 & 120.54$\pm$11.45 & 217.08$\pm$4.71 \\
    \cmidrule(l){2-7}
                                      & \multirow{5}{*}{$\mathcal{M}_\text{Reasoning}$}   & $\mathcal{M}_\text{original}$ & 583.01$\pm$4.57 & 897.61$\pm$8.81 & 487.54$\pm$7.68 & 924.63$\pm$7.90 \\
                                                                      & & $\mathcal{M}^{SA,in}_\text{restoration}$ & 538.26$\pm$6.08 & 844.46$\pm$8.89 & 442.49$\pm$12.38 & 920.88$\pm$4.77 \\
                                                                      & & $\mathcal{M}^{SA,out}_\text{restoration}$ & 518.71$\pm$11.25 & 852.79$\pm$9.55 & 438.20$\pm$4.33 & 912.47$\pm$5.40 \\
                                                                      & & $\mathcal{M}^{FFN,in}_\text{restoration}$ & - & - & - & - \\
                                                                      & & $\mathcal{M}^{FFN,out}_\text{restoration}$ & 504.96$\pm$8.01 & 863.77$\pm$3.59 & 450.66$\pm$10.42 & 875.01$\pm$11.63 \\
    \bottomrule
  \end{tabular}}
\end{table}

\subsection{CKA Analysis of Different Restoration Models}
\label{cka_restoration_models}
\begin{figure}[!htbp]
    \centering
    \includegraphics[width=0.8\linewidth]{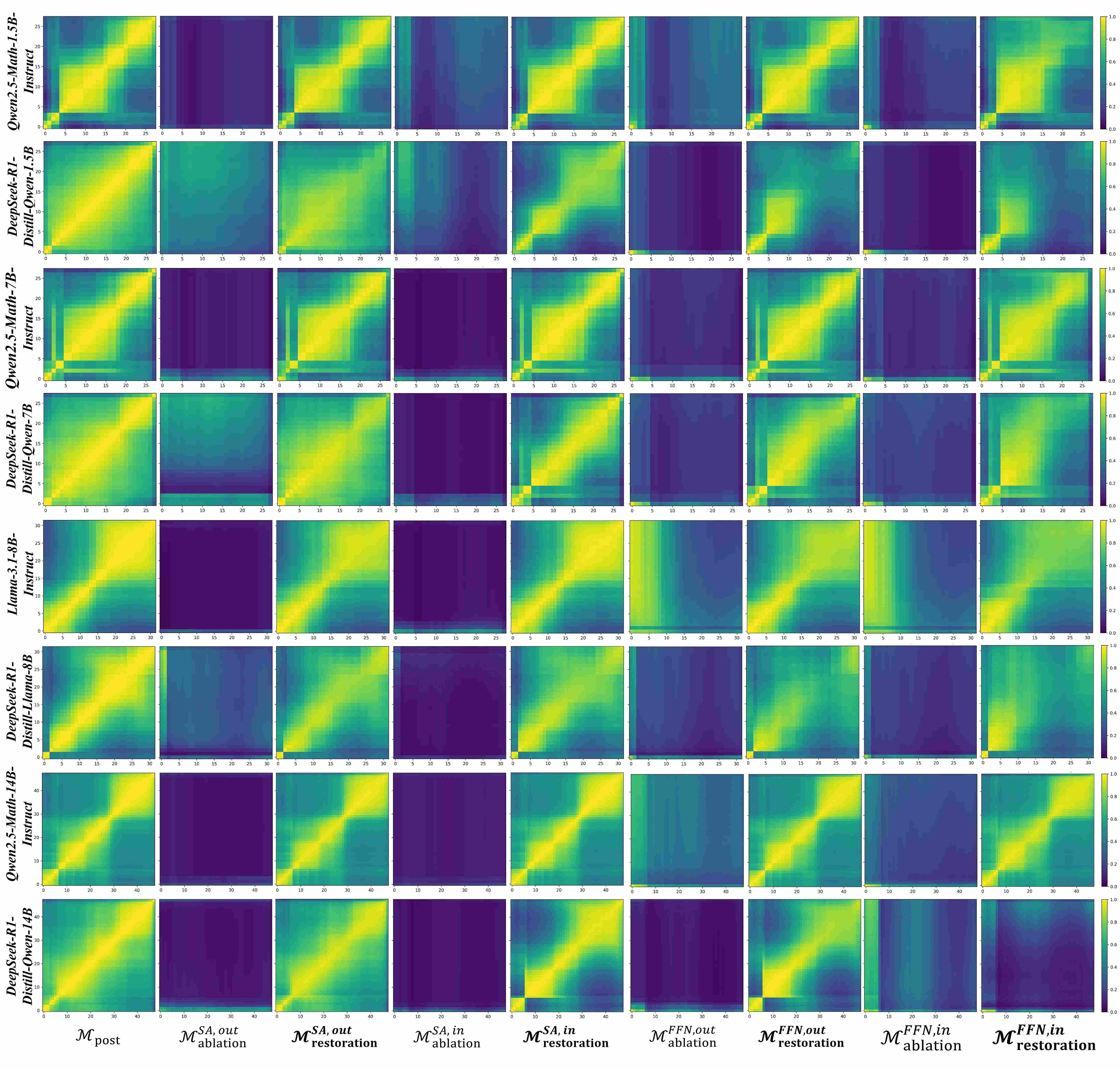}
    \caption{CKA heatmaps generated using $\mathcal{M}_\text{post}$ for $\mathcal{M}_\text{post}$, $\mathcal{M}_\text{ablation}$, and$\mathcal{M}_\text{restoration}$. The results indicate that $\mathcal{M}_\text{Instruct}$ exhibits stronger orthogonal alignment between input and output subspaces compared to $\mathcal{M}_\text{reasoning}$. Additionally, the restoration of orthogonal alignment after perturbation is more robust in the output subspaces than in the input subspaces.}  
    \label{cka_all}
\end{figure}

We then feed $N$ input examples into $\mathcal{M}_\text{post}$, $\mathcal{M}^\text{ablation}_\text{post}$, and $\mathcal{M}^\text{restoration}_\text{post}$, and compute the mean hidden representations $r^{(i)}_\mathcal{M}$ for each layer by averaging their outputs (Equation \ref{hidden_rep}):
\begin{equation}
    r^{(i)}_\mathcal{M}=\frac{1}{N}\sum^{N}_{j=1}\mathcal{M}^{(i)}(T_{j})
    \label{hidden_rep}
\end{equation}
where $T_j$ is the $j$-th input question, and $\mathcal{M}^{(i)}(\cdot)$ denotes the hidden representation produced by the $i$-th Transformer block in model $\mathcal{M}$. We use the first 100 examples from the GSM8K training set for analysis ($N=100$). We compute the CKA heatmap between the average hidden representations of $\mathcal{M}_\text{post}$ and each \textsc{ablation}/\textsc{restoration} variant to assess the impact of orthogonal consistency on internal representations. Figure \ref{cka_all} presents our experimental results. 

Disrupting either the SAs or FFNs compromises the orthogonal alignment between input and output subspaces, impairing the internal structure of $\mathcal{M}_\text{post}$. Restoring this alignment leads to the reemergence of structural symmetry in the CKA heatmaps, indicating a partial recovery of the model’s hidden representations. The weight matrices of $\mathcal{M}_\text{Instruct}$ exhibit stronger orthogonal consistency than those of $\mathcal{M}_\text{reasoning}$. This is evidenced by the restoration variants of $\mathcal{M}_\text{Instruct}$ producing CKA heatmaps that more closely resemble those of $\mathcal{M}_\text{post}$. The CKA heatmaps remain only partially reducible, reflecting the fact that orthogonality is preserved only approximately. This observation is further supported by the correction introduced in Equation \ref{calibration}. The restoration process effectively reinstates the original representational geometry, highlighting the critical structural role of orthogonal transformations.

\section{The Structural Changes in a Broader Range of Models}
\label{case_study}
In the main text, we present a systematic comparison of structural changes in model weights before and after supervised post-training, with a particular focus on the \textit{Qwen} and \textit{LLaMA} families. We also report detailed experimental results that confirm the validity of Equation \ref{approx_W}. These findings naturally motivate several follow-up questions:
\begin{enumerate}[leftmargin=15pt,itemsep=4pt, parsep=0pt, topsep=0pt]
\item How do reinforcement learning (RL)-based post-training methods influence model weights? From the perspective of parameter space, in what ways do their effects differ from those of supervised post-training, and what implications can be drawn?
\item Would modifications to the model architecture or the adoption of different training strategies affect the generalizability of the observed structural changes?
\item Do other components in LLMs with specific functions (such as normalization layers and output projection heads) follow similar patterns?
\end{enumerate}

This section addresses these questions by extending our analysis to a broader set of models. The subsequent case studies provide strong evidence that the validity of Equation \ref{approx_W} is preserved across diverse settings—including supervised post-training, RL-based post-training, and variations in model architecture or training methodology. The two structural changes identified in the main text thus appear to generalize robustly across these scenarios. Furthermore, we observe that this phenomenon persists throughout the entire post-training phase, indicating the continuity of these two structural changes during post-training, as detailed in Appendix \ref{during_post_training}.

\subsection{Structural Changes in LLMs Induced by RL-based Post-Training}
\begin{figure}[!htbp]
    \centering
    \setlength{\abovecaptionskip}{1pt}
    \includegraphics[width=0.8\linewidth]{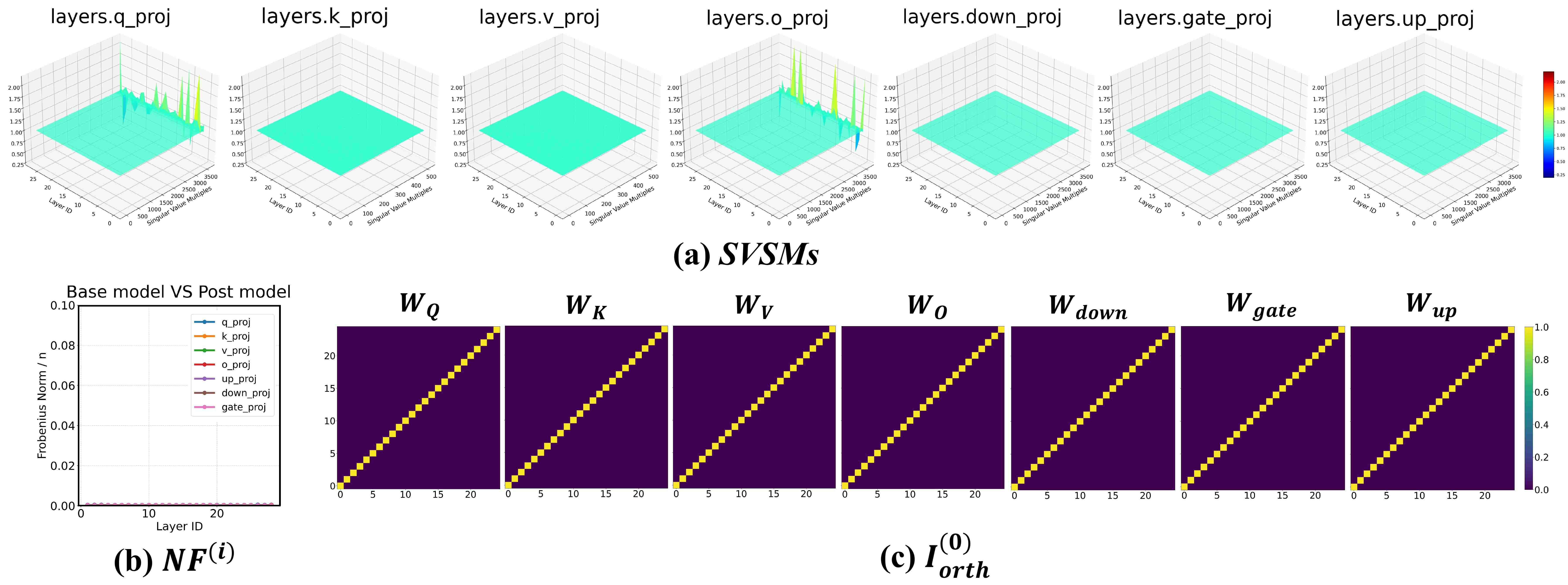}
    \caption{Visualization of structural properties of \textbf{\textit{AceMath-RL-Nemotron-7B}} after post-training. (a) SVSMs reveal that the principal scaling exhibits a near-uniform distribution. (b) $\smash{\mathcal{NF}^{(i)}}$ provides evidence for the consistent orthogonal transformations of the singular vectors. (c) Orthogonality matrices $\smash{I^{(0)}_{orth}}$, shown as an example.
}
    \label{nvidia}
\end{figure}

\begin{figure}[!htbp]
    \centering
    \setlength{\abovecaptionskip}{1pt}
    \includegraphics[width=0.8\linewidth]{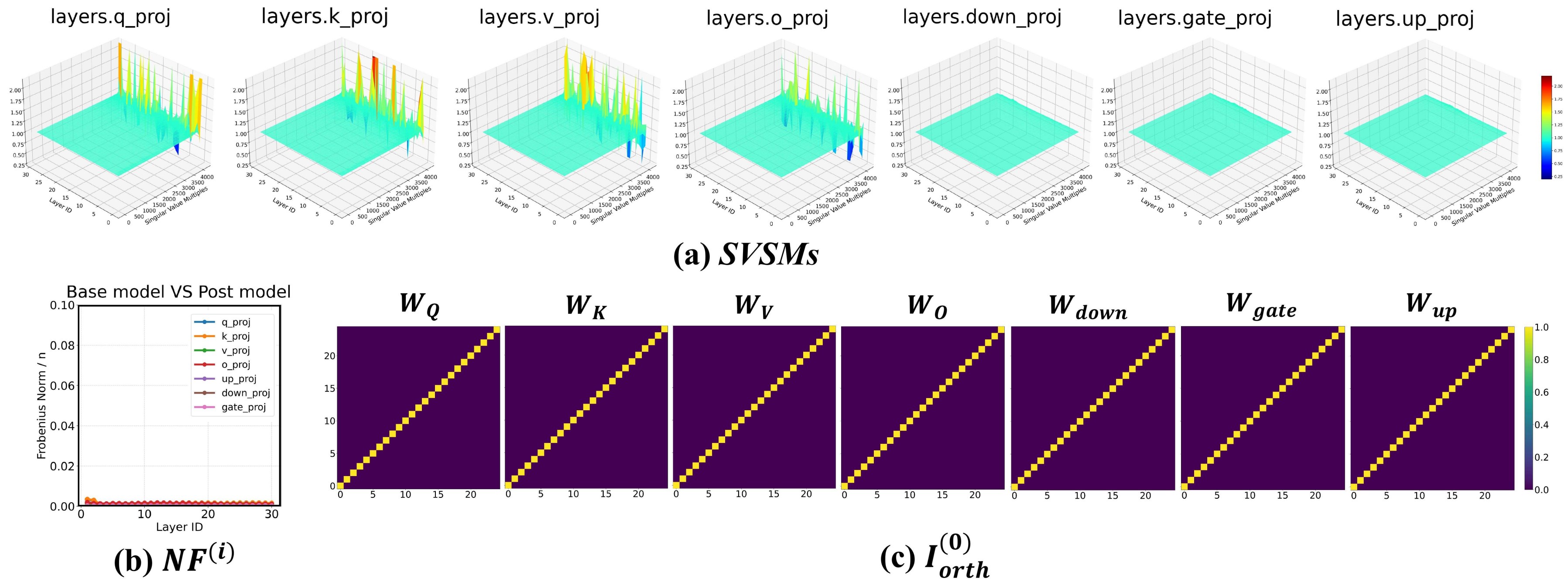}
    \caption{Visualization of structural properties of \textbf{\textit{deepseek-math-7b-rl}} after post-training. The same set of analyses as in Figure \ref{nvidia} is presented, including SVSMs, $\mathcal{NF}^{(i)}$, and orthogonality matrices $\smash{I^{(0)}_{orth}}$.
}
    \label{deepseekmath}
\end{figure}

\begin{figure}[!htbp]
    \centering
    \setlength{\abovecaptionskip}{1pt}
    \includegraphics[width=0.8\linewidth]{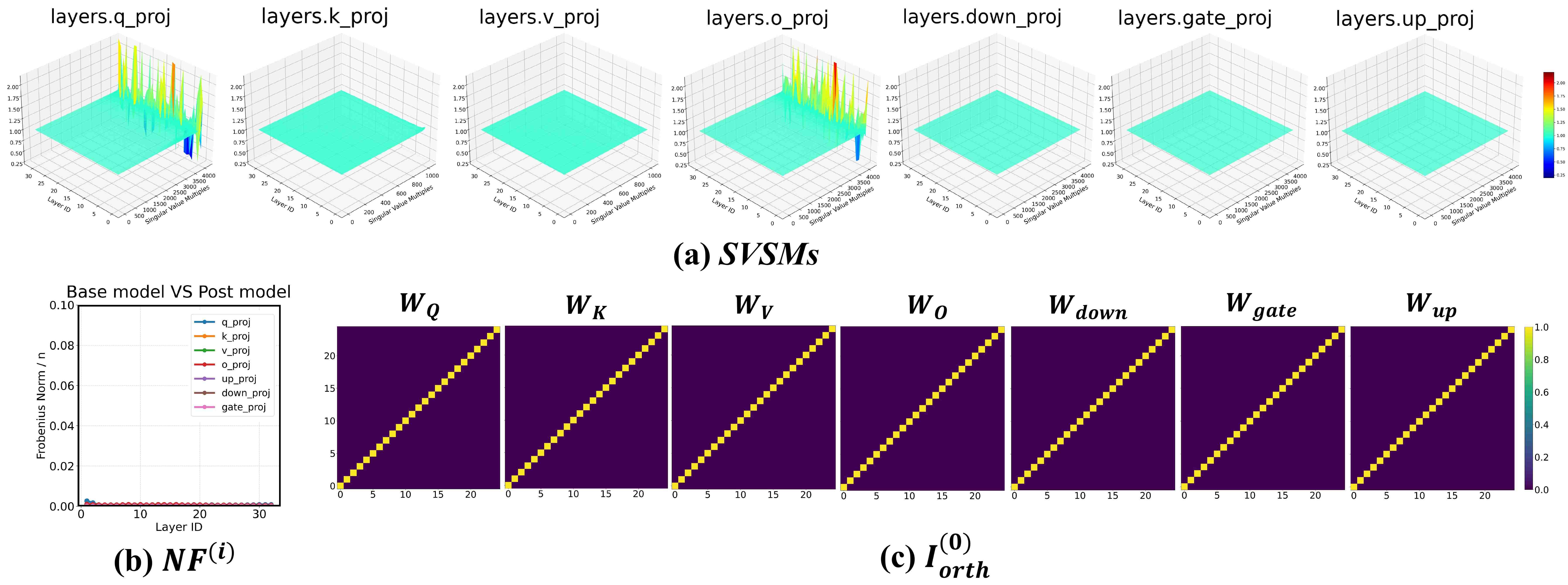}
    \caption{Visualization of structural properties of \textbf{\textit{Seed-X-PPO-7B}} after post-training. The same set of analyses as in Figure \ref{nvidia} is presented, including SVSMs, $\mathcal{NF}^{(i)}$, and orthogonality matrices $\smash{I^{(0)}_{orth}}$.
}
    \label{seed}
\end{figure}

We investigate several state-of-the-art large language models trained with advanced reinforcement learning algorithms, including \textit{AceMath-RL-Nemotron-7B} \cite{acemath2024}, \textit{deepseek-math-7b-rl} \cite{shao2024deepseekmath}, and \textit{Seed-X-PPO-7B} \cite{cheng2025seedxbuildingstrongmultilingual}. These models respectively adopt advanced reinforcement learning approaches such as GRPO \cite{deepseekai2025deepseekr1incentivizingreasoningcapability} and PPO \cite{schulman2017proximalpolicyoptimizationalgorithms}, originate from different research groups, and are built upon diverse training corpora (see Table \ref{version_case_study} for details). This diversity in both algorithmic choices and data sources provides inherent support for the generalizability of our subsequent experimental results. We compute the SVSMs between those models and their \textsc{base} versions, the $\mathcal{NF}^{(i)}$, as well as the orthogonality matrices of the singular vector (e.g., $I^{(0)}_{orth}$ in the first Transformer block), and present the corresponding visualizations in Figures \ref{nvidia}, \ref{deepseekmath}, and \ref{seed}.

From the SVSM heatmaps and the lower values of $\mathcal{NF}^{(i)}$, we observe that models subjected to RL-based post-training exhibit even more consistent structural changes than those trained with SFT-based post-training. \textbf{This strongly suggests that SFT-based and RL-based post-training methods possess a high degree of parameter equivalence, meaning that the effects they impose on model parameters are essentially identical}. Building upon this conclusion, one may infer that RL-based post-training is effectively equivalent to supervised post-training, notwithstanding previous studies \cite{chu2025sft} that have highlighted the ostensibly superior generalization capacity of reinforcement learning algorithms. \textbf{We further conjecture that this generalization advantage does not arise from the intrinsic design of RL algorithms themselves, but rather from the diversity of training data generated through reinforcement learning.} For instance, GRPO encourages the model to produce more diverse responses, which are then incorporated into the training process as additional samples. This analysis further explains the effectiveness of Long-CoT distillation. Its training procedure is equivalent to that of RL-based methods, ensuring comparable effects on model parameters, while its training data are more extensive and diverse than those of instruction tuning, enabling smaller models to achieve reasoning capabilities similar to large-scale RL-based models.

\newpage
\subsection{Generality of Structural Changes Across Training Strategies and Architectures}
We find that \textbf{regardless of architectural modifications or training strategies, LLMs consistently exhibit these two structural changes in their parameters after post-training}. To further examine the universality of this phenomenon, we extend our analysis to \textit{Mistral-7B-Instruct-v0.1} \cite{Mistral_7B}, \textit{Gemma-2-2B-it} \cite{team2024gemma}, and \textit{MediPhi-Instruct} \cite{corbeil2025modular}, each of which incorporates distinct design improvements:

\begin{itemize}[leftmargin=15pt]
    \item \textbf{For \textit{Mistral-7B-Instruct-v0.1},} the model incorporates \textit{Sliding Window Attention} \cite{beltagy2020longformer} and a \textit{Rolling Buffer Cache}. These mechanisms allow each layer's hidden states to access past information within a window size $W$, which is recursively stacked across layers to effectively expand the attention span. As a result, the model achieves a theoretical attention span of approximately 131K tokens. In practice, these improvements substantially reduce memory consumption and enhance computational efficiency without compromising model quality.

    \item \textbf{For \textit{Gemma-2-2B-it},} the model architecture integrates \textit{local sliding window attention} \cite{beltagy2020longformer} and \textit{global attention} \cite{luong2015effectiveapproachesattentionbasedneural}. Local layers operate with a window size of 4096 tokens, global layers extend to 8192 tokens. A \textit{logit soft-capping} \cite{bello2017neuralcombinatorialoptimizationreinforcement} mechanism stabilizes training across attention layers and the final layer, with soft\_cap values set to 50.0 and 30.0. In post-training, the \textsc{base} model firstly undergoes supervised fine-tuning on a mixture of synthetic and human-generated English prompt–response pairs, and then proceeds to \textit{Reinforcement learning with Human Feedback (RLHF)}  \cite{ouyang2022traininglanguagemodelsfollow}, guided by a reward model trained on preference data to align behavior with human intent. The resulting models from each stage are averaged, improving stability and overall performance, and producing an instruction-tuned model optimized for both effectiveness and safety.

    \item \textbf{For \textit{MediPhi-Instruct},} the model still follows a decoder-only Transformer architecture, but the computations of its SAs and FFNs differ from the previously mentioned models. In the case of SAs, given the input $h$, the query ($Q$), key ($K$), and value ($V$) are computed using a single weight matrix $W_{QKV}$:
    \begin{equation}
        Q, K, V = chunk(QKV), \;\;\;\;QKV=hW_{QKV}
        \label{WQKV}
    \end{equation}
    where $\text{chunk}(\cdot)$ splits $QKV$ into $Q,K,V$ along the last dimension. Similarly, for the FFNs, \textit{MediPhi-Instruct} also merges $W_{\text{gate}}$ and $W_{\text{up}}$. 
    As a result, there are only four types of matrices in both the SAs and FFNs, namely $W_{QKV}$, $W_{O}$,
    $W_{\text{gate\_up}}$ and $W_{\text{down}}$. In addition to the architectural modifications, \textit{MediPhi-Instruct} also undergoes an SFT-based post-training stage that integrates domain-specific medical knowledge. Similar to other medical instruction-tuned models such as \textit{Aloe} \cite{gururajan2024aloefamilyfinetunedopen} and \textit{Med42 v2} \cite{christophe2024med42v2suiteclinicalllms}, this stage leverages medical question-answering datasets and benchmark training sets such as \textit{PubMedQA} \cite{jin2019pubmedqadatasetbiomedicalresearch}, thereby aligning the model more closely with medical reasoning and instruction-following tasks.
\end{itemize}

\begin{figure}[!htbp]
    \centering
    \setlength{\abovecaptionskip}{1pt}
    \includegraphics[width=0.8\linewidth]{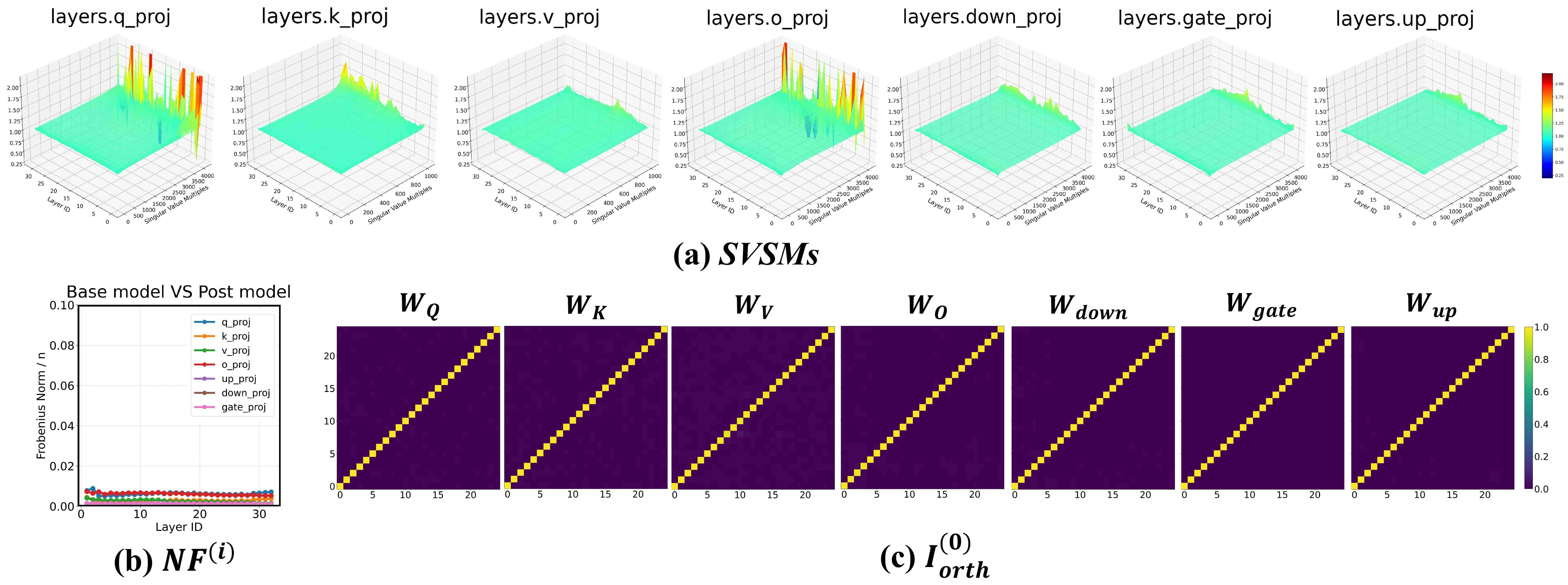}
    \caption{Visualization of structural properties of \textbf{\textit{Mistral-7B-Instruct-v0.1}} after post-training. (a) SVSMs reveal that the principal scaling exhibits a near-uniform distribution. (b) $\smash{\mathcal{NF}^{(i)}}$ provides evidence for the consistent orthogonal transformations of the singular vectors. (c) Orthogonality matrices $\smash{I^{(0)}_{orth}}$, shown as an example.
}
    \label{mistral}
\end{figure}

\begin{figure}[!htbp]
    \centering
    \setlength{\abovecaptionskip}{1pt}
    \includegraphics[width=0.8\linewidth]{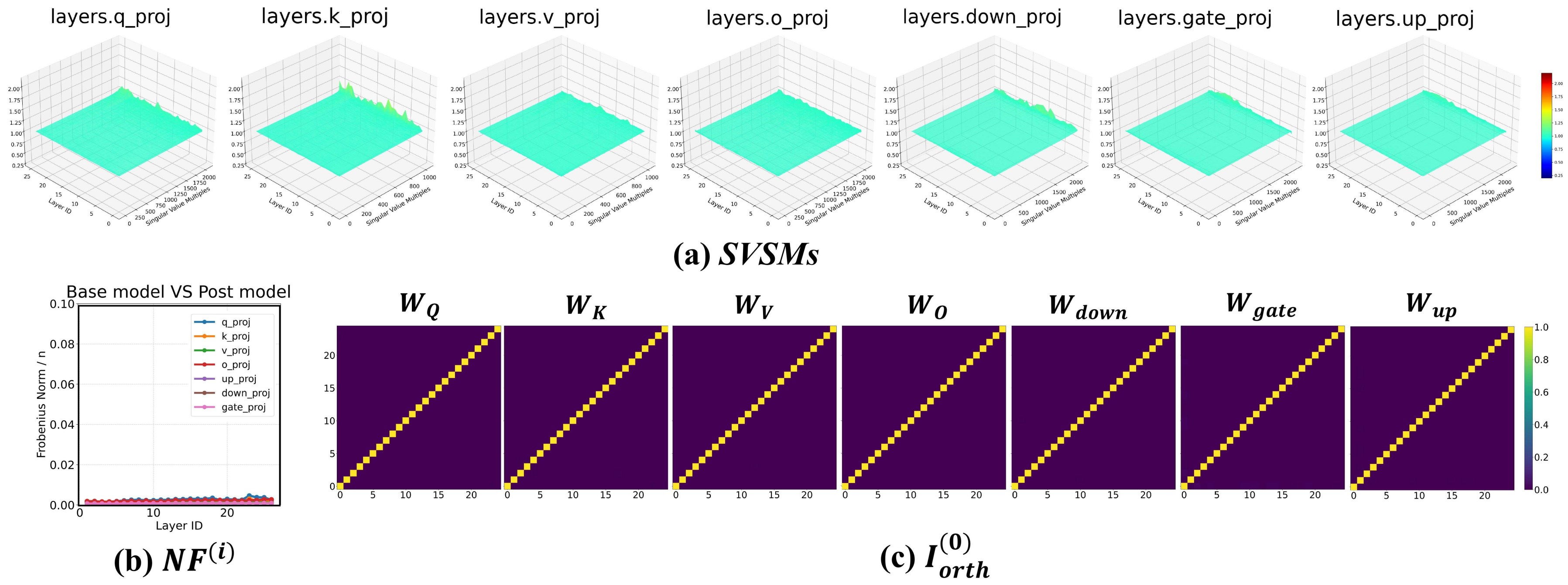}
    \caption{Visualization of structural properties of \textbf{\textit{Gemma-2-2B-it}} after post-training. The same set of analyses as in Figure \ref{mistral} is presented, including SVSMs, $\mathcal{NF}^{(i)}$, and orthogonality matrices $\smash{I^{(0)}_{orth}}$.
}
    \label{gemma2}
\end{figure}

\begin{figure}[!htbp]
    \centering
    \setlength{\abovecaptionskip}{1pt}
    \includegraphics[width=0.8\linewidth]{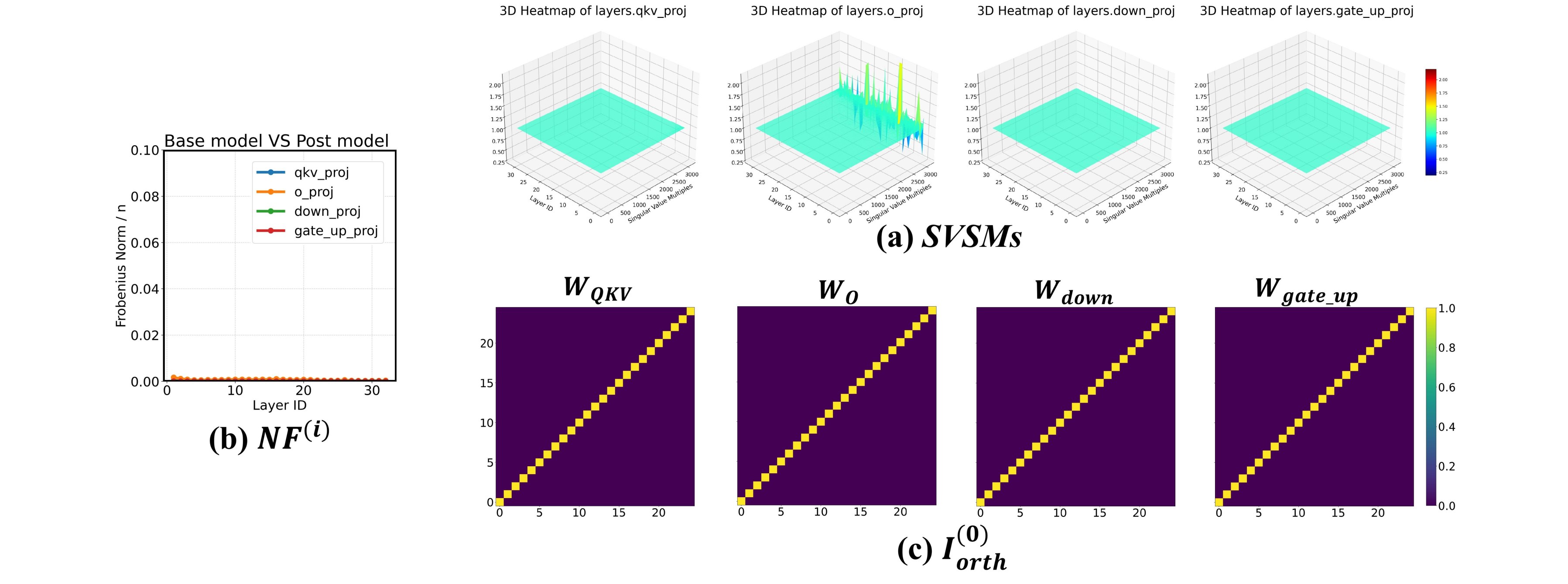}
    \caption{Visualization of structural properties of \textbf{\textit{MediPhi-Instruct}} after post-training. The same set of analyses as in Figure \ref{mistral} is presented, including SVSMs, $\mathcal{NF}^{(i)}$, and orthogonality matrices $\smash{I^{(0)}_{orth}}$.
}
    \label{med_phi}
\end{figure}

More detailed information regarding the aforementioned models will be presented in Table \ref{version_case_study}. We compute the SVSMs between those models and their \textsc{base} versions, the $\mathcal{NF}^{(i)}$, as well as the orthogonality matrices of the singular vector (e.g., $I^{(0)}_{orth}$ in the first Transformer block), and present the corresponding visualizations in Figures \ref{mistral}, \ref{gemma2}, and \ref{med_phi}.

The flattened SVSM heatmaps and a relatively low value of $\mathcal{NF}^{(i)}$ indicate that, regardless of whether the modifications stem from changes in the model architecture or adjustments in the training strategy, this structural property consistently persists in the linear layers of large models. In other words, \textbf{Equation \ref{approx_W} can be employed to characterize the parameter changes of large models before and after post-training}. This provides strong evidence for the universality of such structural transformations and further substantiates the reliability of Equation \ref{approx_W}.

\newpage

\subsection{Structural Changes in Other Components of LLMs}
\label{other_components}

We investigate the structural changes of the main linear layers in LLMs in the main text. Although these layers constitute nearly the entire parameter space, other components also play crucial roles. This subsection therefore extends the exploration to the structural changes in the parameter space of functionally important components such as normalization layers and output projection heads. Specifically, we focus on the models listed in Table \ref{version}, where each transformer block employs two RMSnorm layers \cite{jiang2023prermsnormprecrmsnormtransformersequivalent} that serve as the pre-norms for the attention and FFN modules, respectively, to enhance training stability, and an output projection head is added to the final block to convert hidden vectors into a vocabulary distribution.

We visualize the features of normalization layers and output projection heads and unexpectedly find that \textbf{these components still roughly adhere to the parameter law described in Equation \ref{approx_W}},yet exhibit subtle differences.

\begin{figure}[!htbp]
    \vspace{-8pt}
    \centering
    \includegraphics[width=0.85\linewidth]{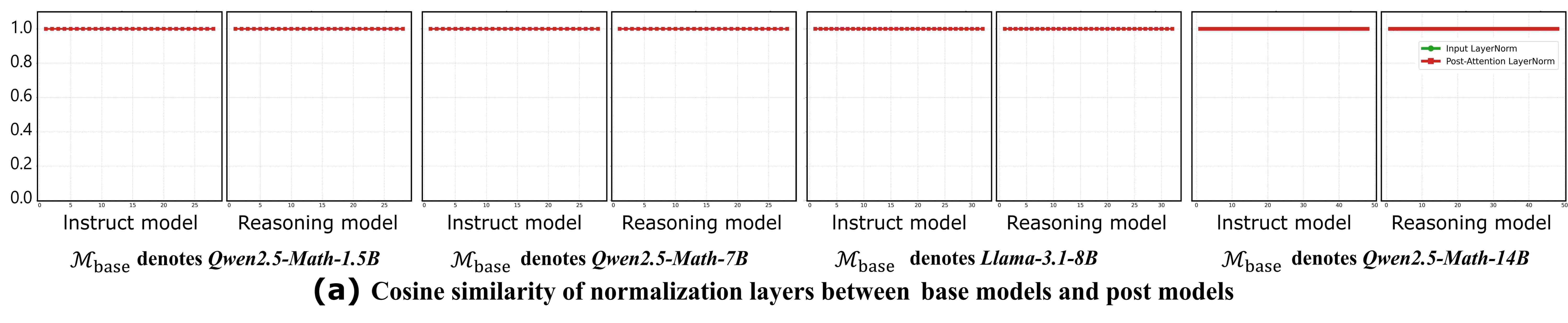}
    \includegraphics[width=0.85\linewidth]{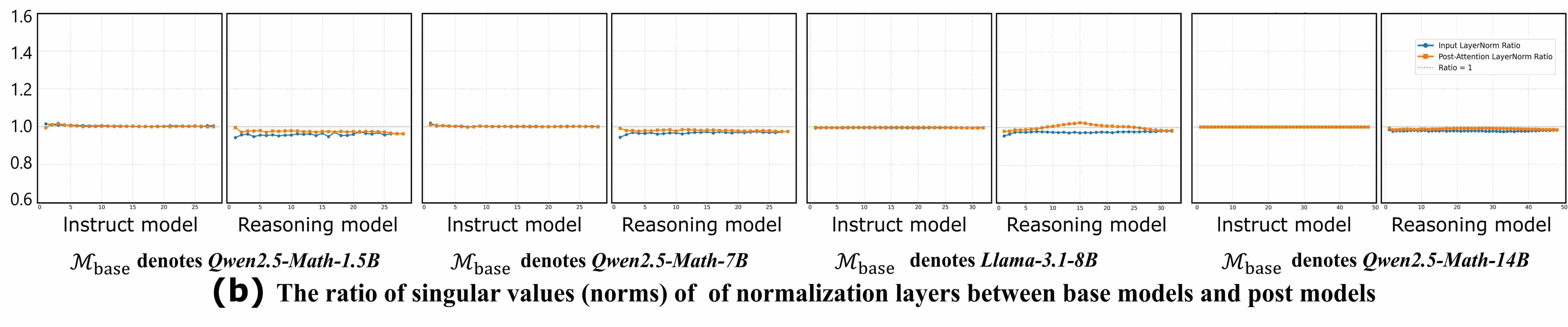}
    \caption{(a) The cosine similarity between the corresponding normalization layers of the \textsc{base} models and \textsc{post} models was calculated. The vast majority of values were equal to 1. (b) The magnitudes of the normalization layers are approximately uniformly scaled but exhibit some fluctuations.}  
    \label{normlization_layers}
    \vspace{-8pt}
\end{figure}

For normalization layers, since the weight often exists as a one-dimensional vector $w$, we consider performing reduced SVD on it:
\begin{equation}
    \label{SVD_on_1d_vector}
    w = a *\sigma * v^T = 1 * ||w||*\frac{w}{||w||}
\end{equation}
For a vector $w$, its left singular vector reduces to $\pm 1$ (assumed to be $1$), its right singular vector becomes the normalized unit vector $\frac{w}{||w||}$, and its singular value is $\|w\|$. For the corresponding normalized weight $w_\text{post}$ of the \textsc{post} model, if Equation \ref{approx_W} holds in Equation \ref{SVD_on_1d_vector}, it implies that the rotation matrix $Q$ of the right singular vector degenerates. In this one-dimensional case, $Q$ becomes a $1 \times 1$ matrix whose sole element is identical to the cosine similarity between $w$ and $w_\text{post}$, which is exactly 1. we can derive that:
\begin{equation}
    \label{SVD_on_1d_vector_co}
    v^Tv_\text{post}=\frac{w}{||w||}\cdot(\frac{w_\text{post}}{||w_\text{post}||})^T=a^Ta_\text{post}=1
\end{equation}
We have experimentally verified this point, as shown in Figure \ref{normlization_layers}a. It can be observed that the cosine similarity between the weights of the normalization layers in the \textsc{post} models and the \textsc{base} models remains consistently at 1. \textbf{It mathematically proves that the normalization layer of each Transformer block only shows uniform and globally consistent scaling during post-training, rather than the channel-wise selective filtering we anticipated}. However, there is some fluctuation in the scaling of their singular values (norms), as shown in Figure \ref{normlization_layers}b. We speculate that this may be related to the unique function of normalization, which involves dynamically adjusting the expressive capacity of the hidden vectors. When the subspace is fixed, this can only be achieved by globally scaling the vector norms, making it difficult for the norms to maintain uniformly consistent scaling across layers.

\begin{figure}[!htbp]
    \centering
    \includegraphics[width=0.8\linewidth]{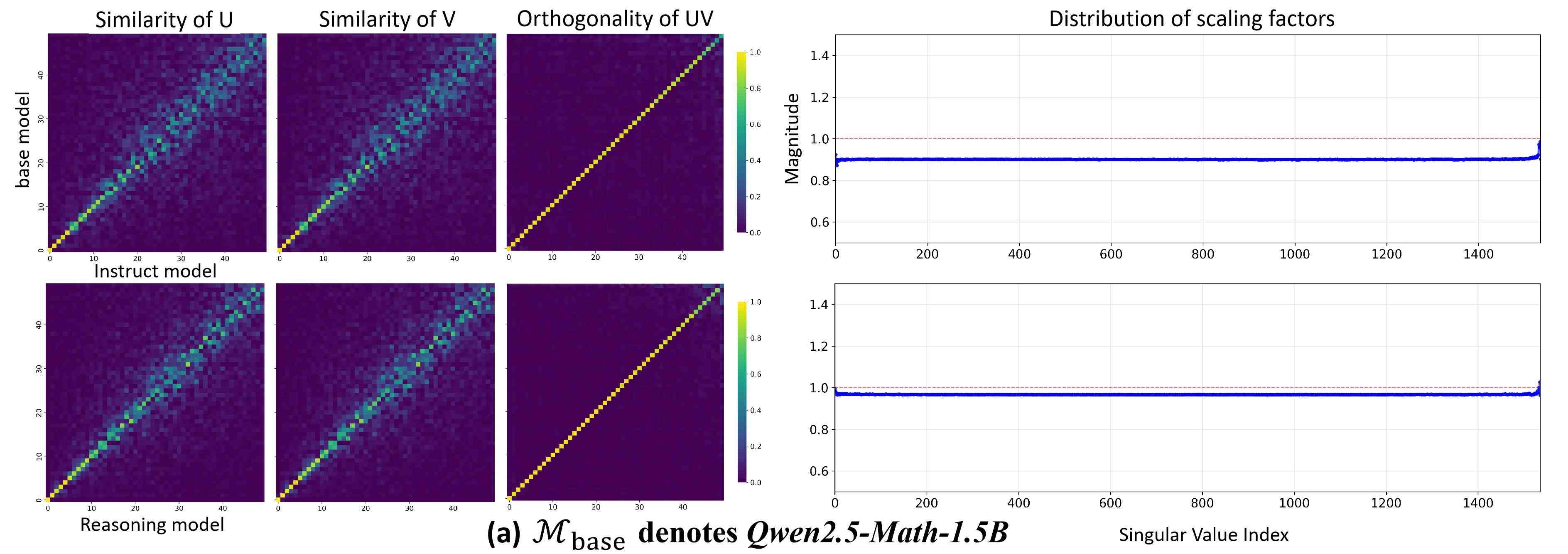}
    \includegraphics[width=0.8\linewidth]{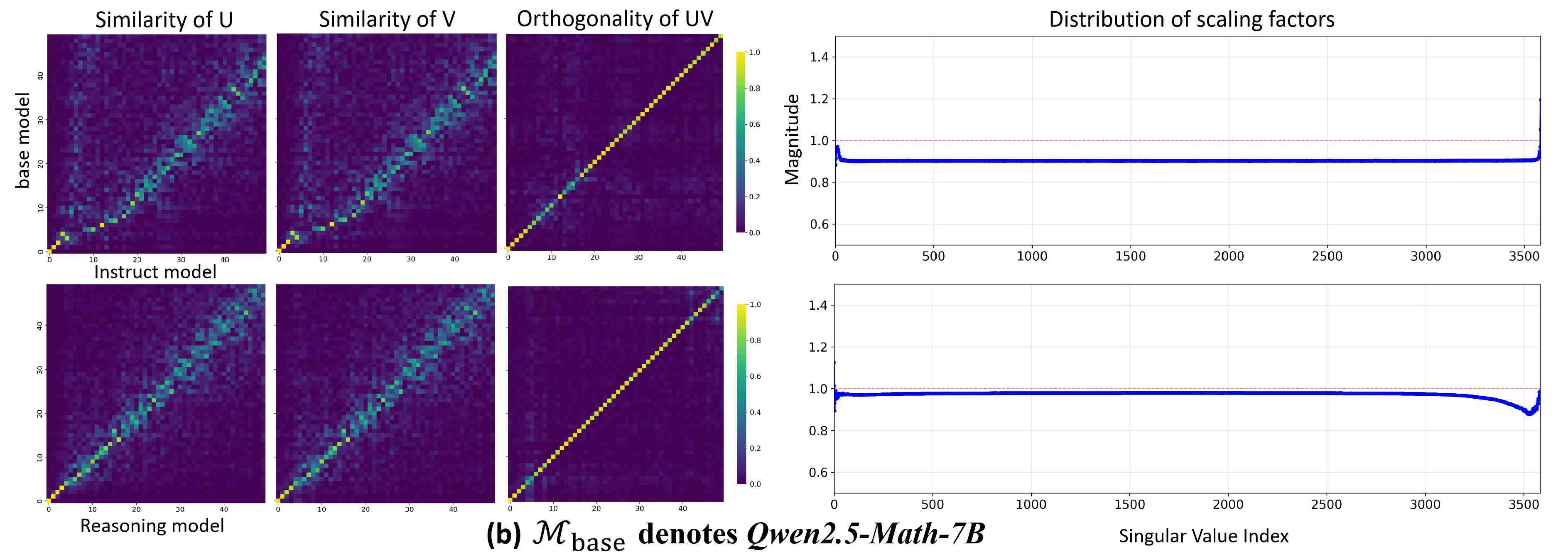}
    \includegraphics[width=0.8\linewidth]{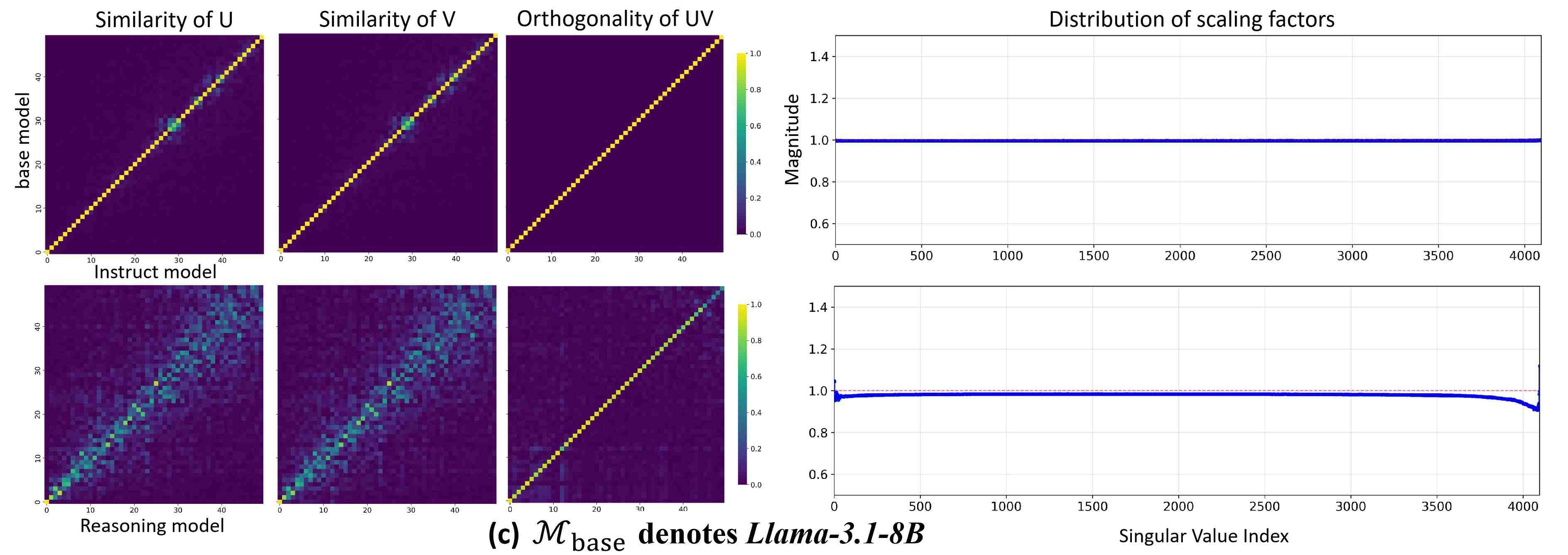}
    \includegraphics[width=0.8\linewidth]{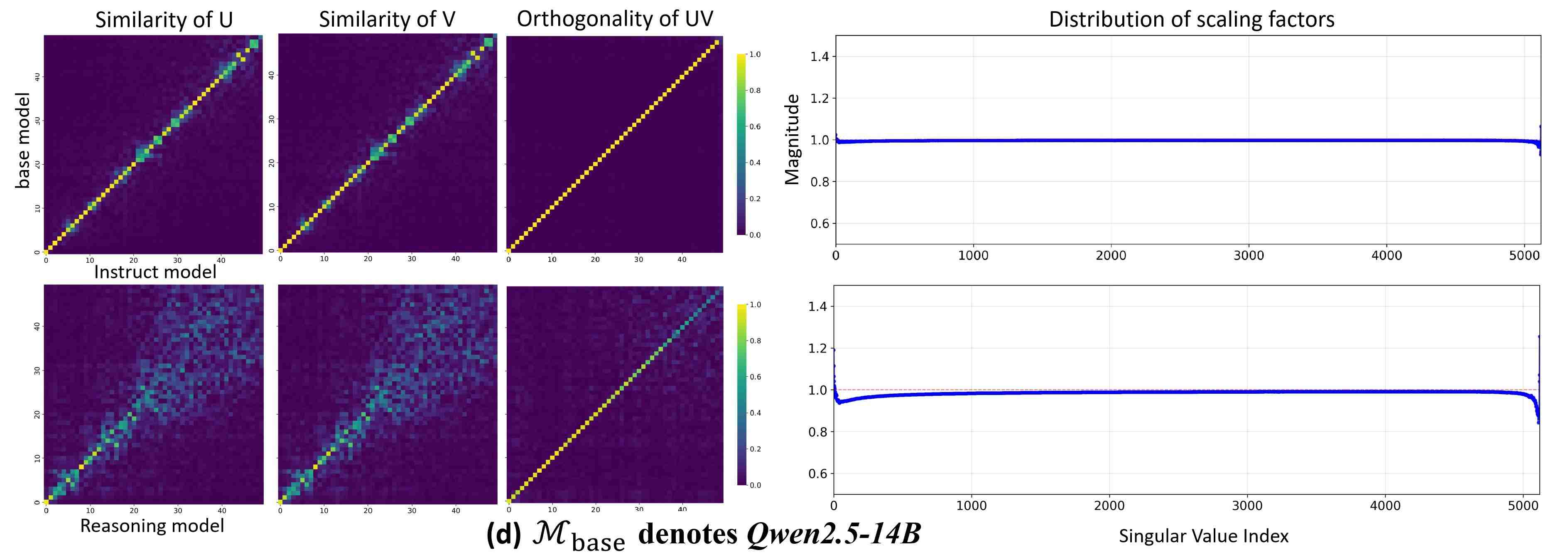}
    \caption{Visualization of the evolution of output projection head properties across model scales. We show the similarity/orthogonality of singular vectors and scaling of singular values before and after post-training.}  
    \label{output_heads}
\end{figure}

Regarding the output projection heads, we plot the left and right similarity matrices against the overall singular value scaling, as shown in Figure \ref{output_heads}. We observe that certain subspaces within the input and output spaces of this component still do not exhibit strong co-rotation. We hypothesize that this stems from the specific function of output projection heads: since they are responsible for mapping hidden states directly to the vocabulary space, \textbf{their parameters are updated directly under the influence of external supervision signals}. As a result, unlike other main linear layers that propagate information through hidden representations, this component experiences greater perturbation of its space during post-training. This makes some of its internal subspaces more susceptible to being reshaped by external supervision, thereby partially hindering appropriate co-rotation. Nevertheless, due to the limited scale of post-training, the structure of the majority of subspaces remains preserved, allowing the output projection heads to largely maintain co-rotation across their subspaces.

\newpage
\subsection{Structural Changes During Post-training}
\label{during_post_training}

\begin{figure}[!htbp]
    \centering
    \includegraphics[width=0.80\linewidth]{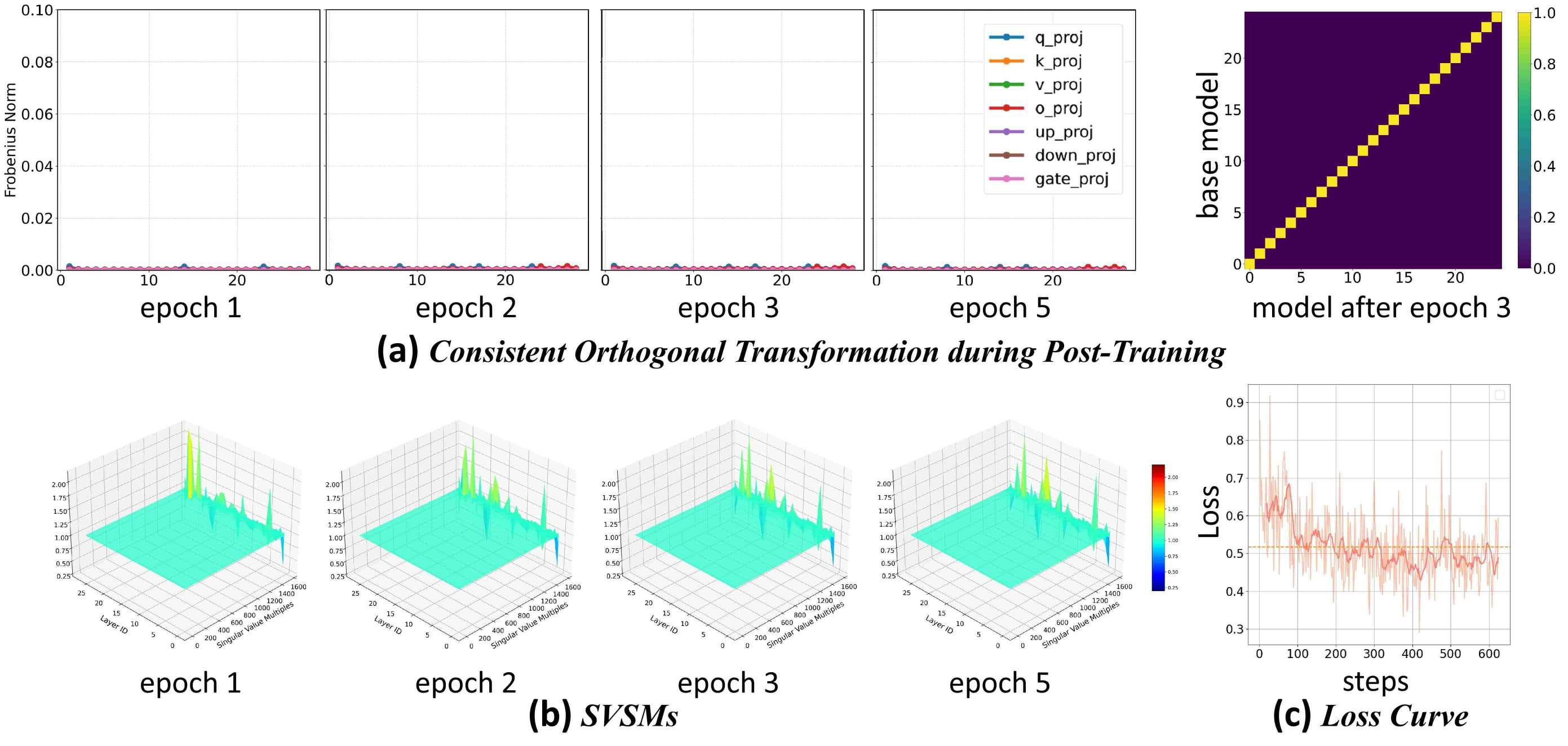}
    \caption{Observation of metrics during the post-training process. (a) presents the $\smash{\mathcal{NF}^{(i)}}$ for each checkpoint relative to the \textsc{base} model, all of which remain at an extremely low level. We also display the $I_{orth}$ of $W_o$ between the first Transformer block of the checkpoints corresponding to epoch 3 and the \textsc{base} model, indicating that consistent orthogonal transformations are highly established. (b) shows SVSMs during post-training, and (c) depicts the loss curve, which gradually converges over epochs.}  
    \label{duringingposttraining}
\end{figure}

To determine whether this phenomenon arises during the post-training process or is specific to the final convergence stage, we design a preliminary investigation. We fine-tune the \textit{Qwen2.5-Math-1.5B} model on the complex dataset \textit{s1K-1.1} \cite{muennighoff2025s1simpletesttimescaling} for 5 epochs using supervised learning. Checkpoints are saved after each training epoch. We subsequently compute the $\smash{\mathcal{NF}^{(i)}}$ metric and the \textit{SVSMs} between these intermediate checkpoints and the original pre-trained \textit{Qwen2.5-Math-1.5B} model. The training configuration is as follows: a maximum sequence length (\texttt{max\_length}) of 1024, a batch size of 16, the \textit{AdamW} optimizer \cite{loshchilov2019decoupledweightdecayregularization}, a learning rate of $2\times10^{-5}$, and no gradient accumulation. The evolution of $\smash{\mathcal{NF}^{(i)}}$ and SVSMs throughout the post-training phase is depicted in Figure \ref{duringingposttraining}.

It can be observed that during the training process, the parameter space of the model still closely adheres to the principle of structural transformation mentioned in the main text. This indicates that this phenomenon is an inherent characteristic of the changes in model parameters, rather than a property that only emerges after model convergence.

\section{Potential Applications of Our Findings}
\label{potential_applications}
\begin{figure}[!htbp]
    \centering
    \includegraphics[width=0.8\linewidth]{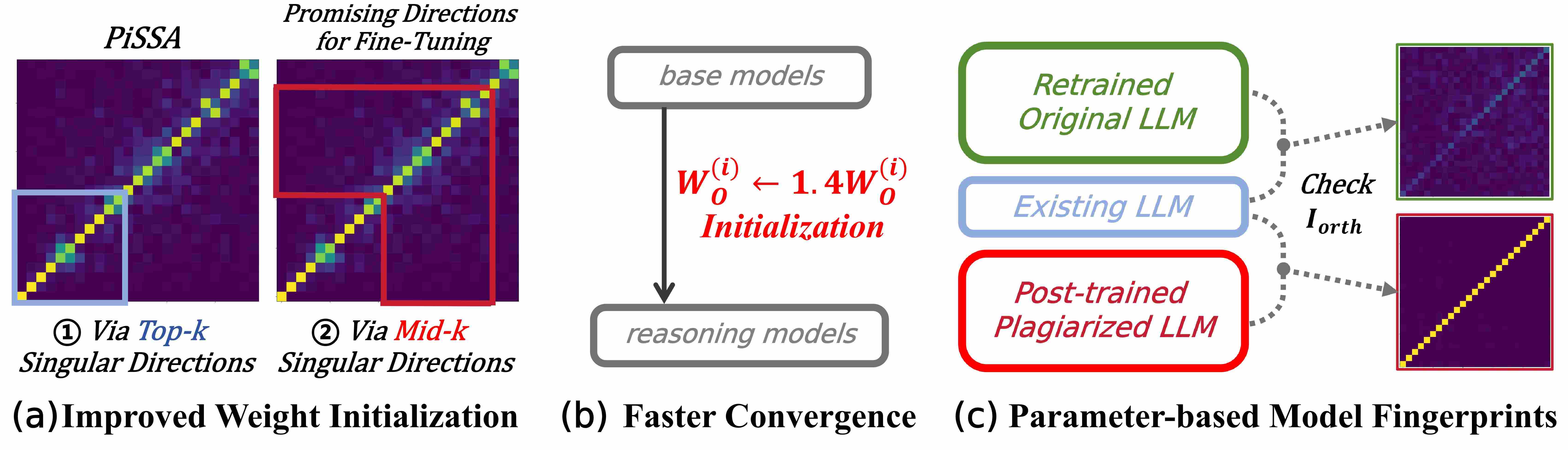}
    \caption{Illustrative overview of potential applications suggested by our findings: (a) fine-grained initialization strategies; (b) accelerated convergence in \textsc{reasoning} models; (c) model fingerprinting based on the detection of $I_{orth}$.}
    \label{fig:applications}
\end{figure}

While our primary focus is to characterize the structural transformations of LLMs induced by post-training, our analysis also points to several promising avenues for application. This section outlines a set of illustrative directions, intended not as definitive claims but as conceptual extensions of our findings, with the goal of inspiring future research and advancing the understanding of parameter-level transformations. An overview of these potential applications is provided in Figure \ref{fig:applications}.

\textbf{Fine-grained initialization strategies. }From a post-training perspective, the observed coordinated rotation of singular vectors could inspire more fine-grained weight initialization strategies. A novel approach, termed \textit{PiSSA} \cite{meng2024pissa}, preserves key components of singular vectors and singular values by initializing them as LoRA weights, while retaining and freezing the remaining singular components. However, \textit{PiSSA} primarily fine-tunes the principal components corresponding to the top-$k$ singular directions. Our analysis of $sim_U$ and $sim_V$ (Figures \ref{orth_all_visi_WQ}–\ref{orth_all_visi}) reveals that the singular vectors associated with the largest singular values ($\sigma_{\text{max}}$) exhibit minimal rotation during post-training. This observation implies that the dominant singular components are not the primary targets of fine-tuning. Consequently, as shown in Figure \ref{fig:applications}a, directing fine-tuning toward the middle-$k$ components rather than the top-$k$ may yield improved performance. This concept closely aligns with the implementation of \textit{MiLoRA} \cite{wang2025miloraharnessingminorsingular}.

\textbf{Potentially accelerated convergence in \textsc{reasoning} models.} We find that the singular value dynamics of \textsc{reasoning} models exhibits unique scaling patterns, particularly in matrices such as $W_{O}$ (as demonstrated in Figures \ref{SVSMofBaseAndInstruct} and \ref{SVSMofBaseAndInstruct_appendix}). Motivated by this observation, one may hypothesize that simple rescaling of pretrained singular values could accelerate convergence during reasoning-oriented training. For instance, initializing $W_O$ as $\alpha W_O$ with $\alpha = 1.4$ provides a lightweight mechanism to impose reasoning-like spectral properties in a single step, potentially reducing the number of iterations required to reach stable performance. While speculative, this perspective highlights the potential to exploit post-training geometry for more efficient model development.

\textbf{Model fingerprints under fully parameterized testing.} Appendix \ref{Visualization_of_other_families_and_scales_orth_sec2} demonstrates that the weight matrices of the same model architecture exhibit markedly different behaviors in $I_{\text{orth}}$ after undergoing distinct pre-training and post-training procedures. This observation provides a practical criterion for distinguishing whether a large language model has been fully developed from scratch or merely obtained through post-training on another model. As illustrated in Figure \ref{fig:applications}c, this distinction can be achieved simply by measuring the deviation between $I_{\text{orth}}$ and the identity matrix $I$. Importantly, since disrupting the coordinated rotational structure directly leads to model collapse, potential plagiarists cannot eliminate the discrepancy between their model and the original one by deliberately altering this property. Consequently, $I_{\text{orth}}$ serves as a robust and discriminative fingerprint for model identification. Moreover, because this method relies solely on parameter-level analysis, it does not require the design of evaluation datasets as in representation-based fingerprinting approaches such as \textit{REEF} \cite{zhang2024reef}. This line of investigation highlights a promising avenue for safeguarding the intellectual property rights of LLM developers.

While the potential applications discussed above represent relatively straightforward extensions of our observations, their concrete implementation and validation require more rigorous empirical investigation. Nevertheless, we hope that these preliminary intuitions will serve to inspire future research and provide readers with a deeper understanding of the broader implications of our findings for model design, optimization, and interpretability.

\section{Proof}
\label{proof}
This section mainly integrates all the mathematical proofs mentioned in the main paper.

\subsection{Training is to Perform Orthogonal Transformation on $U$ and $V$ matrices}
\label{Proof_of_Orthogonality}
Considering $\mathcal{M}_A \rightarrow \mathcal{M}_B$ as the model training process, left singular vectors of $W_A \in \mathcal{M}_A,\;W_A \in \mathbb{R}^{m \times n} $ can be regarded as performing different transformations $Q_U$:
\begin{equation}
    U_B=U_AQ_U
    \label{transformation}
\end{equation}
We first prove that $Q_U$ is an orthogonal matrix. For $Q_U$, we have:
\begin{equation}
    \begin{split}
        U^T_AU_B=U^T_AU_A \cdot Q_U=I \cdot Q_U=Q_U
    \end{split}    
    \label{transformation_2}
\end{equation}
$Q^TQ=I$ is a necessary and sufficient condition for $Q$ to be an orthogonal matrix. We calculate $Q_U^TQ_U$ then have:
\begin{equation}
    \begin{split}
        Q_U^TQ_U=(U^T_AU_B)^T \cdot (U^T_AU_B)=U^T_B\cdot(U_A  U^T_A)\cdot U_B=I
    \end{split}    
    \label{transformation_}
\end{equation}
Therefore $Q_U$ is an orthogonal matrix. 

Through experiments, we observe that $V_A^{T} V_B$ is nearly identical to $Q_U = U_A^{T} U_B$. Under the condition that $V_A^{T} V_B$ is an orthogonal matrix, we aim to prove that the column spaces of $V_A$ and $V_B$ have the same subspace structure, i.e., $\operatorname{col}(V_A) = \operatorname{col}(V_B)$, and that $V_B$ can be obtained from $V_A$ through an orthogonal transformation. Specifically, we will prove that there exists an orthogonal matrix $Q_V$ such that $V_B = V_A Q_V$, where $Q_V = V_A^{T} V_B$.

Because \(V_A\) and \(V_B\) have orthonormal columns, \(V_A^{T}V_B\) is an \(m\times m\) matrix. We are given that \(Q_V=V_A^{T}V_B\) is orthogonal, hence
\begin{equation}
    Q_V^{T}Q_V = I
\end{equation}

We define the orthogonal projector onto the column space of $V_A$ as $P_{V_A} = V_A V_A^{T}$. Decompose \(V_B\) into the sum of its projection onto \(\operatorname{col}(V_A)\) and the orthogonal remainder:
\begin{equation}
    V_B = P_{V_A} V_B + (I - P_{V_A}) V_B = V_A (V_A^{T} V_B) + (I - V_A V_A^{T}) V_B
\end{equation}
Using the definition \(Q_V=V_A^{T}V_B\) this becomes
\begin{equation}
    V_B = V_A Q_V + (I - V_A V_A^{T}) V_B
\end{equation}
To show \((I - V_A V_A^{T}) V_B = 0\), consider its Frobenius norm:
\begin{equation}
    \|(I - V_A V_A^{T}) V_B\|_F^{2} 
= \operatorname{tr}\big( V_B^{T} (I - V_A V_A^{T}) V_B \big)
\end{equation}
Expand the trace:
\begin{equation}
    \operatorname{tr}\big( V_B^{T} (I - V_A V_A^{T}) V_B \big)=\operatorname{tr}\big( V_B^{T} V_B \big) - \operatorname{tr}\big( V_B^{T} V_A V_A^{T} V_B \big)
\end{equation}
Since \(V_B\) has orthonormal columns, \(V_B^{T}V_B=I\), so the first term equals \(\operatorname{tr}(I)=m\). For the second term use cyclicity of trace and the definition of \(Q_V\):
\begin{equation}
    \operatorname{tr}\big( V_B^{T} V_A V_A^{T} V_B \big)
    = \operatorname{tr}\big( (V_A^{T}V_B)^{T} (V_A^{T}V_B) \big)
    = \operatorname{tr}\big( Q_V^{T} Q_V \big)
\end{equation}
Because \(Q_V\) is orthogonal, \(Q_V^{T}Q_V=I\), hence
\begin{equation}
    \operatorname{tr}\big( Q_V^{T} Q_V \big) = \operatorname{tr}(I)=m
\end{equation}
Combining these equalities gives
\begin{equation}
    \|(I - V_A V_A^{T}) V_B\|_F^{2} = m - m = 0
\end{equation}
Therefore
\begin{equation}
    (I - V_A V_A^{T}) V_B = 0
\end{equation}
and consequently
\begin{equation}
    V_B = V_A Q_V
\end{equation}

From \(V_B = V_A Q_V\) and the fact that \(Q_V\) is invertible (orthogonal), the column spaces are identical:
\begin{equation}
    \operatorname{col}(V_B) = \operatorname{col}(V_A Q_V) = \operatorname{col}(V_A)
\end{equation}
This completes the proof. From this perspective, the orthogonal bases utilized during the post-training are essentially \textbf{the same as those formed in the \textsc{base} models}. This fundamentally implies that post-training does not disrupt the output subspaces constructed during pre-training, strongly suggesting that it constitutes merely a reparameterization process of the \textsc{base} models.

\subsection{Singular Value Scaling Modulates the Attention Score}
\label{Proof_of_Replacement}
Under near-uniform geometric scaling with singular values, Equation \ref{approx_W} can be restated as $\smash{W_\text{post} \approx \alpha \cdot U_\text{post}\Sigma_\text{base}V^T_\text{post}=\alpha \cdot W'_\text{post}}$, which means scaling the singular values has the same effect as scaling the entire weight matrix. We uniformly apply this linear scaling effect to all weight matrices in SAs and FFNs, resulting in the following modified forms of Equations \ref{SA} and \ref{FFN}:
\begin{equation}
  SA(h) \approx \text{softmax} \left( \frac{\boldsymbol{\alpha^{2}} \cdot hW_Q' \cdot [K_{\text{cache}}' ; hW_K' ]^T}{\sqrt{d}} \right) \cdot [ V_{\text{cache}}';hW_V' ]\cdot W_O' \cdot \boldsymbol{\alpha \alpha_{O}}
  \label{temperature_SA}
\end{equation}
\begin{equation}
  FFN(z) \approx (SwiGLU(z\cdot W'_{gate}\cdot \boldsymbol{\alpha})\odot (z\cdot W'_{up}))\cdot W'_{down} \cdot \boldsymbol{\alpha}^{\mathbf{2}}
  \label{temperature_FFN}
\end{equation}
The term $\boldsymbol{\alpha}^2$ in Equation~\ref{temperature_SA} corresponds to the inverse of the \textit{attention temperature} \cite{vaswani2023attentionneed}, \textbf{which can be directly expressed by $\boldsymbol{T=1/\alpha^2}$}. In SAs, all $\alpha$ except $\alpha_O$ of \textsc{reasoning} models are consistently below $1$ after post-training  (demonstrated in Table \ref{svsf_table}), which corresponds to a higher attention temperature. This causes the softmax function to produce more uniformly distributed attention scores, encouraging the model to attend more evenly across all tokens and thereby enhancing its ability to capture global contextual information.

\subsection{Proof of Differently Post-Trained Models Sharing a Set of Consistent Orthogonal Transformations}
\label{shared}
We theoretically prove that different \textsc{post} models initialized from the same pretrained parameters and post-trained on data from different distributions can be transformed into each other through a set of shared orthogonal transformations. Assuming there are two \textsc{post} models $\mathcal{M}_\text{post},\mathcal{M}'_\text{post}$, we have:

\begin{equation}
    U_\text{post}=U_\text{base}Q_\text{post},\;\;V_\text{post}=V_\text{base}Q_\text{post}    
    \label{proof_shared}
\end{equation}
\begin{equation}
U'_\text{post}=U_\text{base}Q'_\text{post},\;\;V'_\text{post}=V_\text{base}Q'_\text{post}
\label{proof_shared_2}
\end{equation}
Substituting Equation \ref{proof_shared} into \ref{proof_shared_2}, we have:
\begin{equation}
\begin{split}
   U'_\text{post}=(U_\text{post}Q^T_\text{post})\cdot Q'_\text{post}=U_\text{post}\cdot (Q^T_\text{post}Q'_\text{post})\\
V'_\text{post}=(V_\text{post}Q^T_\text{post})\cdot Q'_\text{post}=V_\text{post}\cdot (Q^T_\text{post}Q'_\text{post}) 
\end{split}
\label{proof_shared_3}
\end{equation}
Let $Q_\text{combined}=Q^T_\text{post}Q'_\text{post}$, then we observe that:
\begin{equation}
Q^T_\text{combined}Q_\text{combined}=(Q^T_\text{post}Q'_\text{post})^T(Q^T_\text{post}Q'_\text{post})=I
\label{combined_shared}
\end{equation}
$Q_\text{combined}$ is an orthogonal matrix. This directly shows that the conversion from $\mathcal{M}_\text{post}\rightarrow \mathcal{M}'_\text{post}$ can be transformed using an approximately consistent orthogonal matrix $Q_\text{combined}$. 

This significant corollary reveal that both in-distribution fine-tuning (e.g., instruction tuning) and out-of-distribution fine-tuning (e.g., Long-CoT distillation) induce equivalent transformations in parameter space—specifically, different post-training methods can be mutually converted through shared orthogonal transformations. This equivalence explains why LLMs can be fine-tuned on arbitrary data distributions to improve task-specific performance: \textbf{the model’s input and output subspaces undergo orthogonal transformations optimized for the target task distribution}.

We believe this insight offers significant promise for future research, particularly in developing methods to mitigate forgetting while preserving adaptability.

\section{Settings}
This section will delve into more detailed experimental setups, including the different system prompts used for various datasets and the precision of models.

\subsection{System Prompts}
\label{system_prompt}
The datasets used in this study include GSM8K, MATH-500, MMLU, and GPQA. Due to time and cost constraints, we limit the output tokens to 1024. If a simple system prompt is used directly, models (particularly \textsc{reasoning} models) often require more tokens to generate correct answers when handling challenging datasets like GPQA. This would result in truncated outputs due to the token limit, preventing us from obtaining valid results for performance evaluation. Therefore, we need to design distinct system prompts for different datasets to facilitate observation of the outcomes.  

Additionally, since some datasets provide descriptive ground-truth answers (e.g., GSM8K and MATH-500) while others present multiple-choice questions (e.g., MMLU and GPQA), we must also process the inputs differently across datasets to ensure accurate performance validation.

For the simple dataset (GSM8K) mentioned in this article, the unified system prompt we adopted is:

{\centering\fbox{\textbf{\textit{Please put your final answer within \textbackslash boxed\{\}.}}}\par}

Additionally, all visualization results, including the tracking of attention entropy and the analysis of CKA heatmaps, also adopt this simple system prompt. This is attributed to the fact that during visual analysis of the model, comprehensive output results or testing performance metrics are not required for evaluation purposes.

For hard datasets (MATH-500, MMLU and GPQA) mentioned in this article, the unified system prompt we adopted is:

{\centering\fbox{\textbf{\textit{Please put your final answer within \textbackslash boxed\{\} and keep your thought process as short as possible.}}}\par}

This system prompt will enable us to effectively measure the performance on hard datasets of models within limited token computations.

For the multiple-choice question datasets (MMLU and GPQA) mentioned in this text, the template we adopted for all input prompts is as follows:

\begin{center}
\fbox{%
  \begin{minipage}{0.8\linewidth} 
    \textit{\{\textbf{ORIGINAL QUESTION}\}}
        
    \textit{You have four options, and they are:}

    \textit{A.\{\textbf{CHOICE A}\}}
    
    \textit{B.\{\textbf{CHOICE B}\}}
    
    \textit{C.\{\textbf{CHOICE C}\}}
    
    \textit{D.\{\textbf{CHOICE D}\}}

    \textit{Please select the correct option and just give A, B, C or D. For example, if you think the answer is A, just give \textbackslash boxed\{A\} as the answer.}
  \end{minipage}%
}
\end{center}

This template design enables us to use the same validation evaluator for both multiple-choice and open-ended answer datasets, thereby reducing our engineering complexity.

\subsection{Introduction to the Models and Model Precision Settings}
\label{models}
The different \textsc{post} versions corresponding to the different \textsc{base} models are shown in Table \ref{version} and \ref{version_case_study}. All experiments in this paper were conducted on two NVIDIA A100 GPUs with 40GB of memory each. 
\begin{table}[!htbp]
  \caption{Different \textsc{post} versions of different \textsc{base} models used in Appendix \ref{Visualization_of_other_families_and_scales}, \ref{Visualization_of_other_families_and_scales_orth}, \ref{replaced_appendix} and \ref{restoration}.}
  \label{version}
  \centering
  \resizebox{0.6\textwidth}{!}{
  \begin{tabular}{llll}
    \toprule
    \multirow{1}{*}{\textsc{base} Models}        & \textsc{post} Types & \textsc{post} Models & Developer\\
    \midrule
    \multirow{2}{*}{\textit{Qwen2.5-Math-1.5B}} & $\mathcal{M}_\text{Instruct}$  & \textit{Qwen2.5-Math-1.5B-Instruct}   & \textit{Qwen Team}\\
                                 & ${\mathcal{M}_\text{reasoning}}$& \textit{DeepSeek-R1-Distill-Qwen-1.5B} & \textit{DeepSeek}\\
    \midrule
    \multirow{2}{*}{\textit{Qwen2.5-Math-7B}} & $\mathcal{M}_\text{Instruct}$  & \textit{Qwen2.5-Math-7B-Instruct} & \textit{Qwen Team}\\
                                 & ${\mathcal{M}_\text{reasoning}}$ & \textit{DeepSeek-R1-Distill-Qwen-7B}& \textit{DeepSeek}\\

    \midrule
    \multirow{2}{*}{\textit{Llama-3.1-8B}} & $\mathcal{M}_\text{Instruct}$  & \textit{Llama-3.1-8B-Instruct}& \textit{Meta}\\
                                 & ${\mathcal{M}_\text{reasoning}}$ & \textit{DeepSeek-R1-Distill-Llama-8B}& \textit{DeepSeek}\\

    \midrule
    \multirow{2}{*}{\textit{Qwen2.5-14B}} & $\mathcal{M}_\text{Instruct}$  & \textit{Qwen2.5-14B-Instruct}& \textit{Qwen Team}\\
                                 & ${\mathcal{M}_\text{reasoning}}$ & \textit{DeepSeek-R1-Distill-Qwen-14B}& \textit{DeepSeek}\\
    \bottomrule
  \end{tabular}}
\end{table}
\begin{table}[!htbp]
  \caption{Different \textsc{post} versions of different \textsc{base} models used in Appendix \ref{case_study}.}
  \label{version_case_study}
  \centering
  \resizebox{0.75\textwidth}{!}{
  \begin{tabular}{llll}
    \toprule
    \multirow{1}{*}{\textsc{base} Models}  & \textsc{post} Models  & post-training method  & Developer\\
    \midrule
    \textit{DeepSeek-R1-Distill-Qwen-7B} & \textit{AceMath-RL-Nemotron-7B}  & RL-based (GRPO)  & \textit{Nvidia}\\
    \midrule
    \textit{deepseek-math-7b-base} & \textit{deepseek-math-7b-rl}  & RL-based (GRPO)  & \textit{Deepseek}\\
    \midrule
    \textit{Seed-X-Instruct-7B} & \textit{Seed-X-PPO-7B}  & RL-based (PPO)  & \textit{ByteDance}\\
    \midrule
    \textit{Mistral-7B-v0.1} & \textit{Mistral-7B-Instruct-v0.1}  & SFT-based & \textit{Mistral AI}\\
    \midrule
    \textit{gemma-2-2b} & \textit{gemma-2-2b-it}  & SFT-based & \textit{Google}\\
    \midrule
    \textit{MediPhi} & \textit{MediPhi-Instruct}  & SFT-based & \textit{Microsoft}\\
    \bottomrule
  \end{tabular}}
\end{table}

All $\mathcal{M}_\text{base}$ and $\mathcal{M}_\text{Instruct}$ use BF16 parameter storage, while $\mathcal{M}_\text{reasoning}$ employ FP32. To address potential precision truncation, we consistently convert all parameters to FP32 before experimentation, ensuring unified numerical precision throughout our evaluations.


\end{document}